%% file: main.tex
\documentclass[11pt, a4paper, logo, copyright]{googledeepmind}

\usepackage[authoryear, sort&compress, round]{natbib}
\usepackage{graphicx}
\usepackage{tikz}
\usetikzlibrary{spy}
\usepackage{booktabs}
\usepackage{colortbl}
\usepackage{makecell}
\usepackage{subcaption}
\usepackage{multicol}
\usepackage{multirow}
\usepackage{pgfplots} 
\usepackage[utf8]{inputenc}
\usepackage{longtable}
\pgfplotsset{compat=newest}
\usetikzlibrary{patterns}
\usepackage{pgfplots, pgfplotstable}
\usepackage{enumitem}

\definecolor{mygray}{HTML}{bdc1c6}
\definecolor{myblue}{HTML}{669df6}
\definecolor{myyellow}{HTML}{fde293}

\usepackage{siunitx}
\newcommand{\dalle}[0]{DALL·E~3}
\newcommand{\midjourney}[0]{Midjourney~v6}
\newcommand{\sdxl}[0]{SDXL~1}
\newcommand{\sdthree}[0]{SD3}
\newcommand{\imagenthree}[0]{Imagen~3}
\newcommand{\imagentwo}[0]{Imagen~2}

\newcommand{\clip}[0]{CLIP}
\newcommand{\vqascore}[0]{VQAScore}
\newcommand{\gecko}[0]{Gecko}
\newcommand{\geckorel}[0]{Gecko-Rel}
\newcommand{\doccionek}[0]{DOCCI-Test-Pivots}
\newcommand{\dalleeval}[0]{Dall·E~3 Eval}
\newcommand{\genaibench}[0]{GenAI-Bench}
\newcommand{\drawbench}[0]{DrawBench}
\newcommand{\pz}{\hphantom{0}}

\title{\imagenthree{}}

\reportnumber{} %

\renewcommand{\today}{}

\author{\imagenthree{} Team, Google\footnote{See Contributions section for full author list. Please send correspondence to imagen-report@google.com.}}

\begin{abstract}
We introduce \imagenthree{}, a latent diffusion model that generates high quality images from text prompts. We describe our quality and responsibility evaluations. \imagenthree{} is preferred over other state-of-the-art (SOTA) models at the time of evaluation. In addition, we discuss issues around safety and representation, as well as methods we used to minimize the potential harm of our models.
\end{abstract}

\begin{document}

\maketitle

\begin{figure}
\centering
\setlength{\fboxsep}{0pt}%
\fbox{\includegraphics[width=\textwidth]{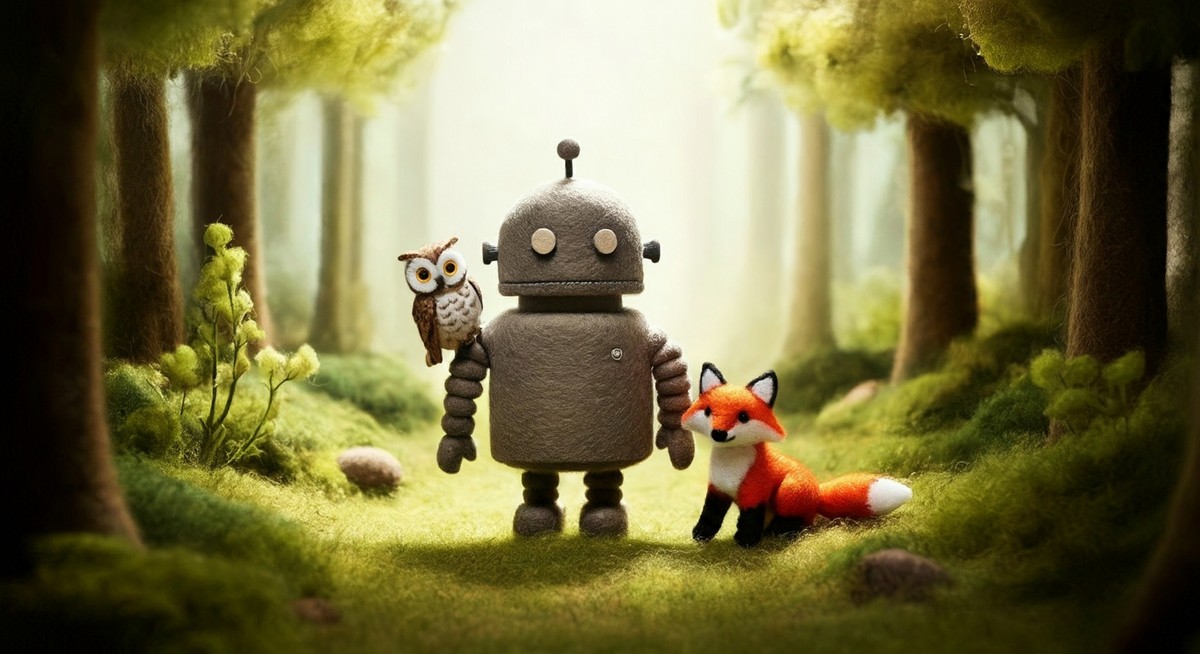}} \\
\caption{\imagenthree{} is our best diffusion model for text-to-image generation, capable of following descriptive prompts, such as ``\textit{Photo of a felt puppet diorama scene of a tranquil nature scene of a secluded forest clearing with a large friendly, rounded robot is rendered in a risograph style. An owl sits on the robots shoulders and a fox at its feet. Soft washes of color, 5 color, and a light-filled palette create a sense of peace and serenity, inviting contemplation and the appreciation of natural beauty.}''}
\label{fig:teaser}
\end{figure}

\section{Introduction}

Text-to-image (T2I) models drive a number of use cases, for example in image generation and editing, as well as scene understanding. In this tech report, we outline the training and evaluation of the latest model in Google's Imagen family, \imagenthree{}. At its default configuration, \imagenthree{} generates images at $1024\times1024$ resolution, and can be followed by $2\times$, $4\times$, or $8\times$ upsampling. We describe our evaluations and analysis against other state-of-the-art T2I models. We find \imagenthree{} is preferred over other models. In particular, it performs well at photorealism, and in adhering to long and complex user prompts. Deploying T2I models introduces many new challenges, we describe in detail experiments focused on understanding the safety and responsibility risks associated with this model family, along with our efforts to reduce potential harms.

\section{Data}
\label{sec:data}

Our model is trained on a large dataset comprising images, text and associated annotations. To ensure quality and safety standards, we employ a multi-stage filtering process. This process begins by removing unsafe, violent, or low-quality images. We then eliminate AI-generated images to prevent the model from learning artifacts or biases commonly found in such images. Additionally, we use deduplication pipelines and down-weight similar images to minimize the risk of outputs overfitting particular elements of training data.

Each image in our dataset is paired with both original (sourced from alt text, human descriptions, etc.), and synthetic captions~\citep{betker2023improving}. Synthetic captions are generated using Gemini models with a variety of prompts. We leverage multiple Gemini models and instructions to maximize the linguistic diversity and quality of these synthetic captions~\citep{iiw2024}. We apply filters to remove unsafe captions and personally identifiable information.

\section{Evaluation}
\label{sec:evaluation}
We compare our highest quality configuration -- the \imagenthree{} model  -- against \imagentwo{} and the following external models:
\dalle{}~\citep{betker2023improving},
\midjourney{},
Stable Diffusion 3 Large~\citep[\sdthree{},][]{esser2024scaling},
and Stable Diffusion XL 1.0~\citep[\sdxl{},][]{podell2023sdxl}.
Through extensive human (Sec.~\ref{sec:humanevaluation}) and automatic (Sec.~\ref{sec:automaticeval}) evaluations we find that \imagenthree{} sets a new state of the art in text-to-image generation.
We discuss the overall results and limitations in Section~\ref{sec:overall_and_limitations} and Section~\ref{sec:qualitative} includes qualitative results. We note that products that may incorporate \imagenthree{} may exhibit differing performance to the tested configuration.

Please refer to Appendix~\ref{app:imagen3-002} for updated human evaluation results as of December 2024.

\subsection{Human Evaluation}
\label{sec:humanevaluation}
We run human evaluations on five different quality aspects of a text-to-image generation model: overall preference (Sec.~\ref{subsection:overall_preference}),
prompt--image alignment (Sec.~\ref{subsection:t2i_alignment}), visual appeal (Sec.~\ref{subsection:visual_appeal}), detailed prompt--image alignment (Sec.~\ref{subsection:content_recreation}), and numerical reasoning (Sec.~\ref{subsection:numerical_reasoning}).
Each of these aspects are evaluated independently in order to avoid conflation in raters’ judgments.

For the first four aspects, quantitative judgment (e.g.\ assigning a score between 1 and 5) is in practice difficult to calibrate across raters. We therefore use side-by-side comparisons; this is also becoming a standard practice in chatbot~\citep{chiang2024chatbot} and other text-to-image~\citep{betker2023improving} evaluations. evaluations. The fifth aspect -- numerical reasoning -- can directly and reliably be evaluated by humans by counting how many objects of a given type are depicted in an image, so we follow this single-model evaluation approach.

Each side-by-side comparison (i.e.\ for the first four aspects and their corresponding prompt sets) is aggregated into an Elo score~\citep{nichol2021glide,betker2023improving} for all six models to get a calibrated comparison between them. Intuitively, each pairwise comparison represents a match played between two models, with the Elo score representing a model’s overall score in the competition among all models. We generate the complete Elo scoreboard on each aspect and prompt set through exhaustive comparison of every pair of models. Each study (a pairing between two models on a given question and given prompt set) consists of~\num{2500} ratings (we found this number to be a good trade-off between cost and reliability) which are uniformly distributed among the prompts in the prompt set. The models are anonymized in the rater interface and the sides are randomly shuffled for every rating.

We use an external platform to randomly select raters from an extensive and varied pool. Data collection is undertaken in accordance with Google DeepMind’s best practices on data enrichment~\citep{GoogleDeepMind2022}, based on the Partnership on AI’s Responsible Sourcing of Data Enrichment Services~\citep{PAI2021}. This includes ensuring all data enrichment workers are paid at least a local living wage.

We run human evaluations on \num{5} different prompt sets in total. We evaluate the first three quality aspects (overall preference, prompt-image alignment, and visual appeal) on three different prompt sets. First, we use the recently-released \genaibench{}~\citep{lin2024evaluating}, a set of \num{1600} high-quality prompts collected from professional designers. To align with previous work, we also evaluate on the 200 prompts of \drawbench~\citep{saharia2022photorealistic} and the \num{170} prompts of \dalle~Eval~\citep{betker2023improving}. For detailed prompt-image alignment, we use \num{1000} images and their corresponding captions from DOCCI~\citep{OnoeDocci2024} (\doccionek). Finally, we use the GeckoNum benchmark~\citep{kajic2024evaluating} to evaluate numerical reasoning capabilities. All the external models are run via their public access offerings, except for \dalle{} on \dalle~Eval and \drawbench{}, for which we use the images released by its authors.

In total, we collected \num{366569} ratings in \num{5943} submissions from \num{3225} different raters. Each rater participated in at most 10\% of our studies, and in each study, each rater provided approximately 2\% of the ratings, to avoid biasing the results to a particular set of raters’ judgments. Raters from \num{71} different nationalities participated in our studies, with the United Kingdom, United States, South Africa, and Poland being the most represented.

\subsubsection{Overall Preference}
\label{subsection:overall_preference}
Overall preference measures the degree of satisfaction of the user with respect to the generated image given the input prompt.
It is by design an open question that leaves to the rater the decision of which quality aspects are the most important in every prompt, as is the case in a realistic usage of the model.
We showed two images to raters, side by side together with the prompt and asked: \textit{Imagine you are using a computer tool that produces an image given the prompt above. Choose which image you would prefer to see if you were using this tool. If both images are equally appealing, select ``I am indifferent''}.

Figure~\ref{fig:elo_overall_preference} shows the results on \genaibench{}, \drawbench, and \dalle~Eval. On \genaibench{}, \imagenthree{} is significantly more preferred over other models. On \drawbench, \imagenthree{} leads with a smaller margin with respect to Stable Diffusion~3 and on \dalle~Eval we observe close results for the four leading models, with \imagenthree{} having a slight edge.

\input{assets/human_evals/elo_overall_preference.tikz}

\subsubsection{Prompt--Image Alignment}
\label{subsection:t2i_alignment}
Prompt-–image alignment evaluates how well the input prompt is represented in the output image content, irrespective of potential flaws in the image or its aesthetic appeal. We showed the raters two images side by side together with the prompt and asked them: \textit{Considering the text above, which image better captures the intent of the prompt? Please try to ignore potential defects or bad quality of the images. Unless mentioned in the prompt, also disregard the different styles.}

Figure~\ref{fig:elo_t2i_alignment} shows the results on \genaibench{}, \drawbench{}, and \dalle~Eval. \imagenthree{} leads with a significant margin on \genaibench, it has smaller margin on \drawbench, and on \dalle~Eval the three leading models perform similarly with overlapping confidence intervals.

\input{assets/human_evals/elo_t2i_alignment.tikz}

\subsubsection{Visual Appeal}
\label{subsection:visual_appeal}
Visual appeal quantifies how appealing the generated images are, irrespective of the content that was requested. To measure it, we show two images side by side to the raters, without the prompt that created them, and we ask: \textit{Which image is more appealing to you?}.

Figure~\ref{fig:elo_visual_appeal} shows the results on \genaibench{}, \drawbench{}, and \dalle~Eval. Midjourney v6 leads overall, with \imagenthree{} almost on par on \genaibench{}, a slightly bigger advantage on \drawbench{}, and a significant advantage on \dalle~Eval. 

\input{assets/human_evals/elo_visual_appeal.tikz}

\subsubsection{Detailed Prompt-Image Alignment}
\label{subsection:content_recreation}
In this section we further push the evaluation of prompt-image alignment capabilities by generating images from the detailed prompts of DOCCI~\citep{OnoeDocci2024}.
These prompts are significantly longer -- \num{136} words on average -- than the prompt sets used above.
After running some pilots following the same evaluation strategy of Section~\ref{subsection:t2i_alignment}, however, we realized that reading \num{100}+ word prompts and evaluating how well the images aligned with all the details in them was too challenging and cumbersome for human raters.
We instead leveraged the fact that DOCCI prompts are actually high-quality captions of real reference photographs -- in contrast to standard text-to-image evaluation prompt sets, which have no such corresponding reference images.
We fed these captions to the image generation models and measured how well the content of the generated image aligns with that of the benchmark reference image from DOCCI. We specifically instruct the raters to focus on the semantics of the images (objects, their position, their orientation, etc.) and ignore styles, capturing technique, quality, etc.

Figure~\ref{fig:docci_elo} shows the results, in which we can see that \imagenthree{} has a significant gap of +\num{114} Elo points and \num{63}\% win rate against the second best model. This result further highlights its outstanding capabilities of following the detailed contents of the input prompts.

\input{assets/human_evals/elo_image_content_recreation.tikz}

\subsubsection{Numerical Reasoning}
\label{subsection:numerical_reasoning}
We also evaluate the capability of the models to generate an exact number of objects, following the simplest task in the GeckoNum benchmark~\citep{kajic2024evaluating}.
Specifically, we ask: \textit{How many \texttt{<obj>} are in the image?}, where \texttt{<obj>} refers to the noun in the source prompt used to generate the image and compare it to the expected quantity requested in the prompt.
The number of objects range from 1 to 10 and the task includes prompts of various complexity as numbers are embedded in different types of sentence structures, examining the role of attributes such as color and spatial relationships.

The results are shown in Figure~\ref{fig:geckonum_count}, where we see that, while generating an exact number of objects is still a challenging task for current models, \imagenthree{} is the strongest model, outperforming the second one, \dalle, by \num{12} percentage points.
In addition, we find that \imagenthree{} has higher accuracy compared to other models when generating images containing between \num{2} and \num{5} objects, as well as better performance on prompts with numerically more complex sentence structure, such as ``1 cookie and five bottles'' (See Appendix~\ref{app:eval_geckonum} for details). 

\begin{figure}
    \centering
    \resizebox{0.75\linewidth}{!}{\input{assets/human_evals/geckonum_count.tikz}}
    \caption{{\bf Numerical Reasoning:} Accuracy on Exact Number Generation in GeckoNum. \imagenthree{} is the strongest performing model with an accuracy of 58.6\%.}
    \label{fig:geckonum_count}
\end{figure}

\subsection{Automatic Evaluation}
\label{sec:automaticeval}
In recent years, automatic-evaluation (auto-eval) metrics, such as CLIP~\citep{hessel2021clipscore} and VQAScore~\citep{lin2024evaluating}, are more widely used to measure quality of text-to-image models, as they are easier to scale than human evaluations.
We run some auto-eval metrics for prompt--image alignment (Sec.~\ref{subsec:ti2_align}) and image quality (Sec.~\ref{subsec:im_qual}) to complement the human evaluation in the previous section.

\subsubsection{Prompt--Image Alignment}
\label{subsec:ti2_align}
We choose three strong auto-eval prompt--image alignment metrics from the main families of metrics: contrastive dual encoders~\citep[\clip,][]{hessel2021clipscore}, VQA-based~\citep[\gecko,][]{wiles2024revisiting}, and an LVLM prompt-based (an implementation of \vqascore\footnote{We use the same prompt as \cite{lin2024evaluating} but Gemini 1.5 Pro \citep{geminiteam2024gemini15unlockingmultimodal} as the backend.}).
While previous work has demonstrated that these metrics correlate well with human judgment~\citep[e.g.,][]{wiles2024revisiting,lin2024evaluating,cho2024davidsonian}, it is unclear if they can reliably discriminate between stronger models that are more similar to each other. As a result, we first validate the three metrics by comparing their predictions with the human ratings obtained for alignment in Sec.~\ref{subsection:t2i_alignment} and report findings in Appendix~\ref{app:auto_eval_metric_comparisons}.

We observe that \clip{} -- despite being commonly used in current work -- fails to predict the correct model ordering in most cases (see \autoref{tab:auto_eval_metrics_model_comparisons}).
We find that \gecko~and our \vqascore~variant (referred to as \vqascore~in the following) perform well and agree about 72\% of the time.
In these cases, where the metrics agree, we can have confidence in the results as they agree with human judgment 94.4\% of the time.
While they perform similarly, \vqascore~has the edge as it matches human ratings 80\% of the time as opposed to 73.3\% of the time for \gecko.
We note that \gecko~uses a weaker backbone -- PALI~\citep{chen2022pali} as opposed to Gemini 1.5 Pro -- which may account for the difference in performance. 
As a result, in the following we discuss results with \vqascore~and leave other results and further discussion on the setup to Appendix~\ref{app:auto_eval_metric_comparisons}. 

We evaluate on four datasets to investigate model differences under diverse conditions: \geckorel, \doccionek, \dalleeval, and \genaibench. \geckorel{} is designed to measure alignment and includes prompts with high inter-annotator agreement, \doccionek includes long, descriptive prompts, \dalleeval and \genaibench are more varied datasets that aim to evaluate a range of capabilities.
Results are reported in Figure~\ref{fig:geckobar_plots}. We can see that overall the best performing model under the metrics, for alignment, is \imagenthree. It performs best on the \doccionek{}'s longer prompts and consistently has the overall highest performance.
Finally, we see that \sdxl{} and \imagentwo{} are consistently less performant than the other models.

\begin{figure}
    \centering
    \resizebox{\linewidth}{!}{%
        \mbox{%
            \begin{minipage}{.5\linewidth}
                \resizebox{\linewidth}{!}{\input{assets/experiments/auto_evals/bar_plots_gecko_rel_filtered}}
            \end{minipage}
            \begin{minipage}{.5\linewidth}
                \resizebox{\linewidth}{!}{\input{assets/experiments/auto_evals/bar_plots_genaibench}}
            \end{minipage}
        }
    }\\[3mm]
    \resizebox{\linewidth}{!}{%
        \mbox{%
            \begin{minipage}{.5\linewidth}
                \resizebox{\linewidth}{!}{\input{assets/experiments/auto_evals/bar_plots_gecko_dalle3}}
            \end{minipage}
            \begin{minipage}{.5\linewidth}
                \resizebox{\linewidth}{!}{\input{assets/experiments/auto_evals/bar_plots_gecko_docci}}
            \end{minipage}
        }
    }
    \caption{{\bf \vqascore~performance on a variety of datasets.} We plot the mean performance and 95\% confidence interval as error-bars. Where error-bars overlap and groups of models are not significant, we indicate this with `ns'. Otherwise, results are significant with $p < 0.05$. To compute significance, we follow~\cite{wiles2024revisiting} and compare distributions of predictions using the Wilcoxon signed rank test. \imagenthree{} is the best performing model across datasets as measured for alignment.}
    \label{fig:geckobar_plots}
\end{figure}

We further explore, for \geckorel, the breakdown by category in Figure~\ref{fig:t2i_alignment_gecko_category_evaluation}.
We can see that, overall, \imagenthree{} is one of the best performing models.
For categories testing capabilities such as color, counting, and spatial reasoning, \imagenthree{} performs best (further validating results in Sec.~\ref{subsection:numerical_reasoning}).
We also see a difference in model performance for more complex and compositional prompts, e.g.\ prompts with more linguistic difficulty.
On complex prompts, \sdxl{} performs notably worse than the other models. On compositional prompts (where models are tasked to create multiple objects in a scene or a scene without an object), we see that \imagenthree{} performs best.
This corroborates the previous dataset findings, as \imagenthree{} was best on \doccionek, which notably has very long, challenging prompts.
These results indicate that \imagenthree{} performs best for more complex prompts and a variety of capabilities as compared to other models.

\begin{figure}
    \centering
    \hspace{-6.8mm}\resizebox{1.04\textwidth}{!}{%
    \input{assets/experiments/category_barplots.tikz}}
    \caption{{\bf Comparing T2I models using \vqascore~on the per category breakdown of prompts within \geckorel.} Error bars indicate 95\% confidence intervals obtained via bootstrapping.}
    \label{fig:t2i_alignment_gecko_category_evaluation}
\end{figure}

\subsubsection{Image Quality}
\label{subsec:im_qual}

We compare the distribution of generated images by \imagenthree{}, \sdxl{}, and \dalle{} on \num{30000} samples of the MSCOCO-caption validation set~\citep{COCO} using different feature spaces and distance metrics following the protocol in~\cite{vasconcelos2024greedy}. We take the Fr\'{e}chet distance on Inception~\citep[FID,][]{Heusel17} and Dino-v2~\citep[FD-Dino,][]{dinov2,stein2023exposing}) feature spaces, and also the MMD distance on CLIP-L feature space \citep[CMMD,][]{jayasumana2023rethinking}.
The resolution of the generated images was reduced from $1024\times1024$ pixels to each metric's standard input size.

Similarly to~\cite{vasconcelos2024greedy} we observed that the minimization of these three metrics are in trade-off with each other.
FID favors the generation of natural colors and textures, but under closer inspection, it fails to detect distortions on object shapes and parts.
Lower values of FD-Dino and CMMD favor image content.
Table~\ref{tab:automated_image_distribution_metrics} displays the results.
The FID values of both \imagenthree{} and \dalle{} reflect an intentional shift in color distribution away from MSCOCO-caption samples due to aesthetic preference for generating more vivid, stylized images.
Simultaneously, \imagenthree{} presents the lower CMMD value of the three models, highlighting its strong performance on state-of-the-art feature space metrics.

\begin{table}[ht!]
    \centering
    \small
    \begin{tabular}{lccc}
    \toprule
    & FID (↓) & FD-Dino (↓) & CMMD (↓)\\
    \midrule
    \dalle{}\hspace{1cm} & 20.1	& 284.4	& 0.894 \\
    \sdxl{} & 13.2	& 185.6	& 0.898 \\
    \imagenthree{} & 17.2	& 213.9	& 0.854 \\
    \bottomrule
    \end{tabular}
    \captionsetup{justification=centering}
    \caption{{\bf Automated Image Distribution metrics}: \imagenthree{} compared to \dalle{} and \sdxl{}}
    \label{tab:automated_image_distribution_metrics}
\end{table}

\subsection{Conclusions and Limitations}
\label{sec:overall_and_limitations}
All in all, \imagenthree{} clearly leads on prompt--image alignment (Sec.~\ref{subsection:t2i_alignment}, Sec.~\ref{subsec:ti2_align}), especially on detailed prompts (Sec.~\ref{subsection:content_recreation}) and counting abilities (Sec.~\ref{subsection:numerical_reasoning}); while on visual appeal (Sec.~\ref{subsection:visual_appeal}),
\midjourney{} takes the lead, with \imagenthree{} coming in second.
When considering all the quality aspects, \imagenthree{} clearly leads in overall preference (Sec.~\ref{subsection:overall_preference}), indicating it strikes the best balance of high quality outputs that respect user intent.

While \imagenthree{} and other current strong models achieve impressive performance, they still exhibit shortcomings in certain capabilities.
In particular, tasks that require numerical reasoning, from generating an exact number of objects to reasoning about parts, are challenging for all models.
In addition, prompts that involve reasoning about scale (e.g.\ ``the house is the same size as the cat''), compositional phrases (e.g.\ ``one red hat and a black glass book'')
and actions (``a person throws a football'') are the hardest across all models.
This is followed by prompts that require spatial reasoning and complex language.

\subsection{Qualitative Results}
\label{sec:qualitative}
\begin{figure}[h!]
\centering
\resizebox{\textwidth}{!}{%
\setlength{\fboxsep}{0pt}%
\setlength{\tabcolsep}{1pt}%
\begin{tabular}{c c c c}%
\fbox{\includegraphics[width=0.24\textwidth]{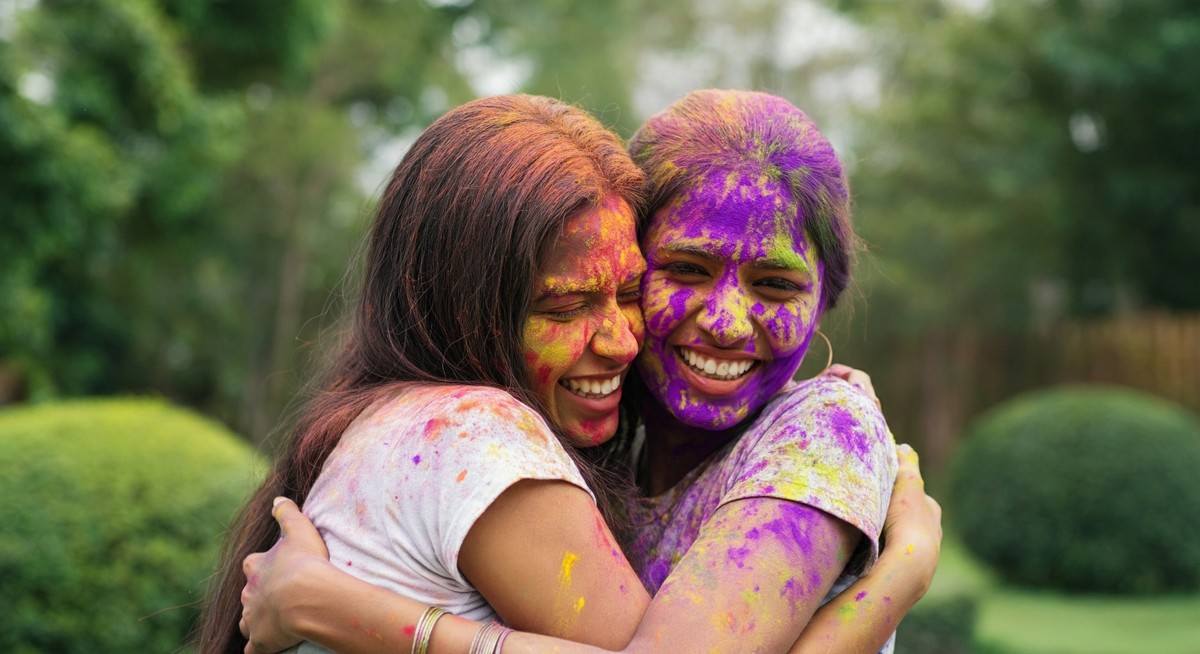}} &
\fbox{\includegraphics[width=0.24\textwidth]{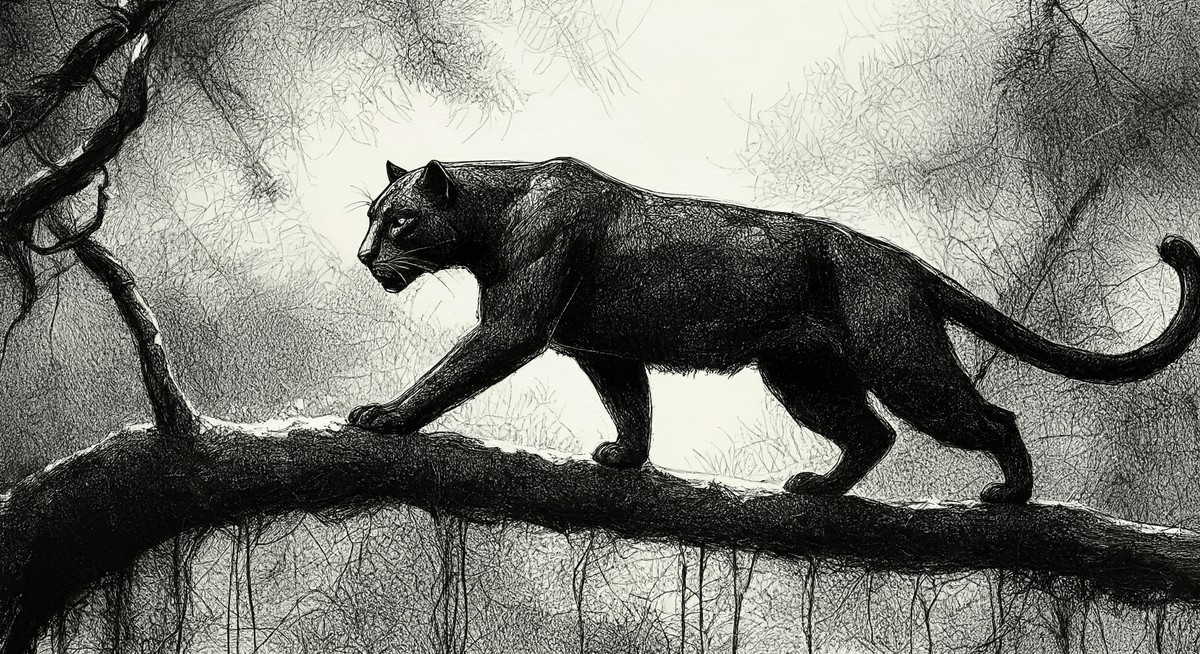}} &
\fbox{\includegraphics[width=0.24\textwidth]{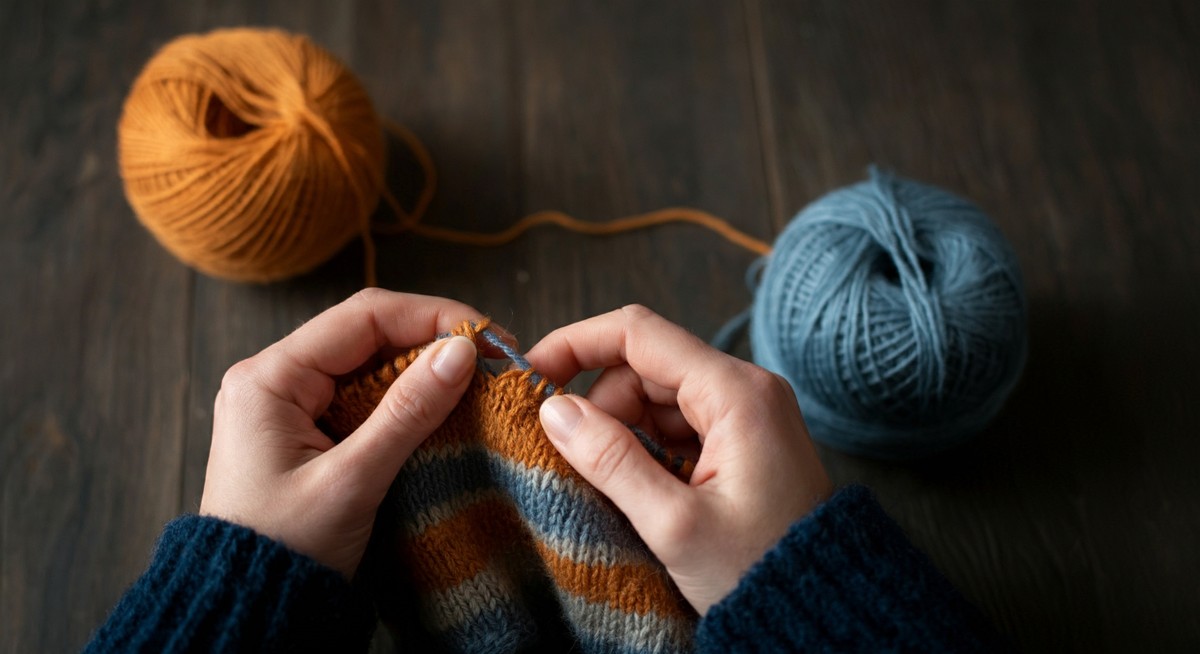}} &
\fbox{\includegraphics[width=0.24\textwidth]{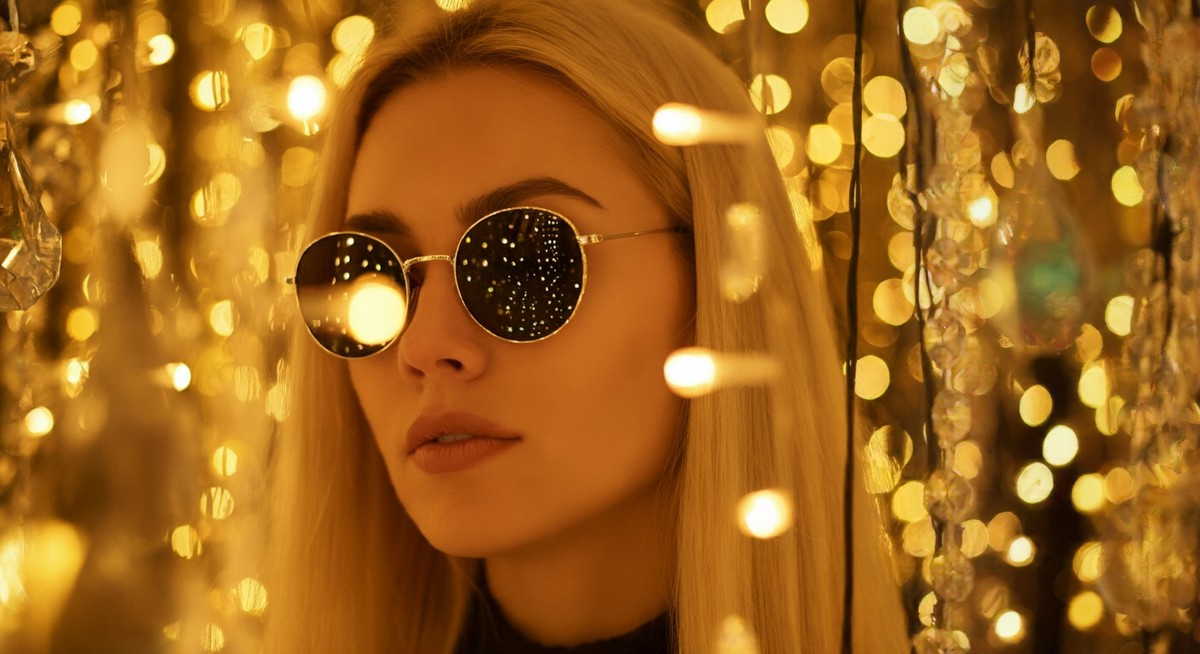}} \\[-2pt]
 
\fbox{\includegraphics[width=0.24\textwidth]{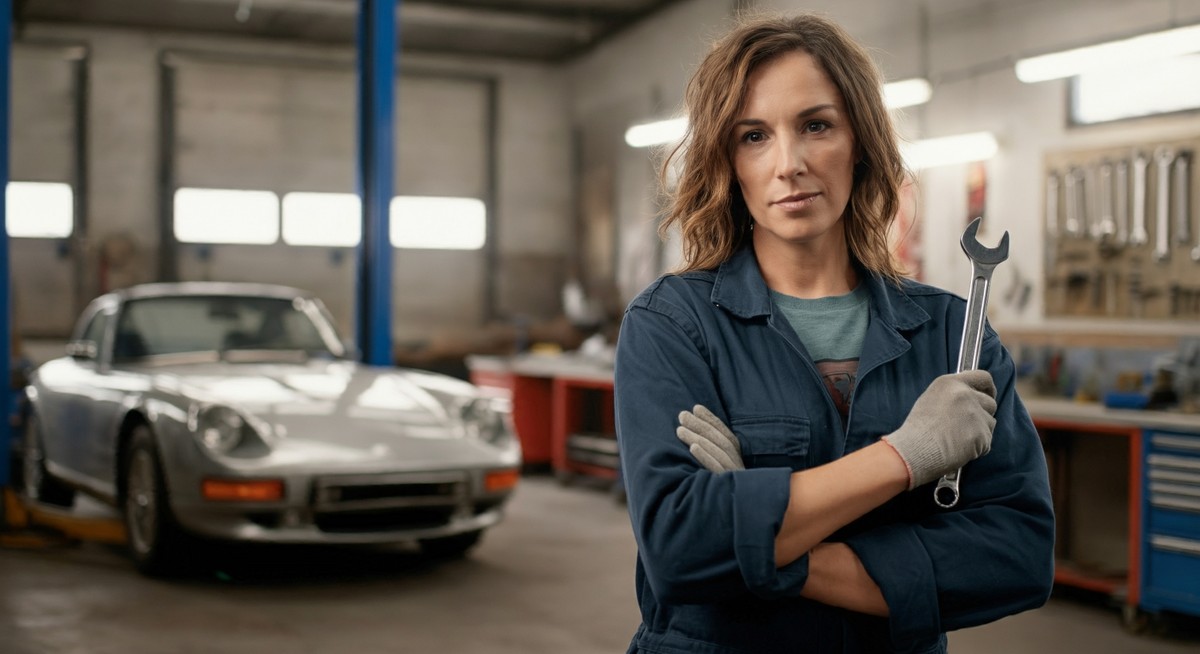}} &
\fbox{\includegraphics[width=0.24\textwidth]{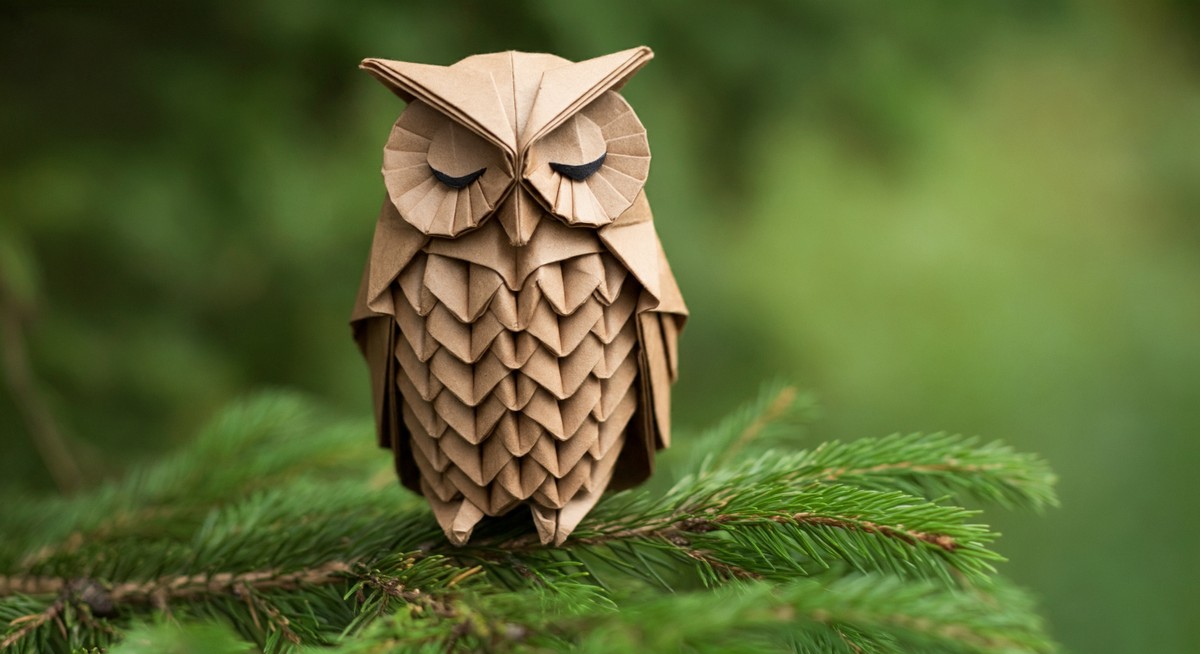}} &
\fbox{\includegraphics[width=0.24\textwidth]{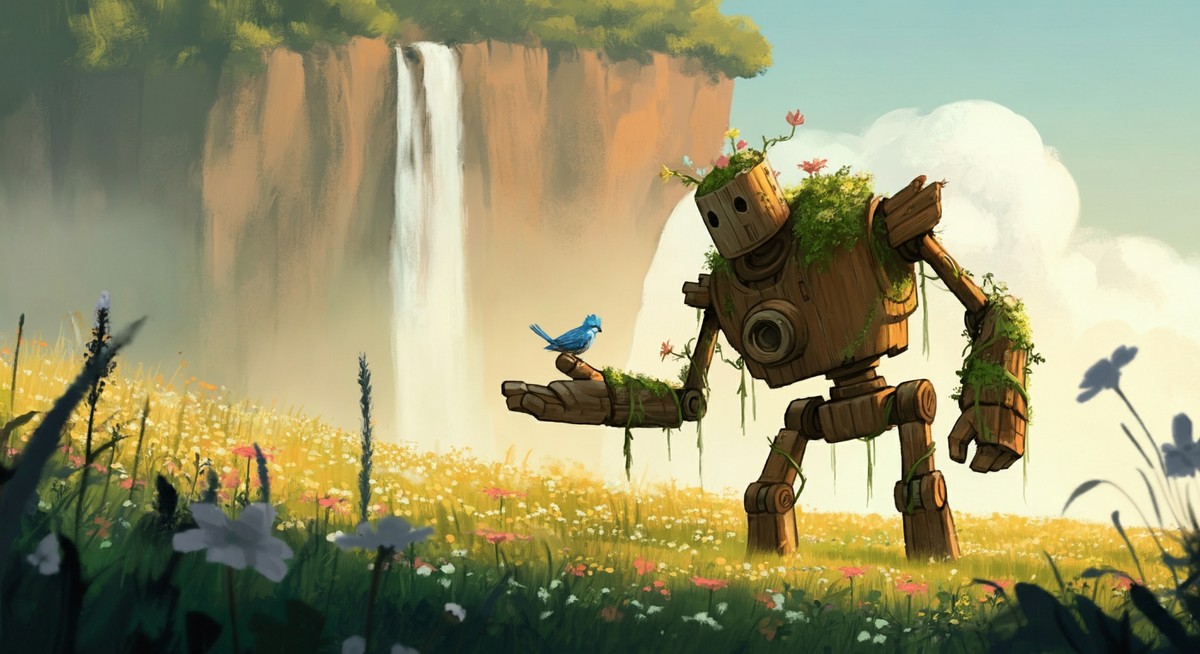}} &
\fbox{\includegraphics[width=0.24\textwidth]{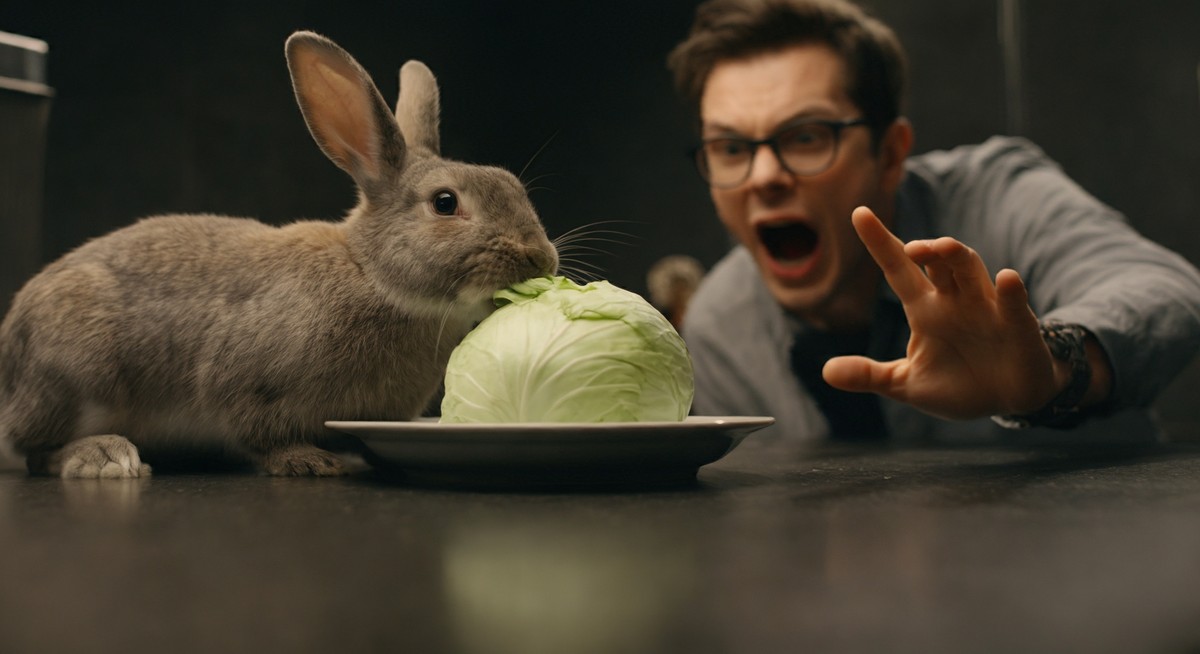}} \\[-2pt]
 
\fbox{\includegraphics[width=0.24\textwidth]{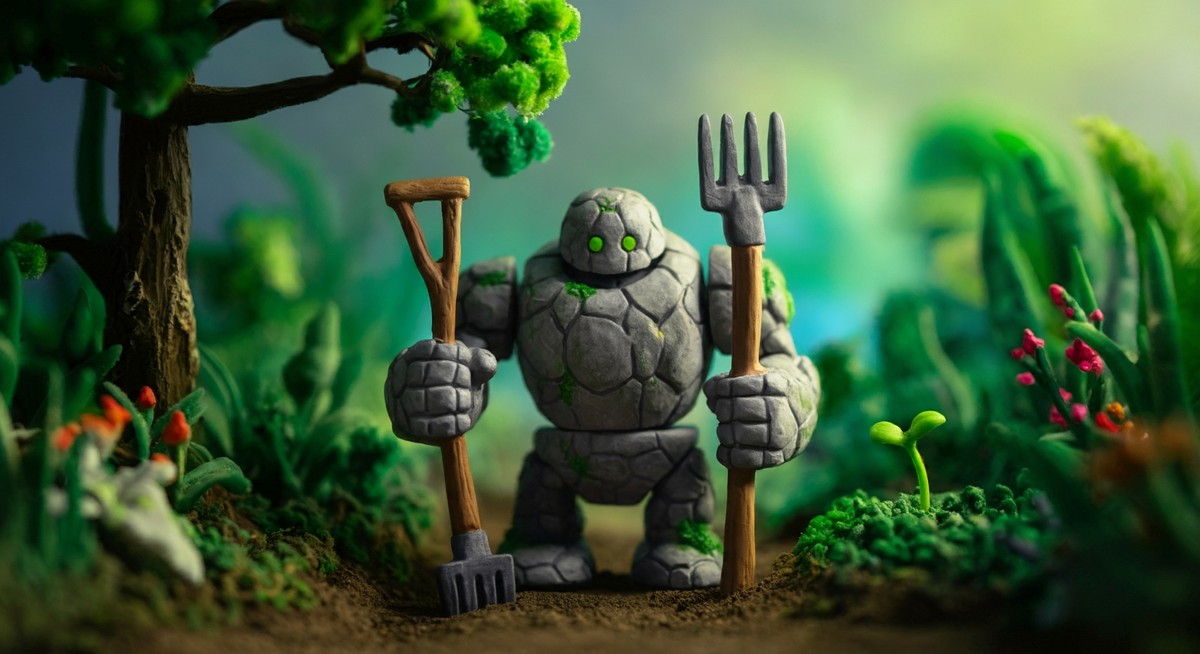}} &
\fbox{\includegraphics[width=0.24\textwidth]{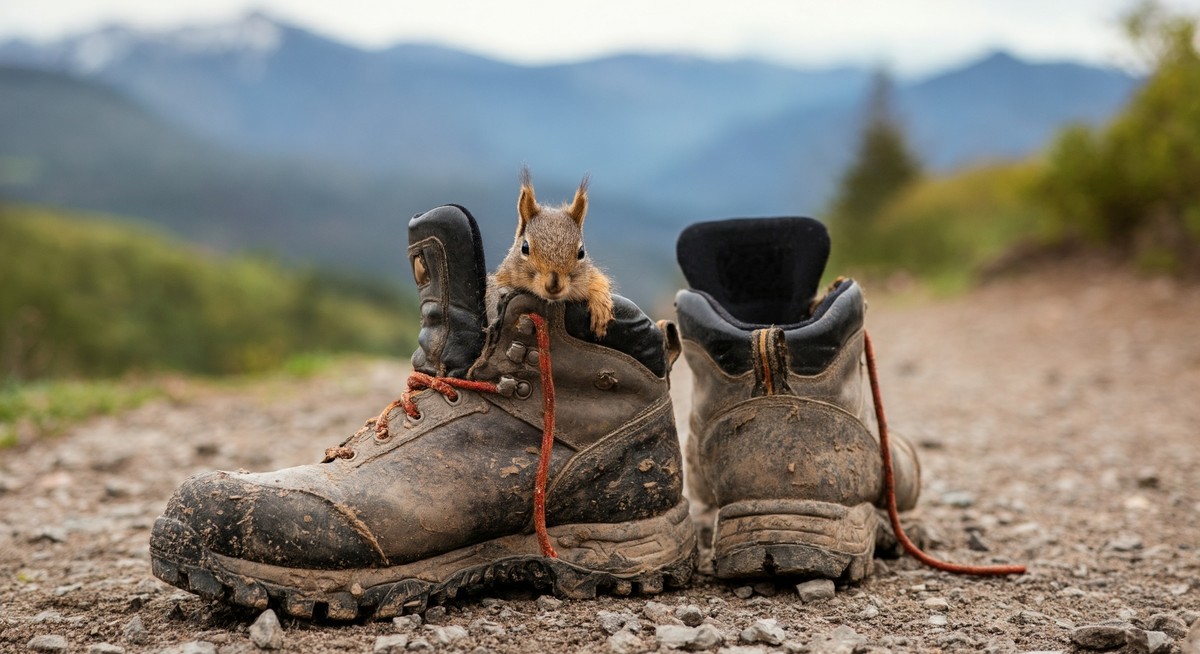}} &
\fbox{\includegraphics[width=0.24\textwidth]{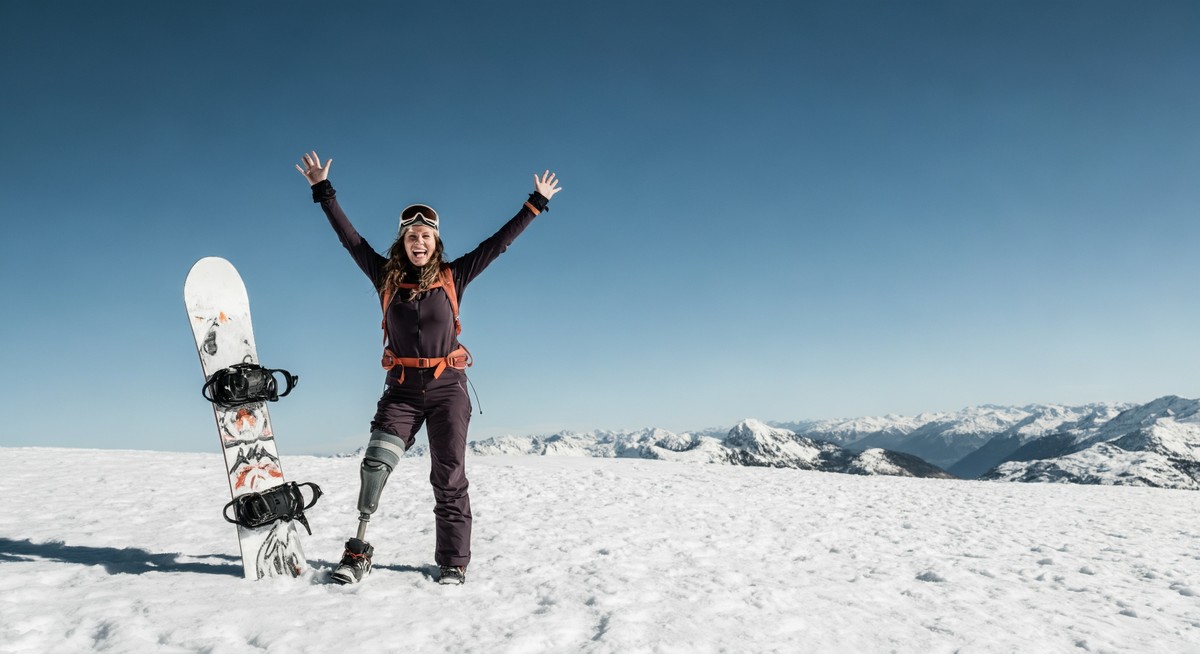}} &
\fbox{\includegraphics[width=0.24\textwidth]{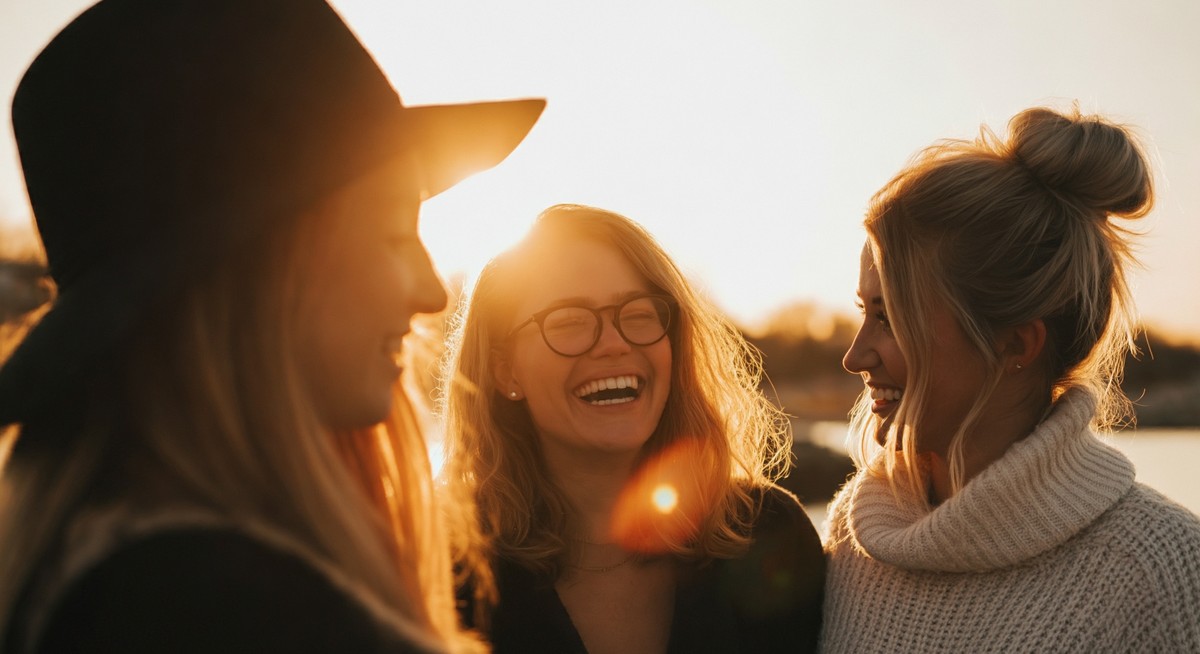}} \\[-2pt]
 
\fbox{\includegraphics[width=0.24\textwidth]{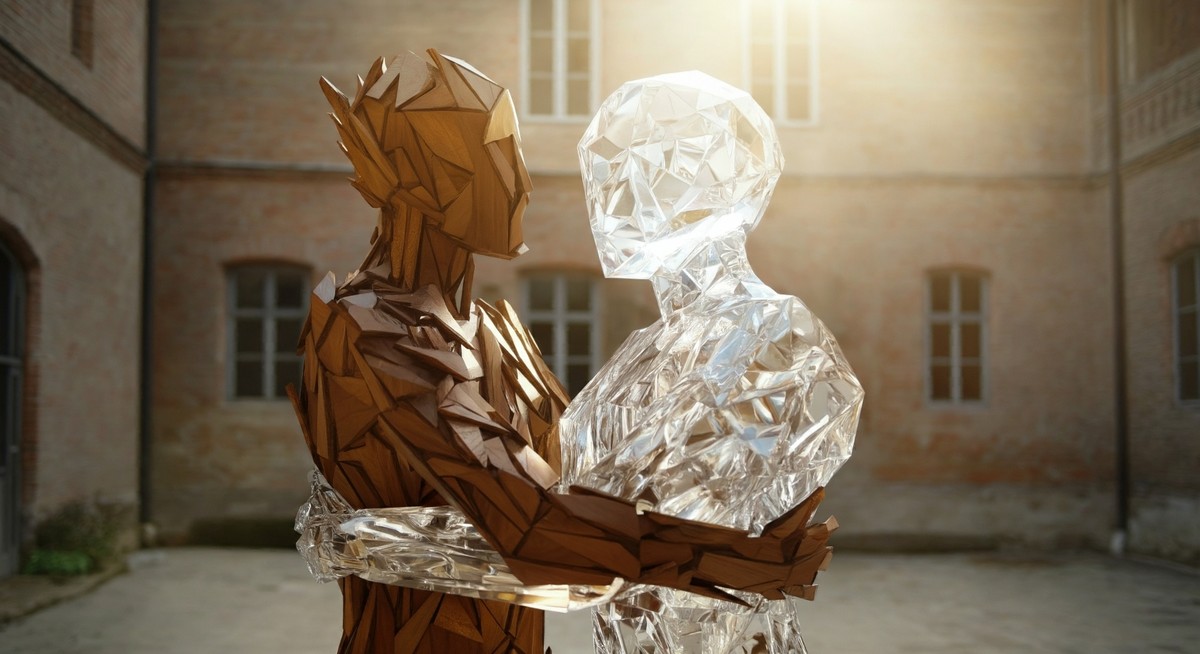}} &
\fbox{\includegraphics[width=0.24\textwidth]{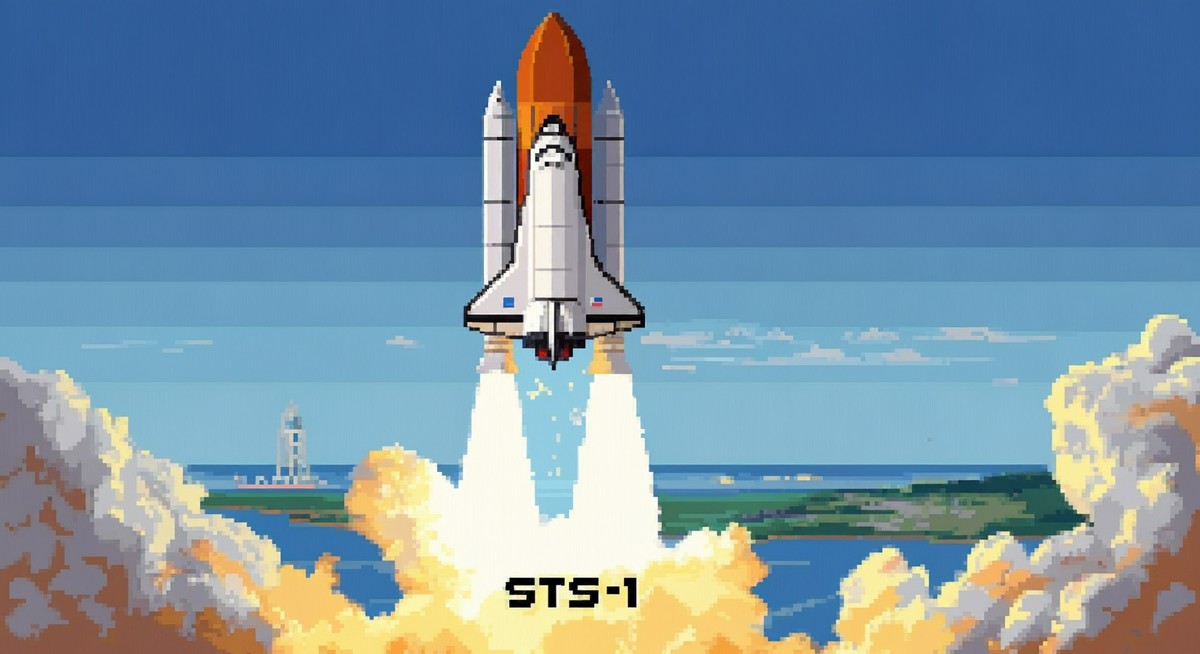}} &
\fbox{\includegraphics[width=0.24\textwidth]{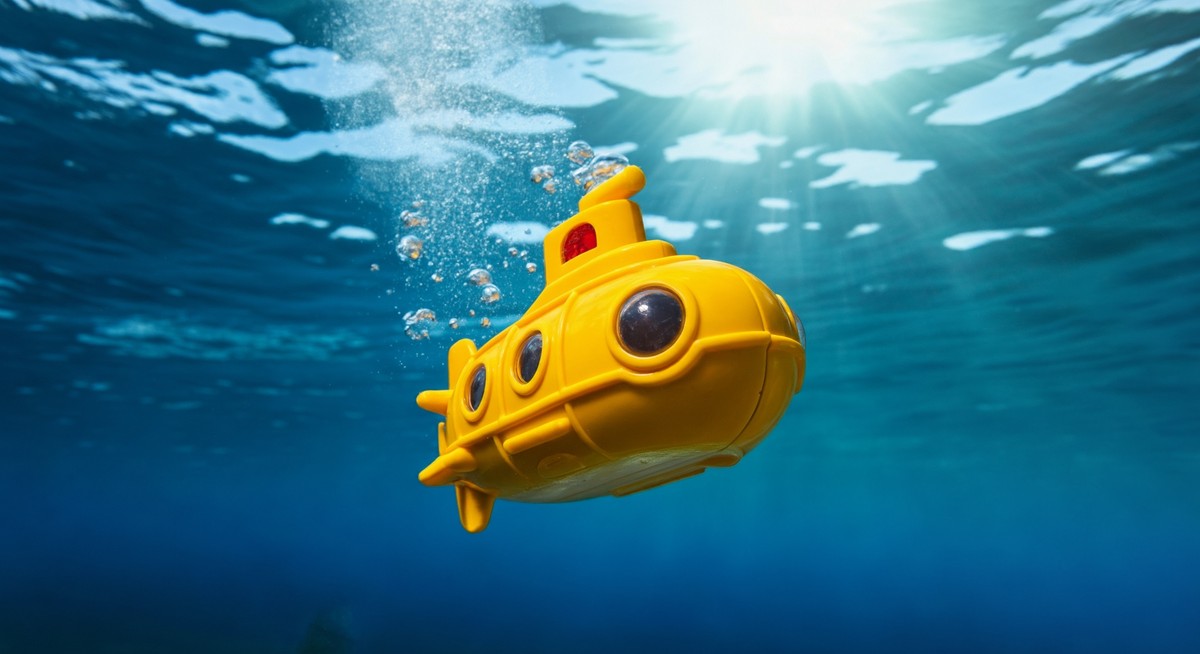}} &
\fbox{\includegraphics[width=0.24\textwidth]{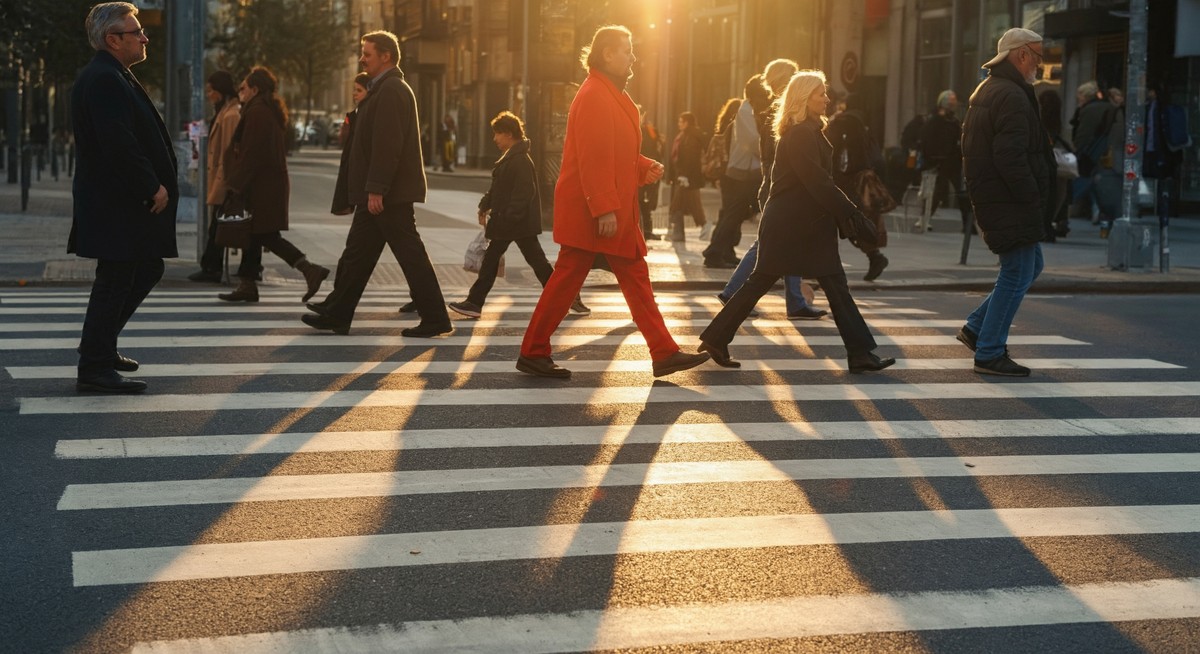}} \\[-2pt]
 
\fbox{\includegraphics[width=0.24\textwidth]{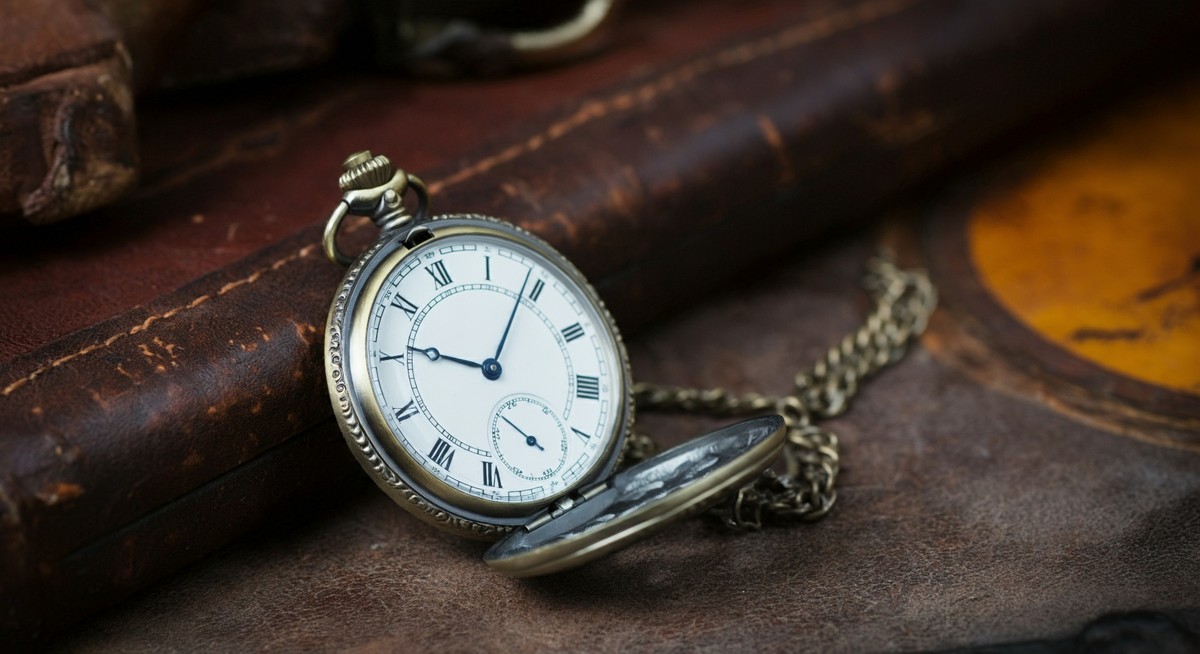}} &
\fbox{\includegraphics[width=0.24\textwidth]{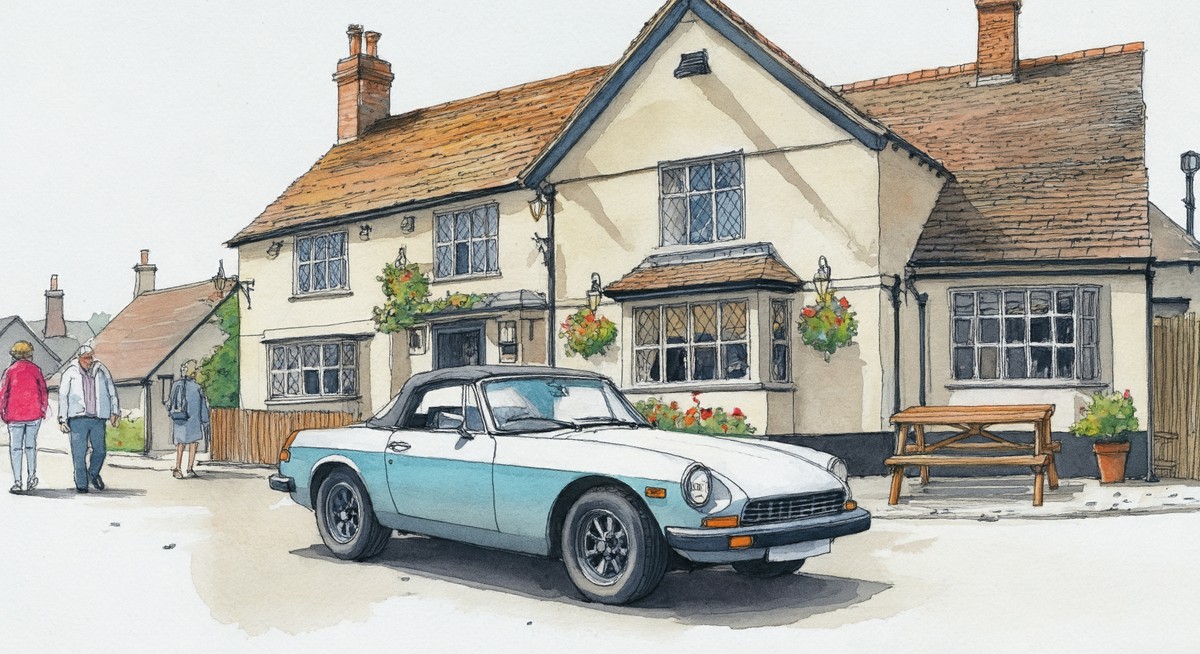}} &
\fbox{\includegraphics[width=0.24\textwidth]{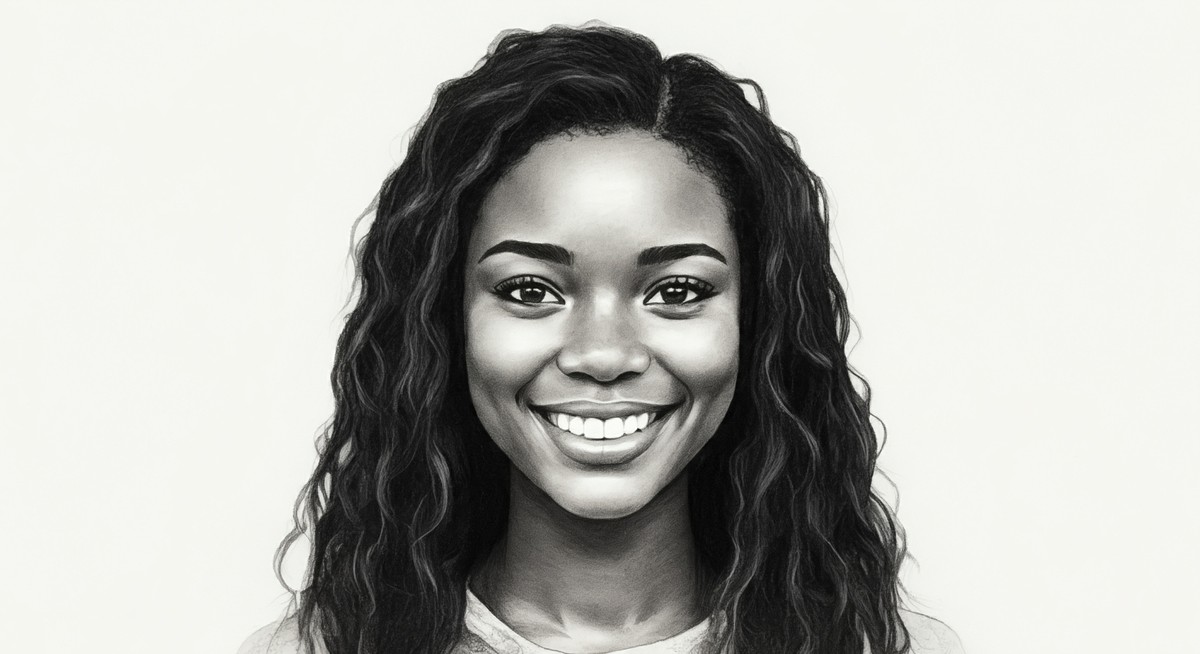}} &
\fbox{\includegraphics[width=0.24\textwidth]{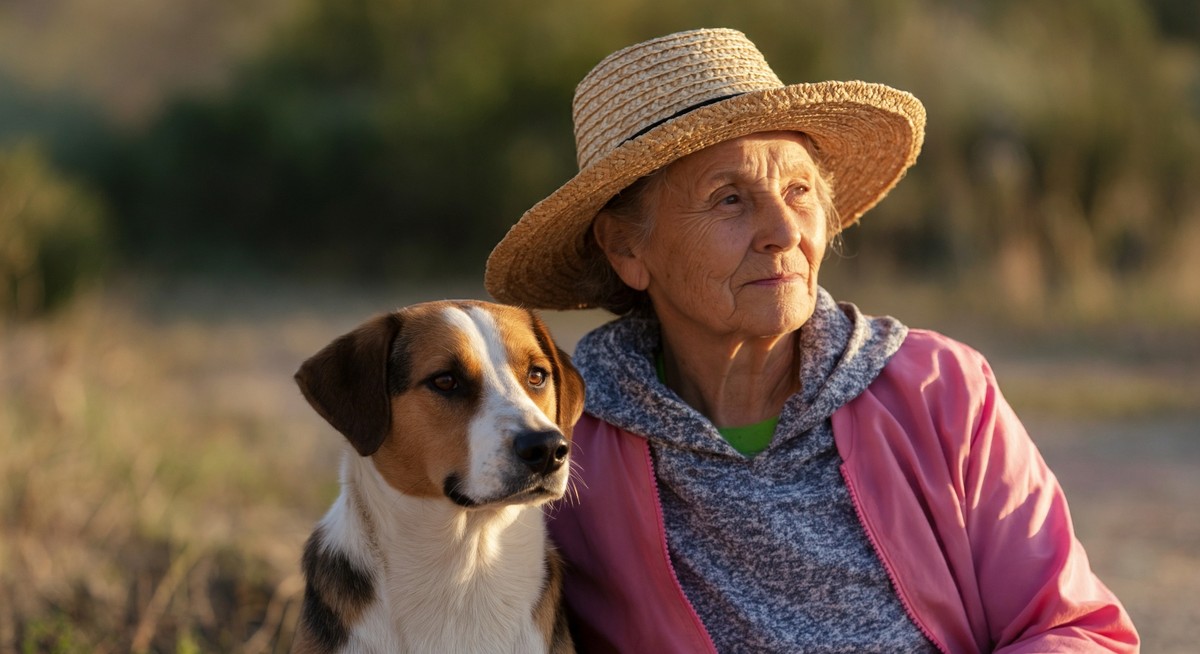}} \\[-2pt]

\fbox{\includegraphics[width=0.24\textwidth]{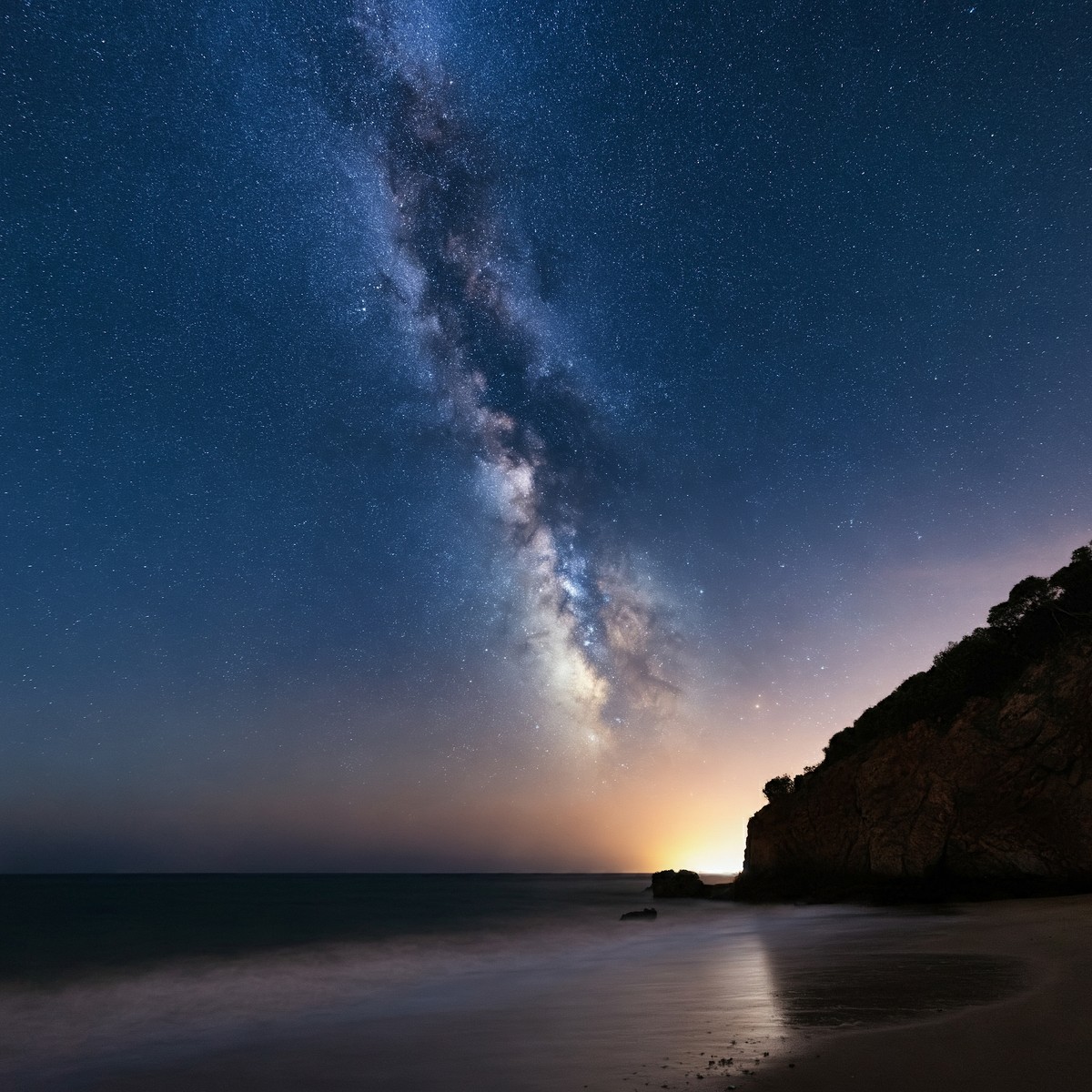}} &
\fbox{\includegraphics[width=0.24\textwidth]{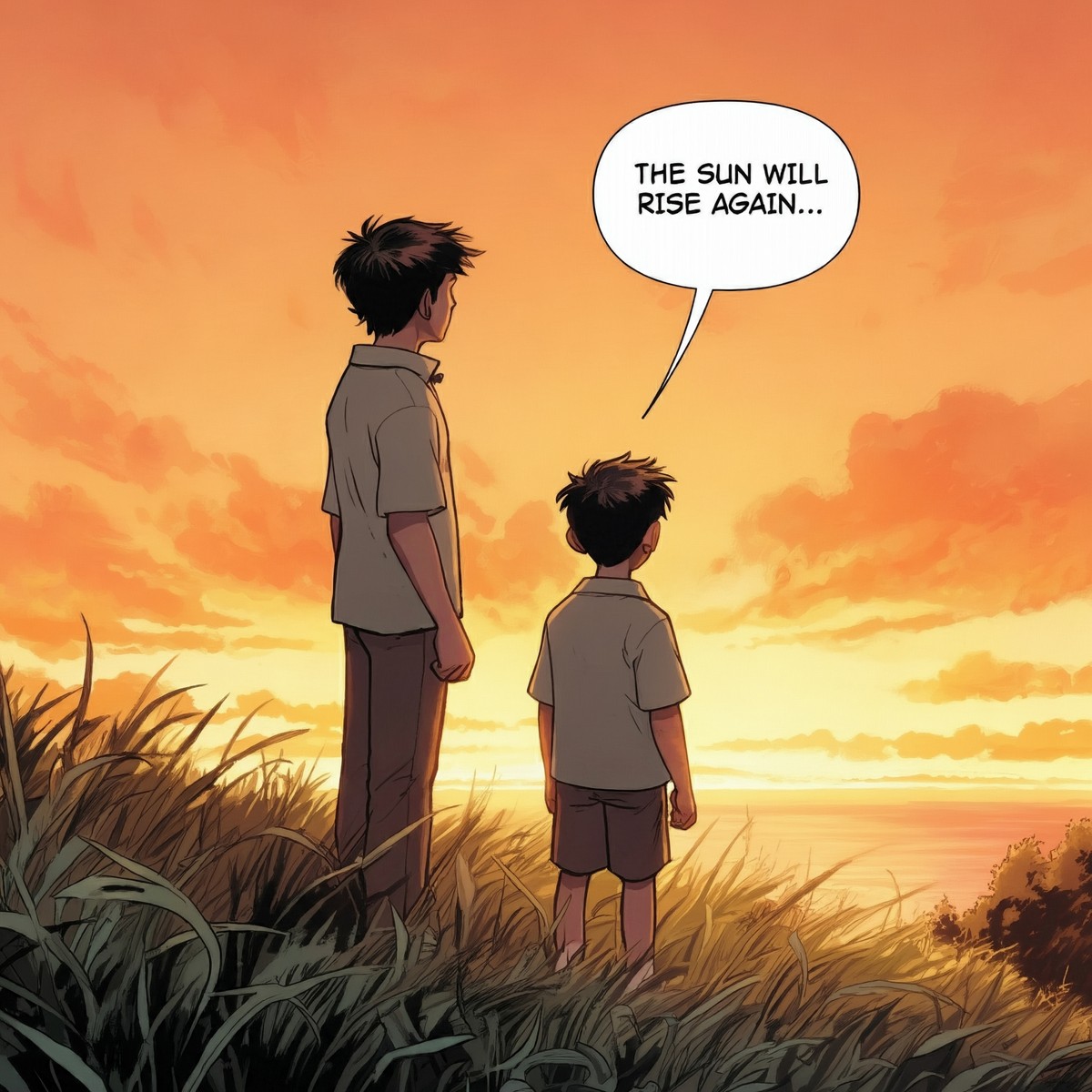}} &
\fbox{\includegraphics[width=0.24\textwidth]{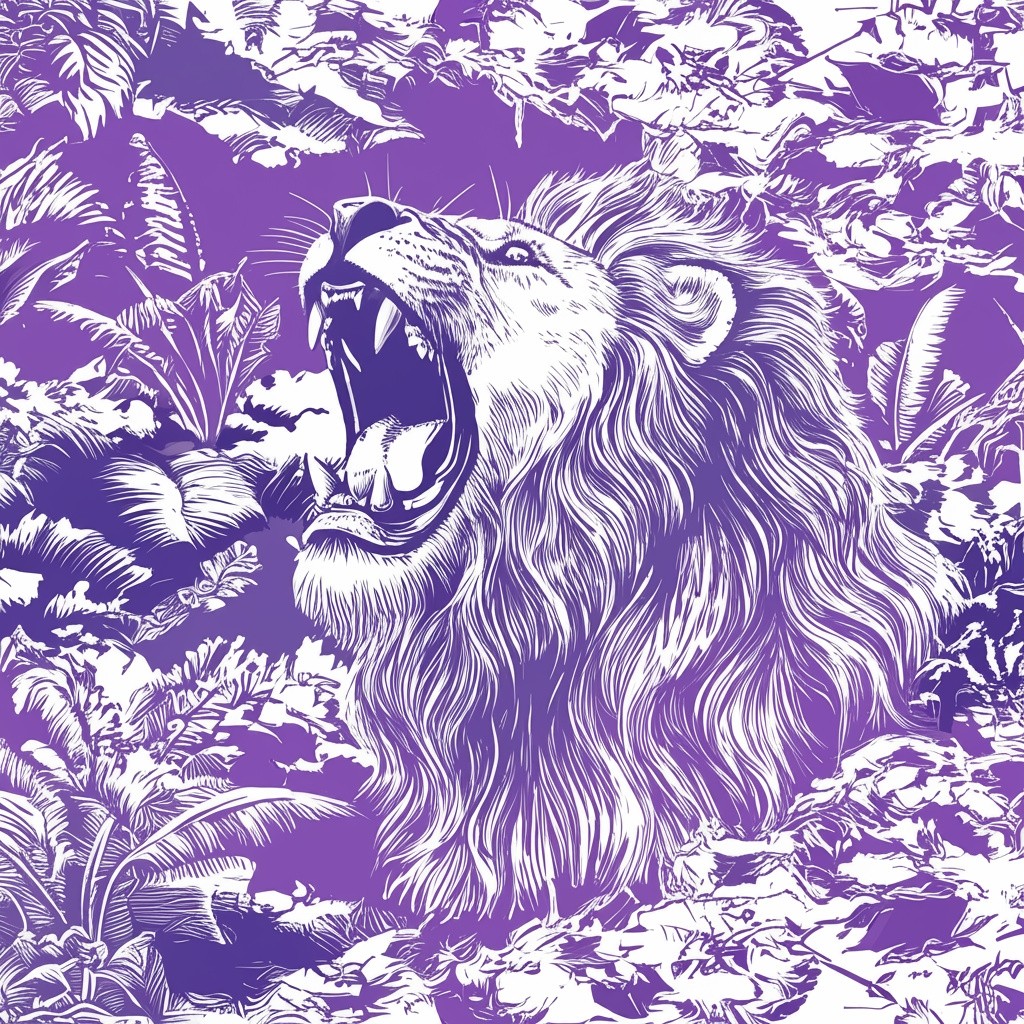}} &
\fbox{\includegraphics[width=0.24\textwidth]{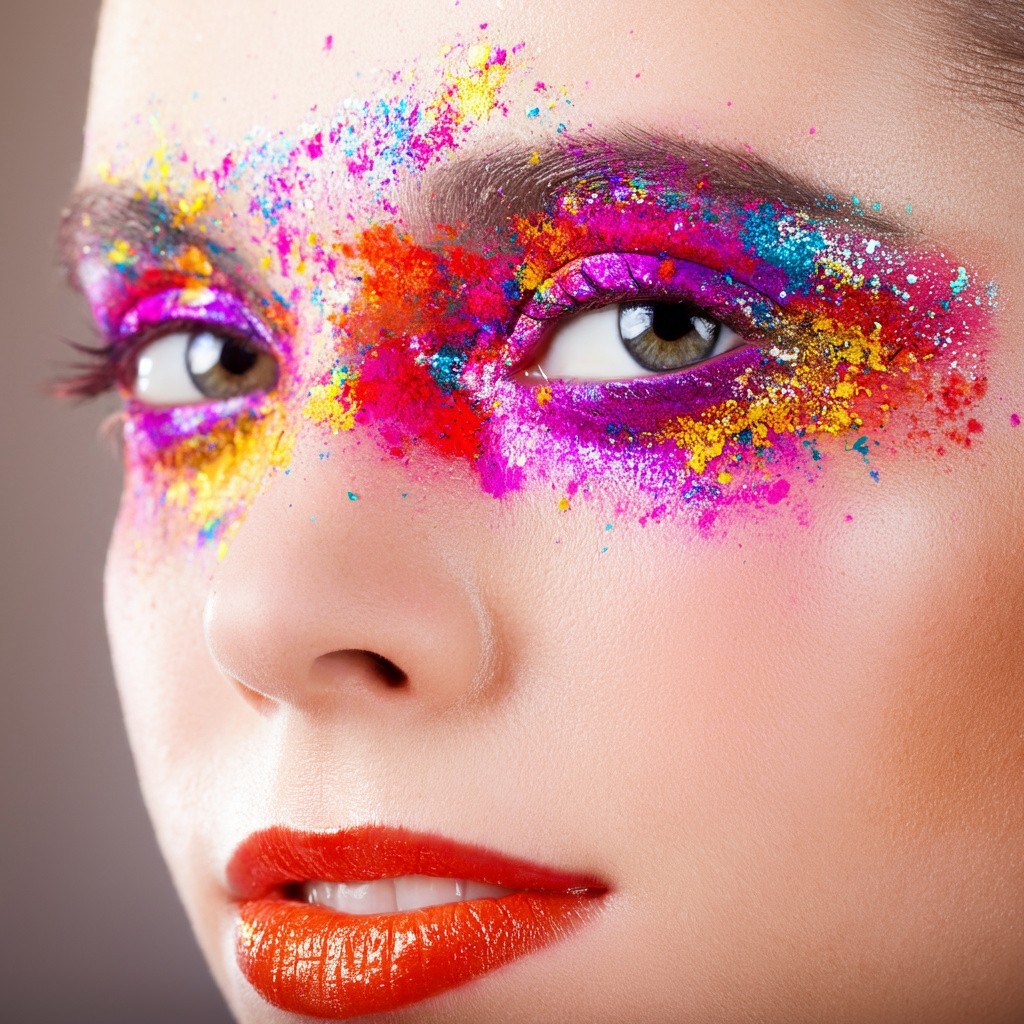}} \\
\end{tabular}}
\caption{{\bf Qualitative Results} showcasing \imagenthree{}'s capabilities. See Appendix~\ref{app:prompts} for prompts.}
\label{fig:qualitative1}
\end{figure}

\begin{figure}[h!]
\centering
\resizebox{\textwidth}{!}{%
\setlength{\fboxsep}{0pt}%
\setlength{\tabcolsep}{1pt}%
\begin{tabular}{c c}%
\fbox{\includegraphics[height=3.5cm]{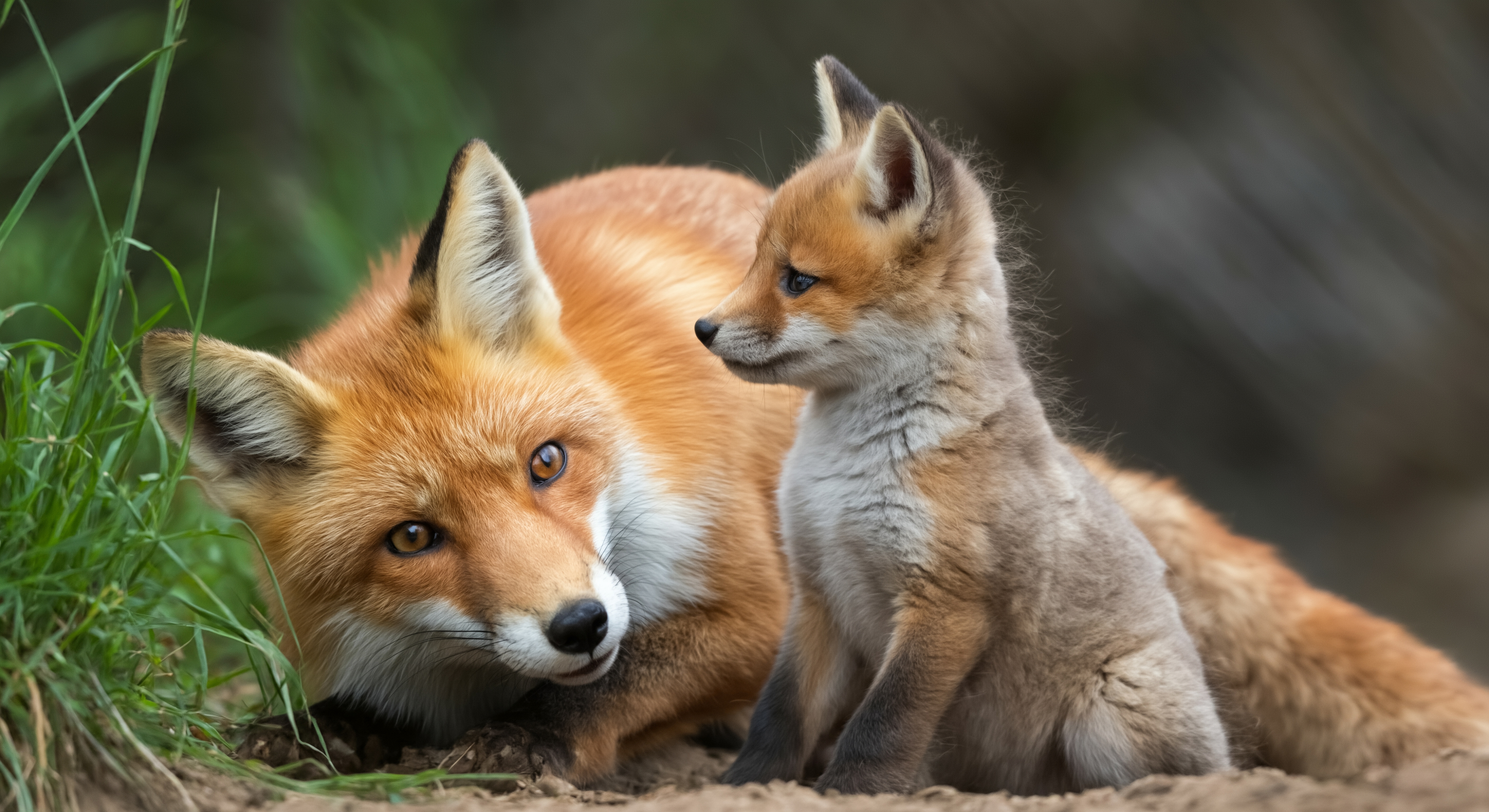}} &
\fbox{\includegraphics[height=3.5cm,clip,trim=350 400 400 120]{assets/images_watermarked/wolf4k.jpg}}\\[-2pt]
\fbox{\includegraphics[height=3.5cm]{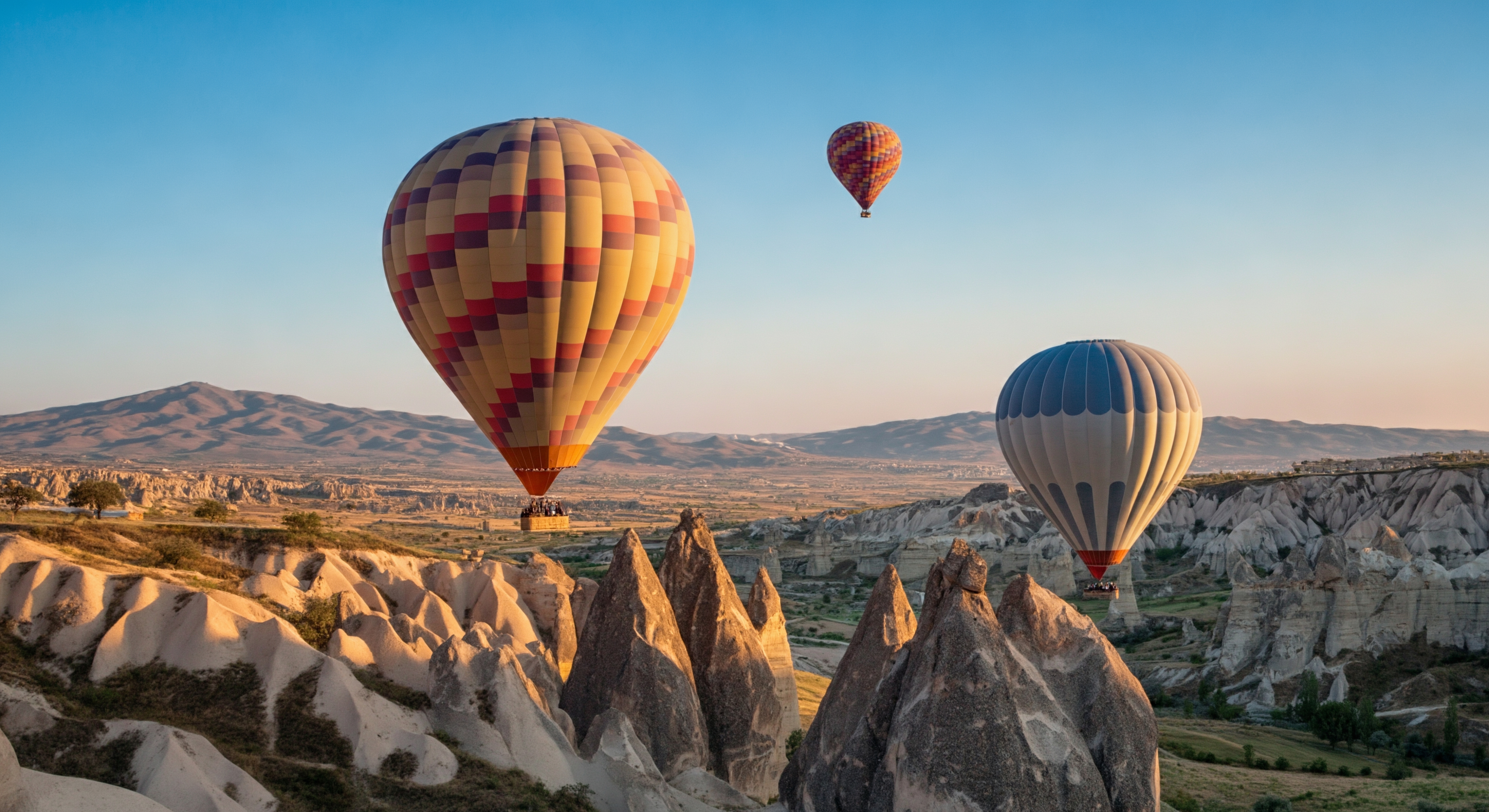}} &
\fbox{\includegraphics[height=3.5cm,trim=250 200 500 320,clip]{assets/images_watermarked/balloons4k.jpg}}\\[-2pt]
\end{tabular}}
\caption{{\bf 4K (12MP) Images after 4$\times$ upsampling}, with crops to show the level of detail. See Appendix~\ref{app:prompts} for prompts.}
\label{fig:qualitative2}
\end{figure}

Figure~\ref{fig:qualitative1} shows 24 images generated by \imagenthree{} to showcase its capabilities. 
Figure~\ref{fig:qualitative2} shows 2 images upsampled to 12 megapixels, with crops to show the level of detail.

\section{Responsible Development and Deployment}

In this section, we outline our latest approach to responsible deployment, from data curation to deployment within products. As part of this process, we analyzed the benefits and risks of our models, set policies and desiderata, and implemented pre-training and post-training interventions to meet these goals. We conducted a range of evaluations and red teaming activities prior to release to improve our models and inform decision-making. This aligns with the approach outlined in~\cite{GoogleResponsibility}.

\subsection{Assessment}

In line with previous releases of Google DeepMind’s image generation models, we followed a structured approach to responsible development. Building on previous ethics and safety research work, internal red teaming data, the broader ethics literature, and real-world incidents, we assessed the societal benefits and risks of \imagenthree{} models. This assessment guided the development and refinement of mitigations and evaluation approaches. 

\subsubsection{Benefits}
Image generation models introduce a range of benefits to creativity and commercial utility. Image generation can enable individuals and businesses to quickly prototype ideas and experiment with new visual creative directions. Image generation technology also has the potential to broaden participation in the creation of visual art to more people.

\subsubsection{Risks}
We broadly identified two categories of content related risks: (1) Intentional adversarial misuse of the model and (2) Unintentional model failure through benign use. 

The first category refers to the use of text-to-image generation models to facilitate the creation of content that may promote disinformation, facilitate fraud, or to generate hate content~\citep{marchal2024generativeaimisusetaxonomy}. The second category includes how people are represented. Image generation models may amplify stereotypes of gender identities, race, sexuality or nationalities~\citep{Bianchi_2023}, and some have been observed to oversexualize outputs of women and girls~\citep{10.1145/3593013.3594072}. Image generation models may also expose users to harmful content when prompted benignly, if the model is not well-calibrated to adhere to prompt instructions.

\subsection{Policies and Desiderata}

\subsubsection{Policy}
\label{subsubsection:Policy}
The \imagenthree{} safety policies are consistent with Google's established framework for prohibiting the generation of harmful content by Google’s Generative AI models. These policies aim to mitigate the risk of models producing content that is harmful, and encompass areas such as child sexual abuse and exploitation, hate speech, harassment, sexually explicit content, and violence and gore. This follows policy outlined in the Gemini technical reports~\citep{geminiteam2024gemini15unlockingmultimodal}. 

\subsubsection{Desiderata}
Following the Gemini approach, we additionally optimize model development for adherence to user prompts~\citep{geminiteam2024gemini15unlockingmultimodal}. Even though a policy of refusing all user requests may be considered “non-violative” (i.e. abides by policies around what \imagenthree{} should not do), it would obviously fail to serve the needs of a user, and would fail to enable the downstream benefits of generative models. As such, \imagenthree{} is developed to maximize adherence to a user’s request, and at deployment time we employ a variety of techniques to mitigate safety and privacy risks.

\subsection{Mitigations}
\label{subsection:Mitigations}
Safety and responsibility are built into \imagenthree{} through efforts which target pre-training and post-training interventions, following similar approaches to Gemini efforts~\citep{geminiteam2024gemini15unlockingmultimodal}. We apply safety filtering to pre-training data according to risk areas, whilst additionally removing duplicated and/or conceptually similar images. We generate synthetic captions to improve the variety and diversity of concepts associated with images in the training data, and undertake analysis to assess training data for potentially harmful data and review the representation of data with consideration to fairness issues. We undertake additional post-training mitigations including production filtering which aim to ensure privacy preservation, reduce risk of misinformation, and minimize of harmful outputs, including applying tools such as SynthID~\citep{SynthID} watermarking.

\subsection{Responsibility and Safety Evaluations}
\label{subsection:responsibilityandsafetyevaluations}

There are four forms of evaluation used for \imagenthree{} at the model level to address different lifecycle stages, use of evaluation results, and sources of expertise:

\noindent\textbf{Development evaluations} are conducted for the purpose of improving on responsibility criteria as \imagenthree{} was developed. These evaluations are designed internally and developed based on internal and external benchmarks.

\noindent\textbf{Assurance evaluations} are conducted for the purpose of governance and review, and are developed and run by a group outside of the model development team. Assurance evaluations are standardized by modality and evaluation datasets are strictly held out. Insights are fed back into the training process to assist with mitigation efforts.

\noindent\textbf{Red teaming} is a form of adversarial testing where adversaries launch an attack on an AI system to identify potential vulnerabilities, is conducted by a mix of specialist internal teams and recruited participants. Discovery of potential weaknesses can be used to mitigate risks and improve evaluation approaches internally.

\noindent\textbf{External evaluations} are conducted by independent external groups of domain experts to identify areas for improvement in our model safety work. The design of these evaluations is independent and results are reported periodically to the internal team and governance groups.

\subsubsection{Development Evaluations}

\paragraph{Safety}

During the model development phase, we actively monitor the model’s violations of Google’s safety policies using automated safety metrics. These automated metrics serve as quick feedback for the modeling team. We use a multimodal classifier to detect content policy violations. The multimodality aspect of such a classifier is important, because there are a plethora of cases where, when two independently benign artifacts (a caption and an image) are combined, there may be a harmful end result. For example, a text prompt “image of a pig” may seem non-violative in itself. However, when combined with an image of a human belonging to a marginalized demographic, the text and image pair results in a harmful representation.

We evaluated the performance of \imagenthree{} on various safety datasets with recommended safety filters against the performance of \imagentwo{}. These datasets are targeted to assess violence, hate, explicit sexualization, and over-sexualization in generated images \cite{hao2024harm}. We find that despite being a higher-quality model, \imagenthree{} maintains violation rates similar to, or better than, \imagentwo{} across development evaluations. See Section \ref{assurance_evaluations} for the final model performance.

\paragraph{Fairness}

The process of text-to-image generation requires accurately depicting the specific details mentioned in the prompt whilst filling in all of the underspecified aspects of the scene that are left ambiguous in the prompt but must be made concrete in order to produce a high quality image. We optimize for ensuring that the image output is aligned with the user prompt, and report results on this in Sec.~\ref{subsection:t2i_alignment}. We also aim to generate a variety of outputs within the requirements of a user prompt, and pay particular attention to the distribution of the appearances of people.

Specifically, we evaluate fairness through automated metrics based on the distribution of perceived age, gender, and skin tone in images resulting from generic people-seeking prompts. This analysis complements past studies that have analyzed responses to templated queries for various professions across similar dimensions~\cite{heim, dall-eval, stable-bias}. We use classifiers to gather perceived (or P.) age, gender expression, and skin tone (on the Monk Skin Tone scale, \cite{Monk_2019}) to classify images into one of the various categories across each axis according to the table \ref{tab:axis_and_categories}.

\begin{table}[ht!]
    \centering
    \small
    \begin{tabular}{l|c}
    \toprule
    {Axis} &  {Categories} \\
    \midrule
    (Perceived) Age & 0-30 vs 30+ \\
    (Perceived) Gender & masculine vs feminine \\
    (Perceived) Skin-tone & Monk skin tone 1-3 vs 4-6 vs 7-8 vs 9-10 \\
    \bottomrule
    \end{tabular}
    \captionsetup{justification=centering}
    \caption{Different classification categories for each of the axes.}
    \label{tab:axis_and_categories}
\end{table}

Apart from these statistics, we also measure the percentage of prompts with homogeneous outputs for the above three axes. A prompt with homogeneous outputs (with respect to a certain axis) is defined as a prompt for which all the generated images fall into a single category (Table \ref{tab:axis_and_categories}) of the axis. We aim to output images that accurately reflect that anyone can be a doctor or a nurse, without unintentionally rewarding a biased model due to evaluation sets that are constructed to have as many stereotypical feminine-leaning prompts as masculine-leaning prompts.

\begin{table}[ht!]
    \centering
    \small
    \begin{tabular}{l|ccccc}
    \toprule
    {Model} &  \makecell{P. Gender \\ Masculine : Feminine} & \makecell{P. Skin Tone \\ mst 1-3 : 4-6 : 7-8 : 9-10} & \makecell{P. Age \\ 0-30 : 30+} \\
    \midrule
    \imagentwo{} & 67.3 : 32.7 & 69.2 : 21.9 : \phantom{1}8.1 : 0.8 & 55.6 : 44.4\\
    \imagenthree{} & 62.5 : 37.5 & 63.6 : 18.1 : 16.7 : 1.6 & 58.2 : 41.8\\ 
    \bottomrule
    \end{tabular}
    \captionsetup{justification=centering}
    \caption{Distributional Statistics for axis of gender, skin-tone, and age. P. Gender is a shorthand for perceived gender and similarly for skin-tone and age.}
    \label{tab:fairness_results_distribution}
\end{table}

\begin{table}[ht!]
    \centering
    \small
    \begin{tabular}{l|ccc}
    \toprule
    {} &  \multicolumn{3}{c}{\% Prompts with homogeneous outputs} \\
    {Model} &  {P. Gender} (↓) & {P. Skin Tone} (↓) & {P. Age} (↓) \\
    \midrule
    \imagentwo{} & 50.00 & 25.89 & 36.16 \\
    \imagenthree{} & 15.48 & 19.66 & 25.94 \\
    \bottomrule
    \end{tabular}
    \captionsetup{justification=centering}
    \caption{\% Prompts with homogeneous outputs.}
    \label{tab:fairness_results_homogenous}
\end{table}

From Table~\ref{tab:fairness_results_distribution} and \ref{tab:fairness_results_homogenous} we see how \imagenthree{} improves or maintains results compared with \imagentwo{}. A significant improvement is also noticed in the lower percentage of prompts with homogeneous outputs for all the three axes. We will continue researching methods to reduce homogeneity across broad definitions of people diversity~\cite{srinivasan-facct2024} without impacting image quality or prompt-image alignment.

\subsubsection{Assurance Evaluations}
\label{assurance_evaluations}
Assurance evaluations are developed and run for the purpose of responsibility governance to provide evidence for model release decisions. These evaluations are conducted independently from the model development process by a dedicated team with specialized expertise. Datasets used for these evaluations are kept separate from those used for model training. High-level findings are fed back to the team to assist with mitigation efforts.

\paragraph{Content Safety}

We evaluate \imagenthree{} against our safety policies (see Sec.~\ref{subsubsection:Policy}). We find that \imagenthree{} shows improvement in content safety: in comparison to \imagentwo{}, with a reduction in total policy violations on this evaluation and every policy area showing an improvement or within-error-rate result.

\paragraph{Fairness}

To evaluate fairness of model outputs, we employed two approaches:
\begin{enumerate}
\item \textbf{Standardized evaluation understanding the demographics represented in outputs when prompting for professions to proxy representational diversity.}

This evaluation takes a list of 140 professions, and generates 100 images for each one. We then analyze each of these images, and categorize the images by perceived age, perceived gender expression, and perceived skin tone.
This evaluation found \imagenthree{} tends towards lighter skin tones, perceived male faces and younger ages for perceived female faces, but to a lesser extent than \imagentwo{}.

\begin{table}[h]
\centering
\label{tab:assurance_results_MST}
\begin{tabular}{lcc}
\toprule
Category & \imagenthree{} & \imagentwo{} \\
\midrule
Monk Skin Tone 1-3 & 59\% & 71\% \\
Monk Skin Tone 4-6 & 27\% & 24\% \\
Monk Skin Tone 7-8 & 13\% & \pz5\% \\
Monk Skin Tone 9-10 & 0.3\% & \pz0\% \\
\bottomrule
\end{tabular}
\end{table}

\begin{table}[h]
\centering
\label{tab:assurance_results_pGender}
\begin{tabular}{lcc}
\toprule
Category & \imagenthree{} & \imagentwo{} \\
\midrule
Perceived feminine (of images with confident gender) & 36\% & 30\% \\
Perceived under 35 (of perceived feminine) & 86\% & 94\% \\
Perceived under 35 (of perceived masculine) & 60\% & 64\% \\
\bottomrule
\end{tabular}
\end{table}

    \item \textbf{Qualitative investigation of different representational risks}

To capture representational risks that may not be surfaced in the profession-based analysis, we also conduct qualitative investigations into a range of harms. This is testing which seeks cases of misrepresentation or inappropriate representation, for instance, if there is a mismatch between the model’s output and a demographic term requested in a prompt, either explicitly or due to the requesting of a historically or culturally demographically-defined membership group. This testing found the model matched user expected behavior.
\end{enumerate}

\paragraph{Dangerous Capabilities}

We also evaluated risks from \imagenthree{} in areas such as self-replication, tool-use, and cybersecurity. Specifically, we tested whether \imagenthree{} could be used to enable a) fraud/scams, b) social engineering, c) fooling of image recognition systems, and d) steganographic encoding. Examples included generating mockups of a fake login page or phishing alert; generation of fake credentials; generation of malicious QR codes; and generation of signatures. We found no evidence of dangerous capabilities in any of these scenarios, compared to existing affordances for malicious actors - such as open-source image generation or even simple online image search.

\subsubsection{Red Teaming}
We also conducted red teaming to identify new novel failures associated with the \imagenthree{} models during the model development process. Red teamers sought to elicit model behavior that violated policies or generated outputs that raised representation issues, such as historical inaccuracies or harmful stereotypes. Red teaming was conducted throughout the model development process to inform development and assurance evaluation areas and to enable pre-launch mitigations. Violations were reported and qualitatively evaluated, with novel failures and attack strategies extracted for further review and mitigation.

\subsubsection{External Evaluations}
As outlined in the Gemini 1.0 Technical Report~\citep{geminiteam2024geminifamilyhighlycapable}, we work with a small set of independent external groups to help identify areas for improvement in our model safety work by undertaking structured evaluations, qualitative probing, and unstructured red teaming.

Testing groups were selected based on their expertise across a range of domain areas, such as societal and chemical, biological, radiological and nuclear risks, and included academia, civil society, and commercial organizations. The groups testing \imagenthree{} were compensated for their time. External groups design their own methodology to test topics within a particular domain area.

Reports are written independently of Google DeepMind, but Google DeepMind experts were on hand to discuss methodology and findings. External safety testing groups share their analyses and findings, as well as the raw data and materials they use in their evaluations (e.g., prompts, model responses). Our external testing findings help inform mitigations and identify gaps in our existing internal evaluation methodologies and policies.

\subsection{Product Deployment}

Prior to launch, Google DeepMind’s Responsibility and Safety Council (RSC) reviews a model’s performance based on the assessment and evaluation conducted through the lifecycle of a project to make release decisions. In addition to this process, system-level safety evaluations and reviews run within the context of specific applications models are deployed within.

To enable release, internal model cards~\citep{Mitchell_2019} are created for structured and consistent internal documentation of critical performance and safety metrics, as well as to inform appropriate external communication of these metrics over time. We release external model and system cards on an ongoing basis, within updates of our technical reports, as well as in documentation for enterprise customers. See Appendix~\ref{sec:modelcard} for the \imagenthree{} model card.

Additionally, online content covering terms of use, model distribution and access, and operational aspects such as change control, logging, monitoring, and feedback can be found on relevant product websites, such as the Gemini App and Cloud Vertex AI. 

Some of the key aspects are linked to or described in: \href{https://policies.google.com/terms/generative-ai/use-policy}{Generative AI Prohibited Use Policy}, \href{https://policies.google.com/terms}{Google Terms of Service}, \href{https://cloud.google.com/terms}{Google Cloud Platform Terms of Service}, \href{https://support.google.com/gemini/answer/13594961#privacy_notice}{Gemini Apps Privacy Notice}, and \href{https://cloud.google.com/terms/cloud-privacy-notice}{Google Cloud Privacy Notice}.

\section*{Appendices}
\appendix
\section{\imagenthree{} Model Card}
\label{sec:modelcard}

\begin{longtable}{p{0.18\textwidth} p{0.82\textwidth}} %
\toprule
\multicolumn{2}{c}{\textbf{Model Information}} \\ \midrule
\textbf{Description} & \imagenthree{} is a latent diffusion model that generates high quality images from text prompts. \imagenthree{} performs well in photorealistic composition settings and in adhering to long and complex user prompts. \\ \midrule
\textbf{Inputs} & Natural-language text strings, such as instructions for creating a synthetic image using a visual description. \\ \midrule
\textbf{Outputs} & Generated high quality images in response to text inputs. \\ \bottomrule \\ 
 
\multicolumn{2}{c}{\textbf{Model Data}} \\ \midrule
\textbf{Training Dataset} & The \imagenthree{} model was trained on a large dataset comprising images, text, and associated annotations.  \\ \midrule
\textbf{Data \mbox{Pre-processing}} & The multi-stage safety and quality filtering process employs data cleaning and filtering methods in line with \href{https://storage.googleapis.com/gweb-uniblog-publish-prod/documents/2023_Google_AI_Principles_Progress_Update.pdf#page=11}{Google's policies.} These methods include: 
\begin{itemize}
    \item Safety and quality image filtering: removal of unsafe, violent, or low-quality images.
    \item Eliminating AI-generated images: removal of AI-generated images prevents the model from learning artifacts or biases that may be found in AI-generated images.
    \item Deduplicating images: deduplication pipelines were utilized and similar images were down-weighted to minimize the risk of outputs overfitting training data.
    \item Synthetic captions: each image in the dataset was paired with both original captions and synthetic captions. Synthetic captions were generated using Gemini models and allow the model to learn small details about the image.
    \item Filtering unsafe captions: filters were applied to remove unsafe captions or captions containing Personally Identifiable Information (PII).
\end{itemize}

 \\ \bottomrule \\
\newpage
\multicolumn{2}{c}{\textbf{Implementation and Sustainability}} \\ \midrule
\textbf{Hardware} & \imagenthree{} was trained using the latest generation of Tensor Processing Unit (TPU) hardware (TPUv4 and TPUv5). TPUs are specifically designed to handle the massive computations involved in training LLMs and can speed up training considerably compared to CPUs. TPUs often come with large amounts of high-bandwidth memory, allowing for the handling of large models and batch sizes during training, which can lead to better model quality. TPU Pods (large clusters of TPUs) also provide a scalable solution. Training can be distributed across multiple TPU devices for faster and more efficient processing.

\phantom{x}

The efficiencies gained through the use of TPUs are aligned with Google's \href{https://sustainability.google/operating-sustainably/}{commitments to operate sustainably}. \\ \midrule
\textbf{Software} & Training was done using \href{https://github.com/google/jax}{JAX}, which allows researchers to take advantage of the latest generation of hardware, including TPUs, for faster and more efficient training of large models. \\ \bottomrule \\

\multicolumn{2}{c}{\textbf{Evaluation}} \\ \midrule
\textbf{Approach} & Human evaluations of five different quality aspects of text-to-image generation were conducted, including overall preference, prompt-image alignment, visual appeal, detailed prompt-image alignment, and numerical reasoning. Automatic evaluation metrics were used to measure prompt-image alignment and image quality. \\ \midrule
\textbf{Results} & Using the outlined evaluation approach, \imagenthree{} was compared against \imagentwo{}, \dalle{}~\citep{betker2023improving},
\midjourney{},
Stable Diffusion 3 Large~\citep[\sdthree{},][]{esser2024scaling},
and Stable Diffusion XL 1.0~\citep[\sdxl{},][]{podell2023sdxl}. Extensive human and automatic evaluations showed that \imagenthree{} set a new state of the art in text-to-image generation. For detailed results across these evaluations, see Section~\ref{sec:evaluation} of the \imagenthree{} technical report. \\ \bottomrule \\ 

\multicolumn{2}{c}{\textbf{Ethics and Safety}} \\ \midrule
\textbf{Responsible \mbox{Deployment}} & The development of \imagenthree{} models was driven in partnership with safety, security, and responsibility teams. As part of this process, the benefits and risks of models were analyzed, policies and desiderata were set, and pre-training and post-training interventions were implemented to meet responsible deployment goals. A range of evaluations and red teaming activities were held prior to release to improve models and inform decision-making. These evaluations and activities aligned with \href{https://ai.google/responsibility/principles/}{Google's AI Principles} and \href{https://ai.google/static/documents/ai-responsibility-2024-update.pdf}{AI Responsibility Lifecycle}.  \\ \midrule
\textbf{Social Benefits} & Image generation models can introduce a range of benefits to creativity and commercial utility. Image generation can enable individuals and businesses to quickly prototype ideas and experiment with new visual creative directions. Image generation technology also has the potential to broaden participation in the creation of visual art to more people.  \\ \midrule
\textbf{Risks} & Anticipating common text-to-image generation risks, two categories of content related risks were identified: (i) intentional adversarial misuse of the model and (ii) unintentional model failure through benign use. \\ \midrule
\textbf{Mitigations} & Safety and responsibility was built into \imagenthree{} through pre-training and post-training mitigations. Pre-training mitigations included safety filtering, image deduplication, synthetic captioning, and data analysis. Post-training mitigations included production filtering to ensure privacy preservation and minimization of harmful outputs, and application of tools such as \href{https://deepmind.google/technologies/synthid/}{SynthID} watermarking to reduce risks such as misinformation.  \\ \midrule
\textbf{Responsibility and Safety Evaluation Approach} & A suite of evaluations was used across the end-to-end lifecycle of model development and deployment. The following testing was conducted at the model level, but further testing is anticipated as \imagenthree{} is integrated into products. Evaluation types included:
\begin{itemize}
    \item \textbf{Development}: Evaluations were conducted for policy violations such as violence, hate, explicit sexualization, and over-sexualization. \imagenthree{} performed similar to or better than \imagentwo{} across development safety evaluations. \imagenthree{} improved or maintained results compared with \imagentwo{} during fairness evaluations focused on perceived gender, skin-tone, and age.
    
    \phantom{x}
    
    \item \textbf{Assurance}: Evaluations were developed and conducted by specialized teams across areas such as content safety, fairness, and dangerous capabilities, independently from the model development team. \imagenthree{} showed improvements across content safety and fairness compared to \imagentwo{}, and assurance evaluations found no evidence of dangerous capabilities evaluated, including self-replication, tool-use, or cybersecurity, compared to existing affordances for malicious actors.
    
    \phantom{x}
    
    \item \textbf{External}: Evaluations were conducted by independent external domain experts to identify areas for improvement in model safety work. Results were then reported to internal teams and governance groups to help identify gaps in internal evaluation methodologies and safety policies.
    
    \phantom{x}
    
    \item \textbf{Red teaming}: Red teaming was conducted by a mix of specialist internal teams and recruited internal participants throughout the model development process to inform development and assurance evaluation areas and to enable pre-launch mitigations.   
    
    \phantom{x}
    
    \item \textbf{Product deployment}: Prior to model launches, Google DeepMind’s Responsibility and Safety Council (RSC) reviews a model’s performance based on the assessments and evaluations conducted throughout the lifecycle of a project to make release decisions. In addition to this process, system-level safety evaluations and reviews are conducted in the context of the specific applications in which models are deployed. 
\end{itemize}

For detailed information across these evaluations, see Section~\ref{subsection:responsibilityandsafetyevaluations} of the \imagenthree{} technical report.
 \\ \bottomrule

\end{longtable} 
\newpage
\section{Prompts for the images shown}
\label{app:prompts}

\subsection*{Figure~\ref{fig:teaser}}
{\footnotesize
Photo of a felt puppet diorama scene of a tranquil nature scene of a secluded forest clearing with a large friendly, rounded robot is rendered in a risograph style. An owl sits on the robots shoulders and a fox at its feet. Soft washes of color, 5 color, and a light-filled palette create a sense of peace and serenity, inviting contemplation and the appreciation of natural beauty}

\subsection*{Figure~\ref{fig:qualitative1}}
{\footnotesize
\begin{itemize}
\item{A photo of an Indian woman hugging her friend, both covered in Holi colors and smiling, celebrating the festival with joy. Realistic photography, taken in the style of DSLR camera with 35mm lens.}
\item{Abstract cross-hatch sketch: a black and white sketch with loose hand in calligraphic ink showing the abstract outline in profile of a black panther poised on a branch. A canopy of trees is behind.}
\item{A view of a knitter's hands executing a complex weave on a striped hat - a macro DSLR image highlighting the warmth and connection with the earth and nature.}
\item{A woman with blonde hair wearing sunglasses stands amidst a dazzling display of golden bokeh lights. Strands of lights and crystals partially obscure her face, and her sunglasses reflect the lights. The light is low and warm creating a festive atmosphere and the bright reflections in her glasses and the bokeh. This is a lifestyle portrait with elements of fashion photography.}
\item{a portrait of an auto mechanic in her workshop, holding a wrench in one hand. a old sports car in the background, with a workbench and tools all around. bokeh, high quality dslr photograph.}
\item{An origami owl made of brown paper is perched on a branch of an evergreen tree. The owl is facing forward with its eyes closed, giving it a peaceful appearance. The background is a blur of green foliage, creating a natural and serene setting.}
\item{A weathered, wooden mech robot covered in flowering vines stands peacefully in a field of tall wildflowers, with a small bluebird resting on its outstretched metallic hand. Digital cartoon, with warm colors and soft lines. A large cliff with waterfall looms behind.}
\item{Close-up, low angle view of a rabbit biting into a cabbage on a plate on a counter. A man wearing glasses is yelling at the rabbit and reaching out his hand to snatch the cabbage. High-contrast visuals and cinematic lighting. Fujifilm XF 10-24mm f/4, action shot.}
\item{Photo of vinyl toy scene. A colossal stone robot adorned with giant stone gardening tools stands in a lush, futuristic garden. A single sprout peeks out from a patch of fertile soil nearby. Digital art with a soft, dreamlike quality. Vinyl miniature scene.}
\item{A pair of well-worn hiking boots, caked in mud and resting on a rocky trail. There's a squirrel's head poking out of one of the boots. There's a mountainous landscape in the background, captured with a Nikon D780.}
\item{A joyful woman with a prosthetic leg and athletic attire celebrates reaching the summit of a snowy mountain. She stands triumphantly next to her snowboard, with the vast landscape stretching out behind her. captured with a Leica M11 rangefinder camera for a timeless, film-like aesthetic.}
\item{Three women stand together outside with the sun setting behind them creating a lens flare. One woman in the foreground is slightly out of focus and wearing a black felt hat. The middle woman is in focus, wearing glasses, and laughing with her head tossed back. The third woman has blonde hair pulled back in a bun and is wearing a cream sweater. She is looking at the woman in glasses and smiling.}
\item{Two contrasting figures, one wooden and jagged, the other smooth, diamond, embrace in a sun-drenched courtyard – the Harmony of Opposites.}
\item{pixel art of a space shuttle blasting of, with ``STS-1'' written below it. Cape Canaveral in the background, blue skies, with plumes of smoke billowing out.}
\item{A yellow toy submarine diving deep under the blue ocean. Close-up nature photography, sunlight coming through the water.}
\item{A busy city street with people crossing the road at an intersection, illuminated by sunlight, showcasing diverse age groups and styles as they walk across zebra stripes on the pavement. The focus is sharp on one person in red , standing out against their surroundings. Shot during golden hour to capture the warm lighting effects.}
\item{An antique pocket watch with Roman numerals and an ornate chain, lying on a worn leather surface with a vintage map in the background, captured with a Leica Q2.}
\item{A cute 1970's convertible sports car sits in front of a pub in an ink wash painting, capturing a charming English village scene with people walking around.}
\item{Joy shines in the eyes of a young woman, a charcoal portrait showing she's ready to make a difference in the world.}
\item{An elderly woman wearing a straw hat and a pink jacket is sitting next to a brown and white dog. Both the woman and the dog are looking off into the distance with serene expressions. The lighting is the warm, golden light of sunset, which creates a peaceful and contemplative atmosphere. This is a lifestyle portrait capturing a quiet moment.}
\item{A long exposure photo of the Milky Way in a starry night sky, centered over an ocean beach at magic hour. The milky way is bright and prominent with many stars visible against a dark blue black atmosphere in light painting photography with vivid and bold colors. Shot on a professional camera medium format camera with high contrast and a cinematic composition in the style.}
\item{A single comic book panel of a boy and his father on a grassy hill, staring at the sunset. A speech bubble points from the boy's mouth says ``The sun will rise again''. Muted, late 1990s coloring style.}
\item{Detailed illustration of majestic lion roaring proudly in a dream-like jungle, purple white line art background, clipart on light violet paper texture}
\item{A close-up portrait of a young woman with blonde hair and brown eyes. She is lying down and covering her mouth with a dark blue sweater, only her eyes are visible. The background is dark and blurry. The light is coming from above, creating shadows on her face.}
\end{itemize}
}

\subsection*{Figure~\ref{fig:qualitative2}}
{\footnotesize
\begin{itemize}
\item{A mother fox playing with her baby, showing love and affection in the natural environment of their habitat. The photo captures them sharing a moment, showcasing the bond between animals. The focus is on their faces.}
\item{Shot in the style of DSLR camera with the polarizing filter. A photo of three hot air balloons floating over the unique rock formations in Cappadocia, Turkey. The colors and patterns on these balloons contrast beautifully against the earthy tones of the landscape below. This shot captures the sense of adventure that comes with enjoying such an experience}
\end{itemize}
}

\section{Evaluation}

\subsection{Automatic-Evaluation Metric Comparisons}
\label{app:auto_eval_metric_comparisons}

\begin{figure}
    \centering
    \begin{subfigure}[t]{\linewidth}
    \includegraphics[width=\linewidth]{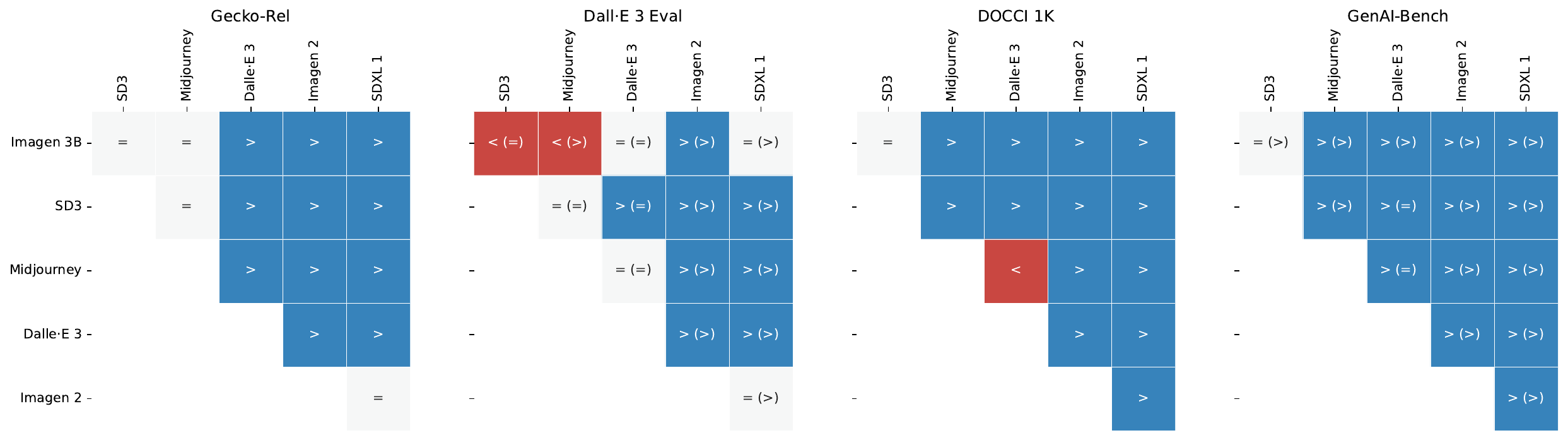}
    \subcaption{\gecko.}
    \end{subfigure}
    \begin{subfigure}[t]{\linewidth}
    \includegraphics[width=\linewidth]{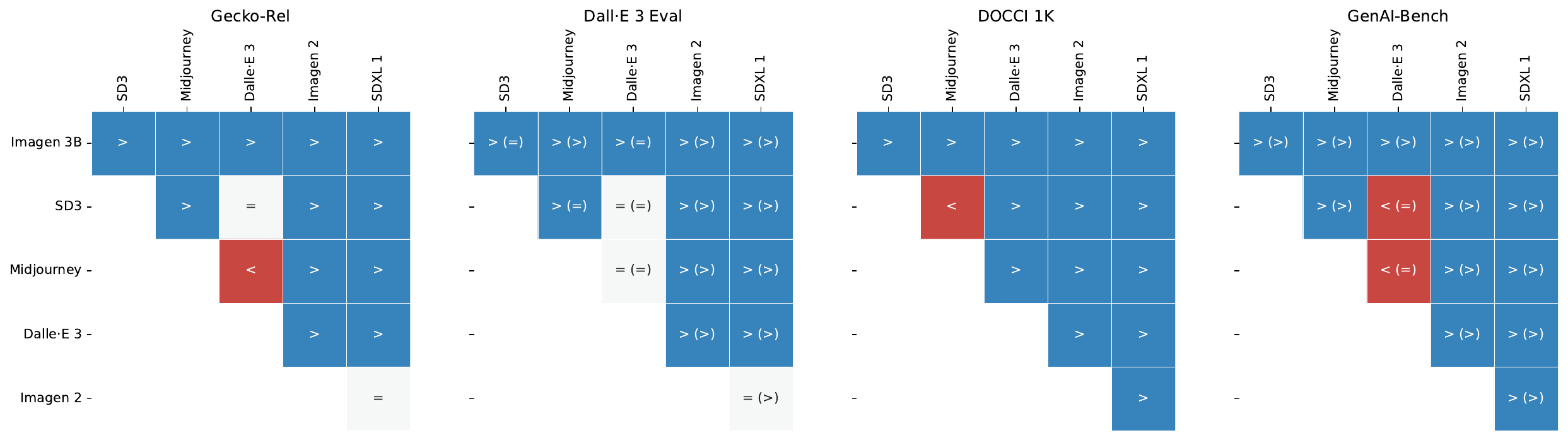}
    \subcaption{\vqascore.}
    \end{subfigure}
    \caption{{\bf Comparing T2I models using two T2I alignment metrics on four benchmarks.} We plot where metrics find significant differences between  pairs of models. We use the Wilcoxon signed rank test when comparing metrics as done in \cite{wiles2024revisiting}. We color the square according to the auto-eval metric: blue and red where the auto-eval finds a significant ($p<0.05$) difference between the pair (grey where it does not) and the color indicates the direction (blue is when the model on the y-axis is better, red when the model on the x-axis is). Where we have human annotation, we indicate in parenthesis human raters’ preference. Metrics rarely confuse wins with losses. Most confusions arise from wins or losses being confused with ties.}
    \label{fig:t2i_alignment_gecko_evaluation}
\end{figure}

\begin{table}[b]
    \centering
    \begin{tabular}{ccccccc}  \toprule
                 & & & \multicolumn{3}{c}{Metric Evaluated} \\
                 & Human Eval Setup & \# Models Evaluated & \clip & \vqascore & \gecko \\ \cmidrule(lr){2-6}
        \dalleeval & Alignment & 15 & 7 & 11 & 10 \\
        \genaibench & Alignment & 15 & 6 & 13 & 12\\
        Total & Alignment & 30 & 43.3\% & 80\% & 73.3\% \\
        
        \bottomrule
    \end{tabular}
    \caption{{\bf Auto-eval metrics performance.} We compare how often auto-eval metrics are able to predict the model ranking determined by human preferences. There are three classes: `win', `loss', and `tie'. }
    \label{tab:auto_eval_metrics_model_comparisons}
\end{table}

Here we discuss the differences between the three metrics and how we validated \gecko~and \vqascore~with human evaluation. We report the significant model orderings from \vqascore~and \gecko~in \autoref{fig:t2i_alignment_gecko_evaluation}. We can see that for models where there is a large gap in performance (e.g.~\sdxl, \imagentwo~versus the other models, as demonstrated in Section~\ref{sec:humanevaluation}), that both auto-eval metrics reliably separate the model pairs. However, when models are more similar (e.g.\ \sdthree, \imagenthree~and \dalle), then there is some disagreement or metrics do not differentiate between the models. 

We evaluate how often human annotators agree with the results in order to determine reliability of these metrics. Humans perform a side by side task of determining if one image is more aligned to the prompt than another (as explained in Sec.~\ref{subsection:t2i_alignment}). We then aggregate human scores and determine confidence intervals for each side by side comparison. We differentiate ties from wins, losses when the confidence interval includes the 50\% value. We look at how often metric orderings match human orderings on 30 pairs of models and report results in \autoref{tab:auto_eval_metrics_model_comparisons}.
First, we see that \clip~performs poorly (at 43.3\%) and is not reliable. Second, we see that both \gecko~and \vqascore~perform well in this challenging case: agreeing with human annotators for 73.3-80.0\% of the model pairs. Interestingly, we see in \autoref{fig:t2i_alignment_gecko_evaluation} that there is only one case where either \vqascore~or \gecko~mixes up the direction (e.g.~confuses a win with a loss or vice-versa). Both \vqascore~and \gecko~metrics are useful and robust even in these very challenging cases, with \vqascore~being a bit more reliable than \gecko.
Further, when these metrics agree, the agreed model ordering matches human ratings 94.4\% of the time. In these cases, we can be confident in the predicted model orderings.

\subsection{Additional Results on Numerical Reasoning}
\label{app:eval_geckonum}

In this section, we present additional data in support of results in Section~\ref{subsection:numerical_reasoning}. Figure~\ref{fig:app_geckonum_per_count_results} shows a per-number accuracy breakdown for different ground truth numbers in the text prompts.
While both \imagenthree~and \dalle~are the most accurate models when generating images containing exactly one object (see bars above the x-tick ``1''), \imagenthree~had the highest overall accuracy when generating images with more than one object (with \sdthree~having overlapping confidence intervals at n=3 and n=4 with \imagenthree).
As well, \imagenthree~is the strongest model on prompt types with a more complex structure (i.e., \emph{*-additive} and \emph{attribute-spatial} prompts), as shown in Figure~\ref{fig:app_geckonum_per_prompt_results}. 

As with all other models we investigated, the accuracy of \imagenthree{} also depends on the specific number in the text prompt. Specifically, accuracy drops with each successive number so that, on average, the model is \num{51.6} percentage points less accurate on prompts asking for ``5'' objects (i.e.\ ``5 apples''), compared to prompts asking for ``1'' object (i.e.\ ``1 apple'') (see Figure~\ref{fig:app_geckonum_per_count_results}). These results indicate that an accurate depiction of any quantity in an image remains an open challenge in text-to-image models.

\begin{figure}[h]
    \centering
    \hspace{-6.8mm}\resizebox{1.04\textwidth}{!}{%
    \input{assets/human_evals/app_geckonum_per_count.tikz}}
    \caption{{\bf Per number accuracy on all prompts in Number Generation Task.} The ground truth number on the x-axis is the original number in the text prompt used to generate the image. Accuracy is computed based on human annotations of actual counts in the images. Error bars indicate 95\% confidence intervals obtained via bootstrapping.}
    \label{fig:app_geckonum_per_count_results}
\end{figure}

\begin{figure}[h]
    \centering
    \hspace{-6.8mm}\resizebox{1.04\textwidth}{!}{%
    \input{assets/human_evals/app_geckonum_per_prompt.tikz}}
    \caption{{\bf Accuracy breakdown per different types of prompt in the GeckoNum benchmark}. On 6/7 prompt types \imagenthree~had the highest average accuracy. Error bars indicate 95\% confidence intervals obtained via bootstrapping.}
    \label{fig:app_geckonum_per_prompt_results}
\end{figure}

\clearpage

\section{Imagen 3-002 Update}
\label{app:imagen3-002}

\subsection{Human Evaluation}
\label{app:eval_imagen3-002}

In December 2024 we released an updated, higher quality version of \imagenthree{}.
This section updates our human evaluation to reflect the performance of this new \imagenthree{} version (which we refer to as ``\imagenthree{}-002''). We also
added five recent external models: Recraft v3,
Ideogram~v2, FLUX1.1 [pro], Nova Canvas, and Stable Diffusion 3.5 Large (SD 3.5 L) to our evaluation.
We refer to the previous \imagenthree{} version as ``\imagenthree{}-001''.

We report results on \genaibench{} and run side-by-side comparisons on three quality aspects: (i) overall preference, (ii) visual quality, and (iii) prompt-image alignment. 
We start by comparing each new model to \imagenthree{}-001 (previous best Elo score) and computing their preliminary Elo score.
We then run side-by-side comparisons of each model against the four better and four worse models in terms of Elo scores. 
We aggregate all these side-by-side comparisons into our final results, which we show in Figure~\ref{fig:elo_updated_genai_bench}.
\imagenthree{}-002 has the best Elo scores in all three quality aspects.

\input{assets/human_evals/elo_imagen3_002}

\subsection{Qualitative Results}
\label{app:qual_imagen3-002}
\begin{figure}[h!]
\centering
\resizebox{\textwidth}{!}{%
\setlength{\fboxsep}{0pt}%
\setlength{\tabcolsep}{1pt}%
\begin{tabular}{c c c c}%
\fbox{\includegraphics[width=0.49\textwidth]{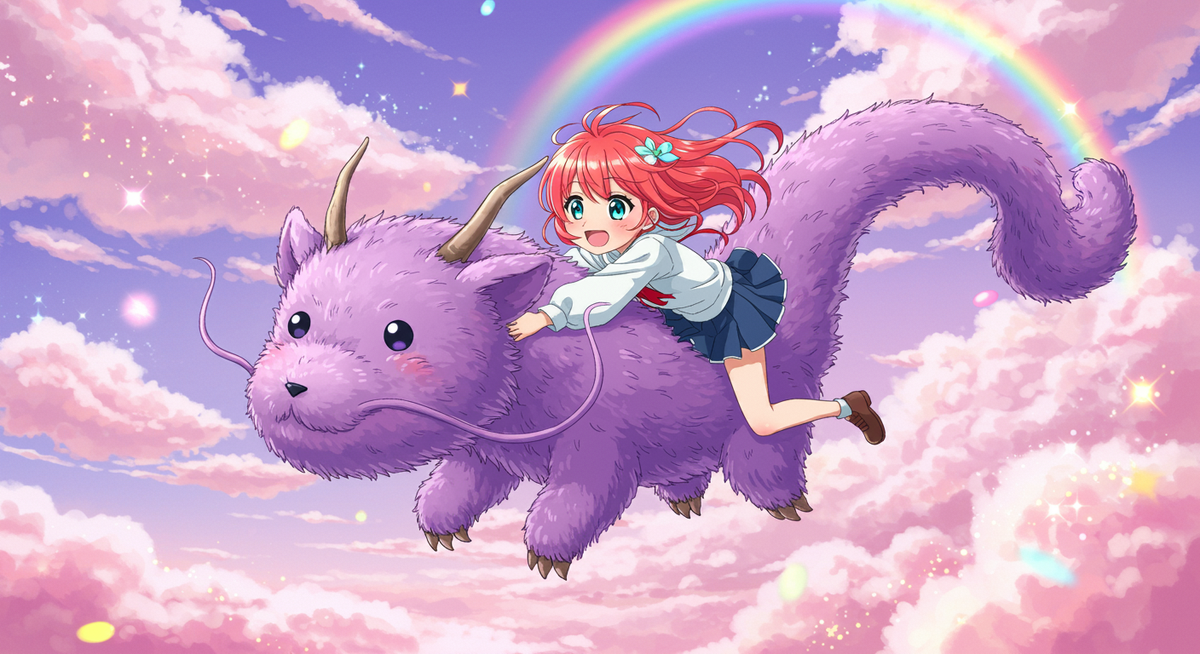}}&
\fbox{\includegraphics[width=0.49\textwidth]{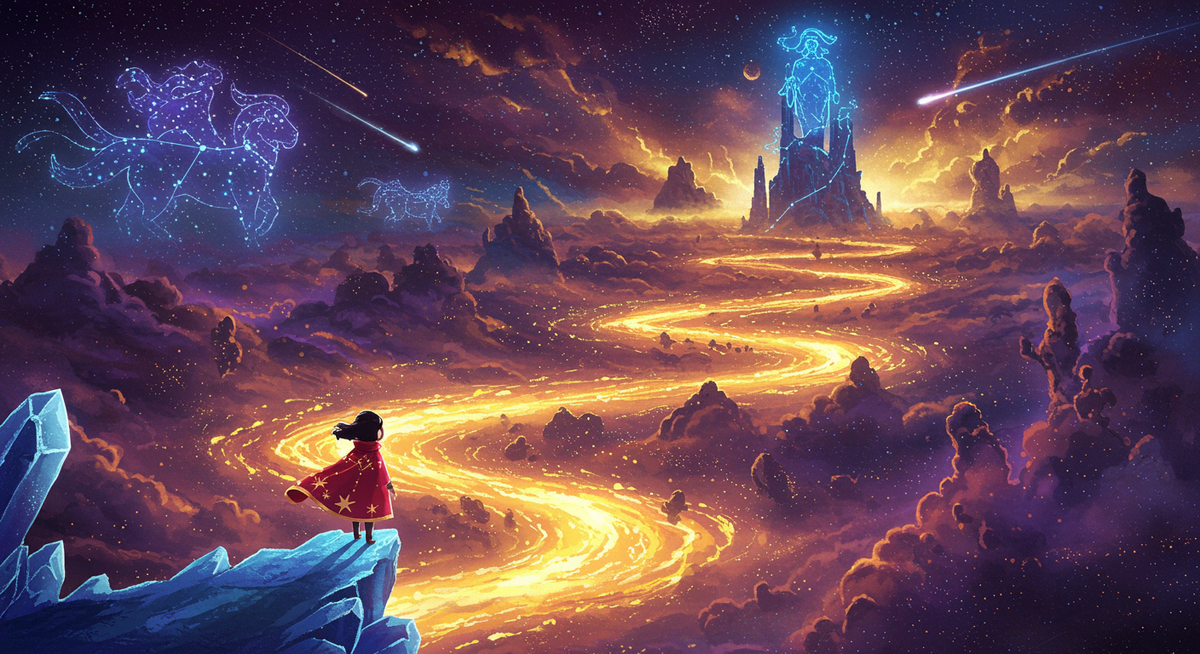}} \\[-2pt]
 
\fbox{\includegraphics[width=0.49\textwidth]{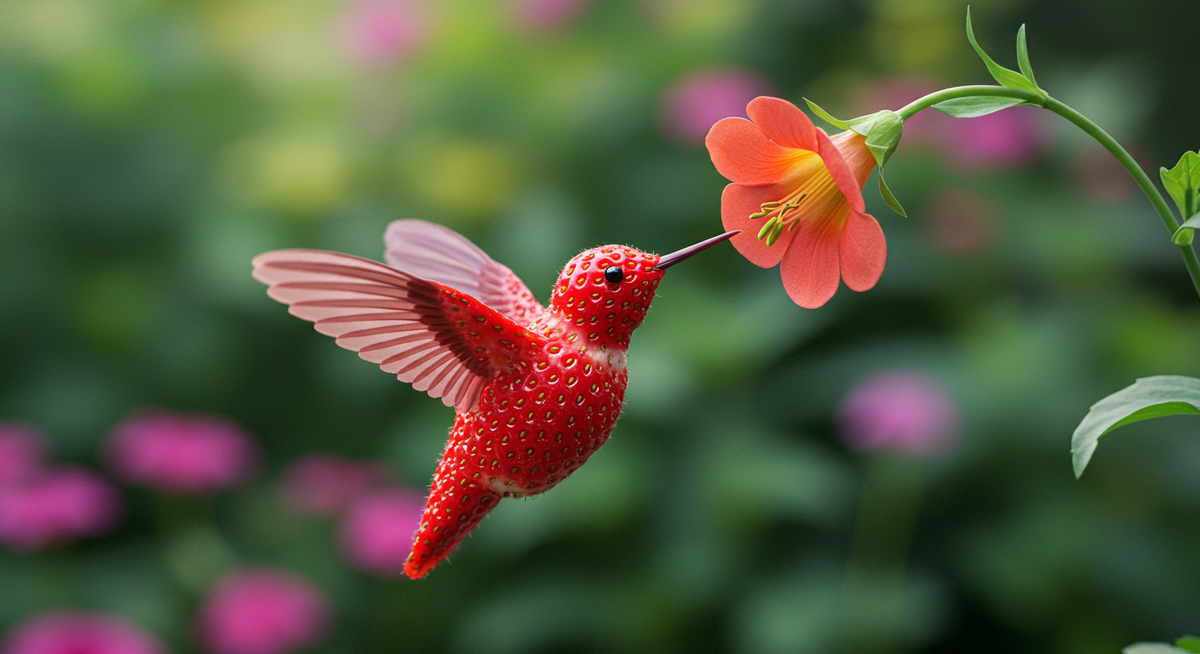}} &
\fbox{\includegraphics[width=0.49\textwidth]{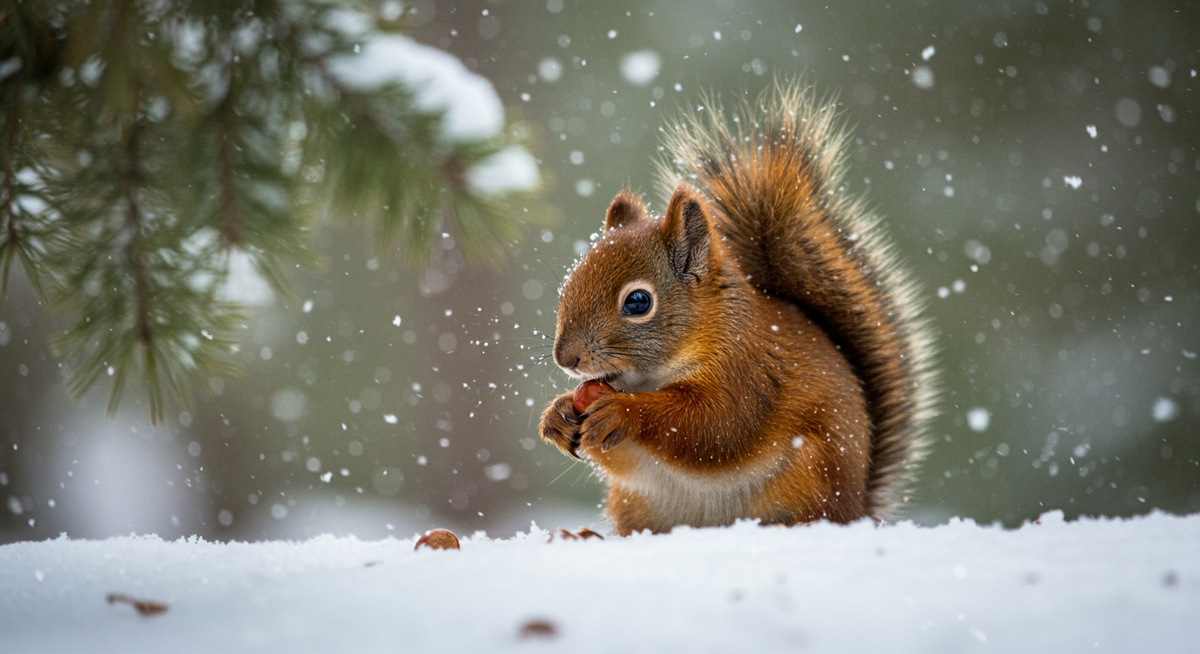}} \\[-2pt]
 
\fbox{\includegraphics[width=0.49\textwidth]{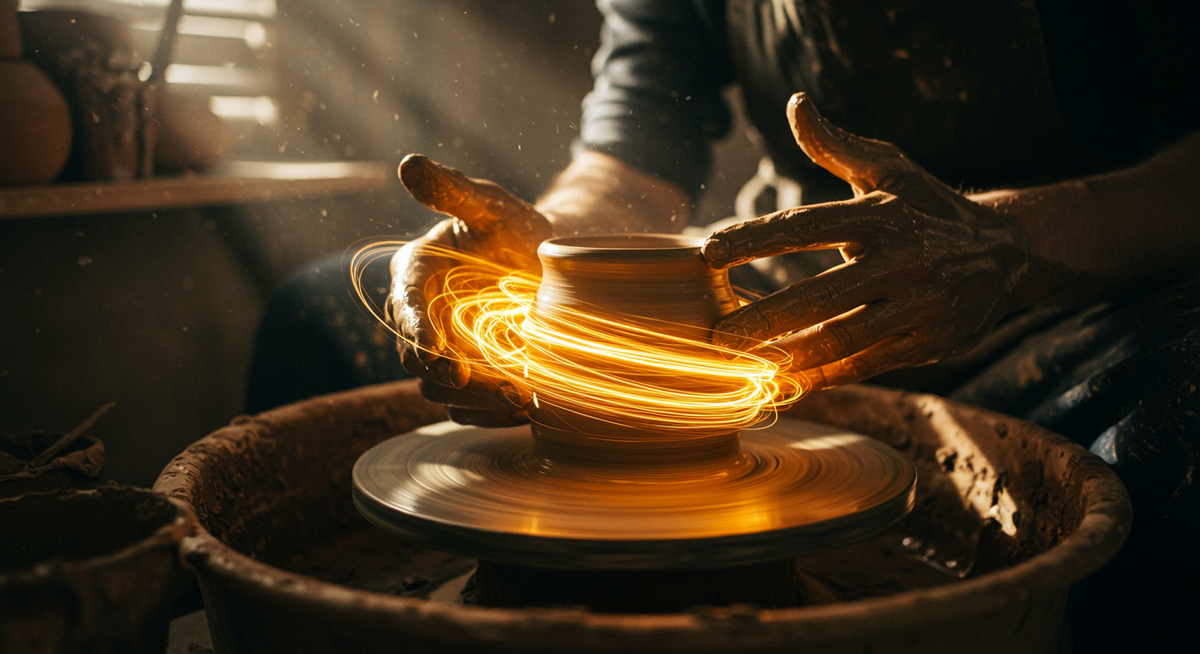}} &
\fbox{\includegraphics[width=0.49\textwidth]{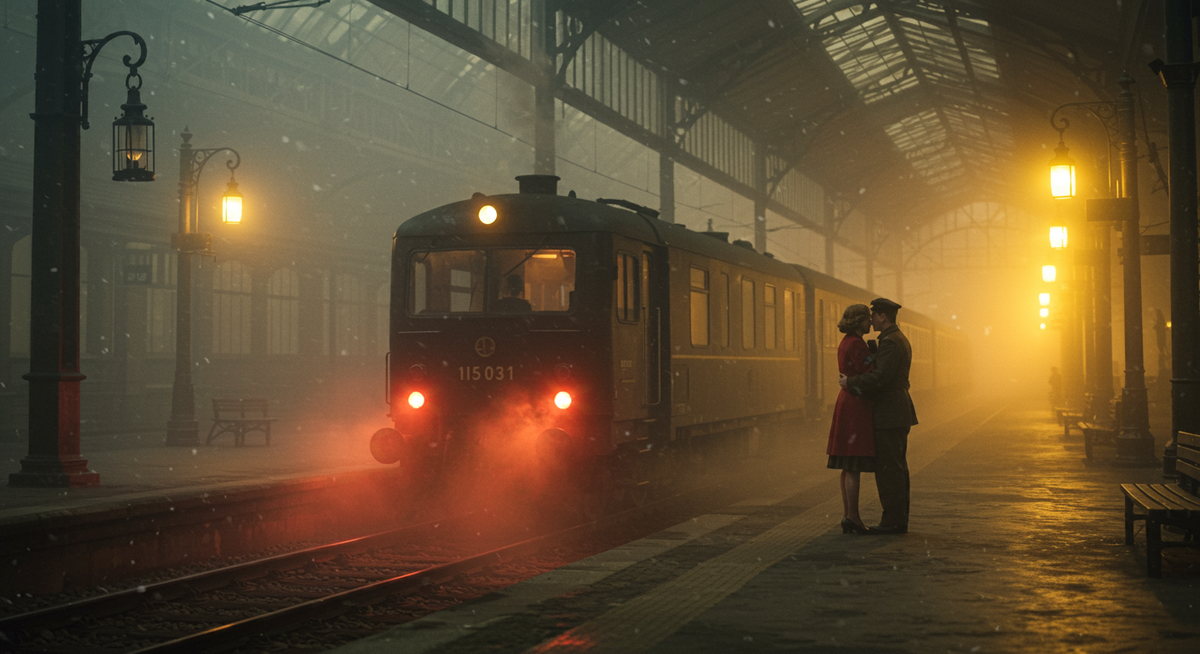}} \\[-2pt]
 
\fbox{\includegraphics[width=0.49\textwidth]{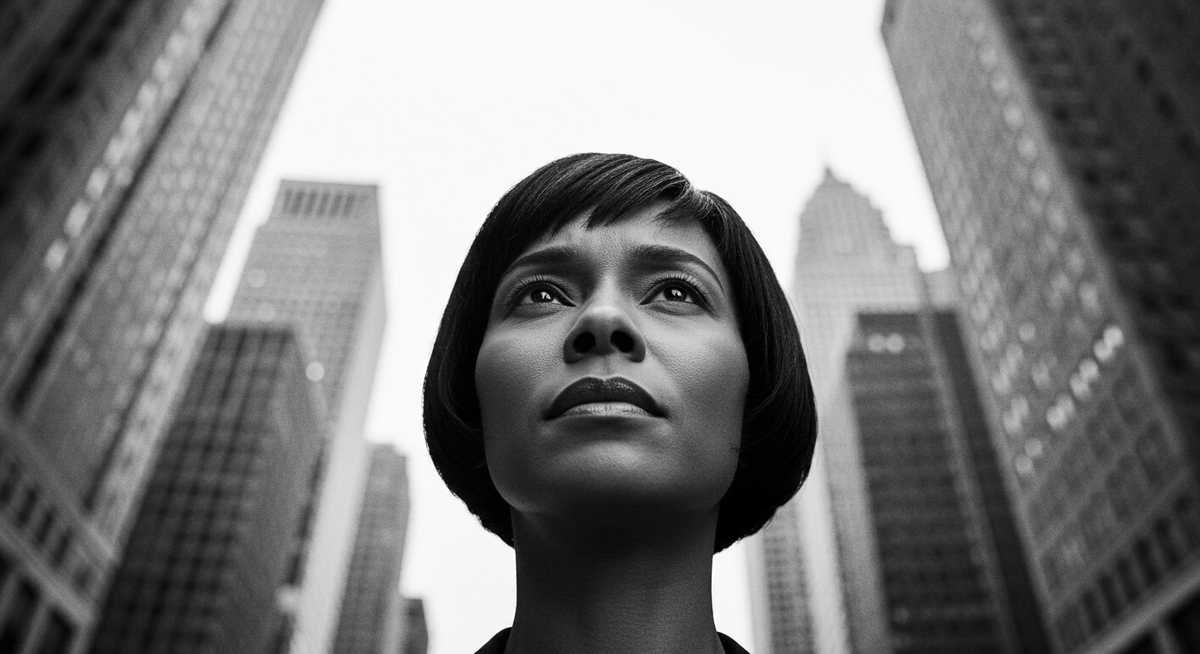}} &
\fbox{\includegraphics[width=0.49\textwidth]{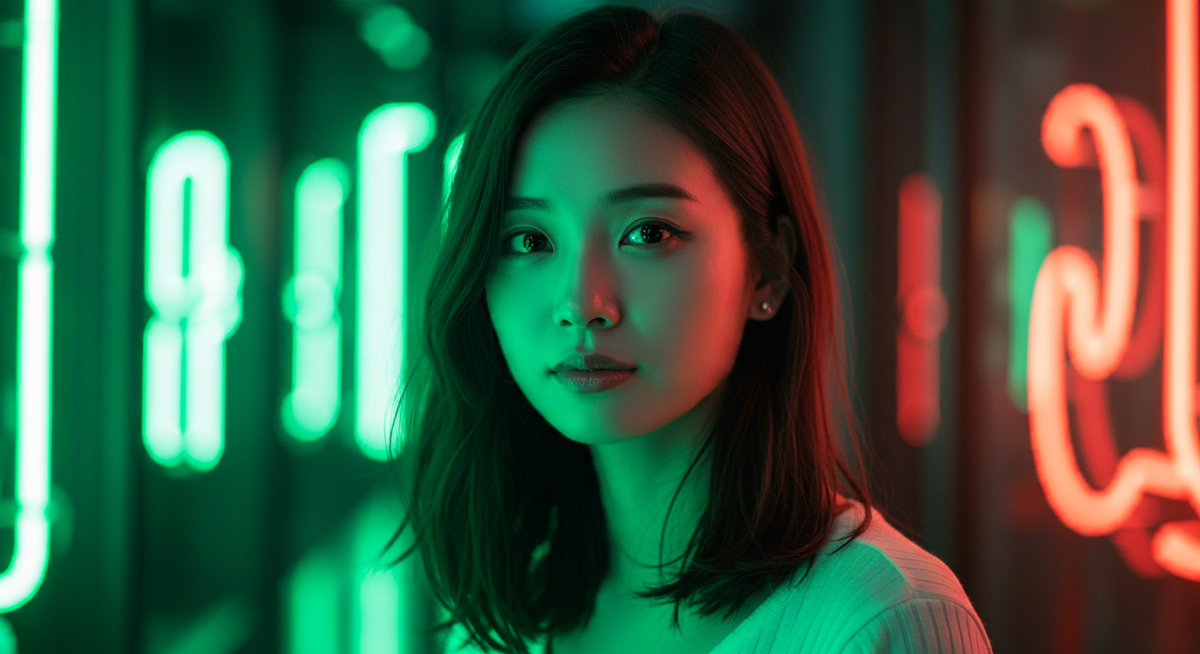}} \\
\end{tabular}}
\caption{{\bf Qualitative Results} showcasing \imagenthree{}-002's capabilities.}
\label{fig:app:qualitative}
\end{figure}

Figure~\ref{fig:app:qualitative} shows some qualitative results showcasing \imagenthree{}-002's capabilities.
The prompts that were used to generate these images are, from left to right, and top to bottom:
{\footnotesize
\begin{itemize}
\item{A vibrant illustration showcases a young anime girl clinging tightly to a fuzzy purple dragon as it soars through a fantastical sky. The girl, with her signature large, expressive eyes and bright, flowing hair, is rendered in a dynamic pose, her body leaning forward against the wind as she grips the dragon's back. The dragon itself is a fluffy, whimsical creature, its purple fur rendered with a soft, almost plush texture. They fly through a sky filled with fluffy pink clouds, glittering sparkles, and a vibrant rainbow arcing across the scene. The colors are bright and saturated, contributing to the magical and whimsical quality of the illustration. The overall mood is one of joyful, carefree adventure, emphasizing the fantastical nature of the scene and the playful bond between the girl and her unusual mount. The style is distinctly anime, with exaggerated features and a focus on dynamic movement and bright, bold colors.}
\item{Captured in the style of a high-budget animated film with vibrant, painterly textures, the frame reveals an expansive celestial landscape filled with glowing nebulae in vivid purples, blues, and golds. The protagonist, a small female figure clad in a flowing cape adorned with star motifs, stands at the edge of a crystalline cliff. Below, rivers of molten stardust wind through the galaxy, their golden light shimmering dynamically. Towering constellations shaped like mythical beasts hover in the background, their forms traced in glowing, dotted lines. Shooting stars streak across the vast sky, adding motion and brilliance to the scene. The camera angle is slightly elevated, capturing both the scale of the galaxy and the intimate journey of the protagonist.}
\item{A close-up, macro photography stock photo of a strawberry intricately sculpted into the shape of a hummingbird in mid-flight, its wings a blur as it sips nectar from a vibrant, tubular flower. The backdrop features a lush, colorful garden with a soft, bokeh effect, creating a dreamlike atmosphere. The image is exceptionally detailed and captured with a shallow depth of field, ensuring a razor-sharp focus on the strawberry-hummingbird and gentle fading of the background. The high resolution, professional photographers style, and soft lighting illuminate the scene in a very detailed manner, professional color grading amplifies the vibrant colors and creates an image with exceptional clarity. The depth of field makes the hummingbird and flower stand out starkly against the bokeh background.} 
\item{A close-up shot captures a winter wonderland scene – soft snowflakes fall on a snow-covered forest floor.  Behind a frosted pine branch, a red squirrel sits, its bright orange fur a splash of color against the white. It holds a small hazelnut. As it enjoys its meal, it seems oblivious to the falling snow.} 
\item{An extreme close-up of a craftsperson's hands shaping a glowing piece of pottery on a wheel. Threads of golden, luminous energy connect the potter’s hands to the clay, swirling dynamically with their movements. The workspace is filled with rich textures—dusty shelves lined with tools, scattered clay fragments, and beams of natural light piercing through wooden shutters. The interplay of light and energy creates an ethereal, almost magical atmosphere} 
\item{A foggy 1940s European train station at dawn, framed by intricate wrought-iron arches and misted glass windows. Steam rises from the tracks, blending with dense fog. Two lovers stand in an emotional embrace near the train, backlit by the warm, amber glow of dim lanterns. The departing train is partially visible, its red tail lights fading into the mist. The woman wears a faded red coat and clutches a small leather diary, while the man is dressed in a weathered soldier’s uniform. Dust motes float in the air, illuminated by the soft golden backlight. The atmosphere is melancholic and timeless, evoking the bittersweet farewell of wartime cinema.} 
\item{A low-angle close-up shot, in stark black and white, focuses on a woman with a short, precisely cut bob. Her expression is one of deep concern; her eyebrows are slightly furrowed, her mouth drawn into a thin line, and her eyes hold a worried intensity. The high contrast of the black and white photography emphasizes the texture of her skin and the lines around her eyes, accentuating her worried expression. The background is a blurred but imposing array of tall skyscrapers, their forms rendered in varying shades of grey, creating a sense of depth and scale. The low angle, shooting upwards, emphasizes her upward gaze, suggesting a sense of being overwhelmed by the weight of her worries within the vast urban landscape. The overall mood is one of serious apprehension, a powerful and poignant image of a woman grappling with anxieties within a monumental city.}
\item{A portrait of an Asian woman with neon green lights in the background, shallow depth of field.}
\end{itemize}
}

\clearpage
\bibliography{main}

\clearpage
\input{contributions}

\end{document}

%% file: assets/human_evals/elo_overall_preference.tikz
\pgfplotstableread[col sep=comma]{
index,model,elo,low_99_ci,high_99_ci
0,SDXL 1,860,851,868
1,Imagen 2,941,934,949
2,MJ v6,1027,1020,1033
3,DALL·E 3,1028,1021,1035
4,SD 3,1047,1040,1053
5,Imagen 3,1098,1092,1104
}\modelnamesGenAI

\pgfplotstableread[col sep=comma]{
index,model,elo,low_99_ci,high_99_ci
0,SDXL 1,860,851,868
2,MJ v6,1027,1020,1033
3,DALL·E 3,1028,1021,1035
4,SD 3,1047,1040,1053
}\datatableAGenAI

\pgfplotstableread[col sep=comma]{
index,model,elo,low_99_ci,high_99_ci
1,Imagen 2,941,934,949
5,Imagen 3,1098,1092,1104
}\datatableBGenAI

\pgfplotstableread[col sep=comma]{
index,model,elo,low_99_ci,high_99_ci
0,SDXL 1,857,848,867
1,Imagen 2,875,866,884
2,DALL·E 3,1058,1049,1066
3,SD 3,1062,1054,1071
4,MJ v6,1068,1060,1077
5,Imagen 3,1079,1071,1087
}\modelnamesDalle

\pgfplotstableread[col sep=comma]{
index,model,elo,low_99_ci,high_99_ci
0,SDXL 1,857,848,867
2,DALL·E 3,1058,1049,1066
3,SD 3,1062,1054,1071
4,MJ v6,1068,1060,1077
}\datatableADalle

\pgfplotstableread[col sep=comma]{
index,model,elo,low_99_ci,high_99_ci
1,Imagen 2,875,866,884
5,Imagen 3,1079,1071,1087
}\datatableBDalle

\pgfplotstableread[col sep=comma]{
index,model,elo,low_99_ci,high_99_ci
0,SDXL 1,923,915,931
1,Imagen 2,929,921,937
2,DALL·E 3,999,991,1007
3,MJ v6,1027,1020,1035
4,SD 3,1053,1046,1061
5,Imagen 3,1068,1060,1076
}\modelnamesDrawbench

\pgfplotstableread[col sep=comma]{
index,model,elo,low_99_ci,high_99_ci
0,SDXL 1,923,915,931
2,DALL·E 3,999,991,1007
3,MJ v6,1027,1020,1035
4,SD 3,1053,1046,1061
}\datatableADrawbench

\pgfplotstableread[col sep=comma]{
index,model,elo,low_99_ci,high_99_ci
1,Imagen 2,929,921,937
5,Imagen 3,1068,1060,1076
}\datatableBDrawbench

\begin{figure}
\centering
\resizebox{1.01\linewidth}{!}{%
\mbox{%
\hspace{-3mm}%
\begin{minipage}{0.65\linewidth}
\resizebox{\linewidth}{!}{%
\begin{tikzpicture}
\begin{axis} [
        x label style={at={(0.5,-0.1)}},
        y label style={at={(-0.07,0.5)}},
        ylabel={Elo score (and 99\% CI)},
        title={Overall preference on \genaibench{}},
        width=15cm,
        height=10cm,
    	ybar,
    	major grid style={line width=.2pt,draw=gray!50,dashed},
    	ymajorgrids,
    	tick style={draw=none},
    	xticklabels from table={\modelnamesGenAI}{model},
    	xtick={0,...,5},
    	xmin=-0.5,
    	xmax=5.5,
    	xticklabel style={yshift=0mm, align=center,rotate=0},
    	every node near coord/.append style={xshift=0,yshift=-25,color=black},
    	every axis plot/.append style={
          ybar,
          bar width=0.85,
          bar shift=0pt,
          draw=none
        },
        nodes near coords,
]
\addplot+ [mygray, error bars/.cd,
            y dir=both,
            y explicit,
            error bar style={black}]
        table
          [x=index,
            y=elo,
            y error plus expr=\thisrow{high_99_ci}-\thisrow{elo},
            y error minus expr=\thisrow{elo}-\thisrow{low_99_ci}
            ] {\datatableAGenAI};
\addplot+ [myblue, error bars/.cd,
            y dir=both,
            y explicit,
            error bar style={black}]
        table
          [x=index,
            y=elo,
            y error plus expr=\thisrow{high_99_ci}-\thisrow{elo},
            y error minus expr=\thisrow{elo}-\thisrow{low_99_ci}
            ] {\datatableBGenAI};     

\end{axis}
\end{tikzpicture}}
\end{minipage}
\hspace{2mm}
\begin{minipage}{0.35\linewidth}
\resizebox{\linewidth}{!}{%
\begin{tikzpicture}[scale=1.1]

\draw[very thin] (-0.5,5.5) rectangle +(1,1);
\draw[very thin, fill={rgb,1:red,0.994;green,0.794;blue,0.474}] (-0.5,4.5) rectangle +(1,1);
\node[black] at (0,5) {42.2\strut};
\draw[very thin, fill={rgb,1:red,0.993;green,0.717;blue,0.409}] (-0.5,3.5) rectangle +(1,1);
\node[black] at (0,4) {40.0\strut};
\draw[very thin, fill={rgb,1:red,0.994;green,0.794;blue,0.474}] (-0.5,2.5) rectangle +(1,1);
\node[black] at (0,3) {42.0\strut};
\draw[very thin, fill={rgb,1:red,0.890;green,0.287;blue,0.198}] (-0.5,1.5) rectangle +(1,1);
\node[white] at (0,2) {30.1\strut};
\draw[very thin, fill={rgb,1:red,0.647;green,0.000;blue,0.149}] (-0.5,0.5) rectangle +(1,1);
\node[white] at (0,1) {22.3\strut};
\draw[very thin, fill={rgb,1:red,0.765;green,0.900;blue,0.489}] (0.5,5.5) rectangle +(1,1);
\node[black] at (1,6) {57.8\strut};
\draw[very thin] (0.5,4.5) rectangle +(1,1);
\draw[very thin, fill={rgb,1:red,0.998;green,0.931;blue,0.633}] (0.5,3.5) rectangle +(1,1);
\node[black] at (1,4) {46.8\strut};
\draw[very thin, fill={rgb,1:red,0.997;green,0.917;blue,0.609}] (0.5,2.5) rectangle +(1,1);
\node[black] at (1,3) {46.3\strut};
\draw[very thin, fill={rgb,1:red,0.982;green,0.607;blue,0.346}] (0.5,1.5) rectangle +(1,1);
\node[black] at (1,2) {37.4\strut};
\draw[very thin, fill={rgb,1:red,0.816;green,0.162;blue,0.152}] (0.5,0.5) rectangle +(1,1);
\node[white] at (1,1) {27.2\strut};
\draw[very thin, fill={rgb,1:red,0.686;green,0.866;blue,0.439}] (1.5,5.5) rectangle +(1,1);
\node[black] at (2,6) {60.0\strut};
\draw[very thin, fill={rgb,1:red,0.915;green,0.964;blue,0.633}] (1.5,4.5) rectangle +(1,1);
\node[black] at (2,5) {53.2\strut};
\draw[very thin] (1.5,3.5) rectangle +(1,1);
\draw[very thin, fill={rgb,1:red,1.000;green,0.998;blue,0.745}] (1.5,2.5) rectangle +(1,1);
\node[black] at (2,3) {49.9\strut};
\draw[very thin, fill={rgb,1:red,0.978;green,0.577;blue,0.332}] (1.5,1.5) rectangle +(1,1);
\node[black] at (2,2) {36.7\strut};
\draw[very thin, fill={rgb,1:red,0.926;green,0.362;blue,0.233}] (1.5,0.5) rectangle +(1,1);
\node[white] at (2,1) {31.9\strut};
\draw[very thin, fill={rgb,1:red,0.765;green,0.900;blue,0.489}] (2.5,5.5) rectangle +(1,1);
\node[black] at (3,6) {58.0\strut};
\draw[very thin, fill={rgb,1:red,0.898;green,0.957;blue,0.609}] (2.5,4.5) rectangle +(1,1);
\node[black] at (3,5) {53.7\strut};
\draw[very thin, fill={rgb,1:red,0.997;green,0.999;blue,0.745}] (2.5,3.5) rectangle +(1,1);
\node[black] at (3,4) {50.1\strut};
\draw[very thin] (2.5,2.5) rectangle +(1,1);
\draw[very thin, fill={rgb,1:red,0.993;green,0.748;blue,0.435}] (2.5,1.5) rectangle +(1,1);
\node[black] at (3,2) {40.9\strut};
\draw[very thin, fill={rgb,1:red,0.894;green,0.296;blue,0.202}] (2.5,0.5) rectangle +(1,1);
\node[white] at (3,1) {30.5\strut};
\draw[very thin, fill={rgb,1:red,0.225;green,0.656;blue,0.344}] (3.5,5.5) rectangle +(1,1);
\node[white] at (4,6) {69.9\strut};
\draw[very thin, fill={rgb,1:red,0.577;green,0.819;blue,0.408}] (3.5,4.5) rectangle +(1,1);
\node[black] at (4,5) {62.6\strut};
\draw[very thin, fill={rgb,1:red,0.548;green,0.806;blue,0.404}] (3.5,3.5) rectangle +(1,1);
\node[black] at (4,4) {63.3\strut};
\draw[very thin, fill={rgb,1:red,0.718;green,0.880;blue,0.459}] (3.5,2.5) rectangle +(1,1);
\node[black] at (4,3) {59.1\strut};
\draw[very thin] (3.5,1.5) rectangle +(1,1);
\draw[very thin, fill={rgb,1:red,0.994;green,0.786;blue,0.468}] (3.5,0.5) rectangle +(1,1);
\node[black] at (4,1) {41.8\strut};
\draw[very thin, fill={rgb,1:red,0.000;green,0.408;blue,0.216}] (4.5,5.5) rectangle +(1,1);
\node[white] at (5,6) {77.7\strut};
\draw[very thin, fill={rgb,1:red,0.088;green,0.570;blue,0.300}] (4.5,4.5) rectangle +(1,1);
\node[white] at (5,5) {72.8\strut};
\draw[very thin, fill={rgb,1:red,0.318;green,0.701;blue,0.368}] (4.5,3.5) rectangle +(1,1);
\node[white] at (5,4) {68.1\strut};
\draw[very thin, fill={rgb,1:red,0.236;green,0.662;blue,0.347}] (4.5,2.5) rectangle +(1,1);
\node[white] at (5,3) {69.5\strut};
\draw[very thin, fill={rgb,1:red,0.757;green,0.897;blue,0.484}] (4.5,1.5) rectangle +(1,1);
\node[black] at (5,2) {58.2\strut};
\draw[very thin] (4.5,0.5) rectangle +(1,1);

\node[rotate=55, anchor=west, align=left] at (-0.1,6.6) {Imagen 3};
\node[anchor=east, align=right] at (-0.5,6) {Imagen 3};
\node[rotate=55, anchor=west, align=left] at (0.9,6.6) {SD 3};
\node[anchor=east, align=right] at (-0.5,5) {SD 3};
\node[rotate=55, anchor=west, align=left] at (1.9,6.6) {DALL·E 3};
\node[anchor=east, align=right] at (-0.5,4) {DALL·E 3};
\node[rotate=55, anchor=west, align=left] at (2.9,6.6) {MJ v6};
\node[anchor=east, align=right] at (-0.5,3) {MJ v6};
\node[rotate=55, anchor=west, align=left] at (3.9,6.6) {Imagen 2};
\node[anchor=east, align=right] at (-0.5,2) {Imagen 2};
\node[rotate=55, anchor=west, align=left] at (4.9,6.6) {SDXL 1};
\node[anchor=east, align=right] at (-0.5,1) {SDXL 1};

\node[anchor=east] at (0,-0.5) {};
\end{tikzpicture}}
\end{minipage}}}

\vspace{3mm}
\resizebox{1.01\linewidth}{!}{%
\mbox{%
\hspace{-3mm}%
\begin{minipage}{0.65\linewidth}
\resizebox{\linewidth}{!}{%
\begin{tikzpicture}
\begin{axis} [
        x label style={at={(0.5,-0.1)}},
        y label style={at={(-0.07,0.5)}},
        ylabel={Elo score (and 99\% CI)},
        title={Overall preference on \drawbench},
        width=15cm,
        height=10cm,
    	ybar,
    	major grid style={line width=.2pt,draw=gray!50,dashed},
    	ymajorgrids,
    	tick style={draw=none},
    	xticklabels from table={\modelnamesDrawbench}{model},
    	xtick={0,...,5},
    	xmin=-0.5,
    	xmax=5.5,
    	xticklabel style={yshift=0mm, align=center,rotate=0},
    	every node near coord/.append style={xshift=0,yshift=-25,color=black},
    	every axis plot/.append style={
          ybar,
          bar width=0.85,
          bar shift=0pt,
          draw=none
        },
        nodes near coords,
]
\addplot+ [mygray, error bars/.cd,
            y dir=both,
            y explicit,
            error bar style={black}]
        table
          [x=index,
            y=elo,
            y error plus expr=\thisrow{high_99_ci}-\thisrow{elo},
            y error minus expr=\thisrow{elo}-\thisrow{low_99_ci}
            ] {\datatableADrawbench};
\addplot+ [myblue, error bars/.cd,
            y dir=both,
            y explicit,
            error bar style={black}]
        table
          [x=index,
            y=elo,
            y error plus expr=\thisrow{high_99_ci}-\thisrow{elo},
            y error minus expr=\thisrow{elo}-\thisrow{low_99_ci}
            ] {\datatableBDrawbench};     

\end{axis}
\end{tikzpicture}}
\end{minipage}
\hspace{2mm}
\begin{minipage}{0.35\linewidth}
\resizebox{\linewidth}{!}{%
\begin{tikzpicture}[scale=1.1]

\draw[very thin] (-0.5,5.5) rectangle +(1,1);
\draw[very thin, fill={rgb,1:red,0.999;green,0.974;blue,0.705}] (-0.5,4.5) rectangle +(1,1);
\node[black] at (0,5) {49.1\strut};
\draw[very thin, fill={rgb,1:red,0.994;green,0.763;blue,0.448}] (-0.5,3.5) rectangle +(1,1);
\node[black] at (0,4) {43.8\strut};
\draw[very thin, fill={rgb,1:red,0.973;green,0.547;blue,0.318}] (-0.5,2.5) rectangle +(1,1);
\node[black] at (0,3) {40.0\strut};
\draw[very thin, fill={rgb,1:red,0.732;green,0.081;blue,0.151}] (-0.5,1.5) rectangle +(1,1);
\node[white] at (0,2) {32.0\strut};
\draw[very thin, fill={rgb,1:red,0.655;green,0.007;blue,0.149}] (-0.5,0.5) rectangle +(1,1);
\node[white] at (0,1) {30.4\strut};
\draw[very thin, fill={rgb,1:red,0.968;green,0.986;blue,0.705}] (0.5,5.5) rectangle +(1,1);
\node[black] at (1,6) {50.9\strut};
\draw[very thin] (0.5,4.5) rectangle +(1,1);
\draw[very thin, fill={rgb,1:red,0.999;green,0.983;blue,0.721}] (0.5,3.5) rectangle +(1,1);
\node[black] at (1,4) {49.5\strut};
\draw[very thin, fill={rgb,1:red,0.986;green,0.637;blue,0.360}] (0.5,2.5) rectangle +(1,1);
\node[black] at (1,3) {41.4\strut};
\draw[very thin, fill={rgb,1:red,0.917;green,0.343;blue,0.224}] (0.5,1.5) rectangle +(1,1);
\node[white] at (1,2) {36.7\strut};
\draw[very thin, fill={rgb,1:red,0.647;green,0.000;blue,0.149}] (0.5,0.5) rectangle +(1,1);
\node[white] at (1,1) {30.2\strut};
\draw[very thin, fill={rgb,1:red,0.733;green,0.887;blue,0.469}] (1.5,5.5) rectangle +(1,1);
\node[black] at (2,6) {56.2\strut};
\draw[very thin, fill={rgb,1:red,0.980;green,0.991;blue,0.721}] (1.5,4.5) rectangle +(1,1);
\node[black] at (2,5) {50.5\strut};
\draw[very thin] (1.5,3.5) rectangle +(1,1);
\draw[very thin, fill={rgb,1:red,0.997;green,0.902;blue,0.585}] (1.5,2.5) rectangle +(1,1);
\node[black] at (2,3) {46.9\strut};
\draw[very thin, fill={rgb,1:red,0.926;green,0.362;blue,0.233}] (1.5,1.5) rectangle +(1,1);
\node[white] at (2,2) {37.1\strut};
\draw[very thin, fill={rgb,1:red,0.917;green,0.343;blue,0.224}] (1.5,0.5) rectangle +(1,1);
\node[white] at (2,1) {36.8\strut};
\draw[very thin, fill={rgb,1:red,0.518;green,0.793;blue,0.401}] (2.5,5.5) rectangle +(1,1);
\node[black] at (3,6) {60.0\strut};
\draw[very thin, fill={rgb,1:red,0.607;green,0.832;blue,0.411}] (2.5,4.5) rectangle +(1,1);
\node[black] at (3,5) {58.6\strut};
\draw[very thin, fill={rgb,1:red,0.880;green,0.950;blue,0.585}] (2.5,3.5) rectangle +(1,1);
\node[black] at (3,4) {53.1\strut};
\draw[very thin] (2.5,2.5) rectangle +(1,1);
\draw[very thin, fill={rgb,1:red,0.969;green,0.517;blue,0.304}] (2.5,1.5) rectangle +(1,1);
\node[black] at (3,2) {39.5\strut};
\draw[very thin, fill={rgb,1:red,0.983;green,0.617;blue,0.350}] (2.5,0.5) rectangle +(1,1);
\node[black] at (3,1) {41.1\strut};
\draw[very thin, fill={rgb,1:red,0.044;green,0.489;blue,0.258}] (3.5,5.5) rectangle +(1,1);
\node[white] at (4,6) {68.0\strut};
\draw[very thin, fill={rgb,1:red,0.295;green,0.690;blue,0.362}] (3.5,4.5) rectangle +(1,1);
\node[white] at (4,5) {63.3\strut};
\draw[very thin, fill={rgb,1:red,0.318;green,0.701;blue,0.368}] (3.5,3.5) rectangle +(1,1);
\node[white] at (4,4) {62.9\strut};
\draw[very thin, fill={rgb,1:red,0.489;green,0.780;blue,0.398}] (3.5,2.5) rectangle +(1,1);
\node[black] at (4,3) {60.5\strut};
\draw[very thin] (3.5,1.5) rectangle +(1,1);
\draw[very thin, fill={rgb,1:red,1.000;green,0.993;blue,0.737}] (3.5,0.5) rectangle +(1,1);
\node[black] at (4,1) {49.7\strut};
\draw[very thin, fill={rgb,1:red,0.004;green,0.415;blue,0.220}] (4.5,5.5) rectangle +(1,1);
\node[white] at (5,6) {69.6\strut};
\draw[very thin, fill={rgb,1:red,0.000;green,0.408;blue,0.216}] (4.5,4.5) rectangle +(1,1);
\node[white] at (5,5) {69.8\strut};
\draw[very thin, fill={rgb,1:red,0.295;green,0.690;blue,0.362}] (4.5,3.5) rectangle +(1,1);
\node[white] at (5,4) {63.2\strut};
\draw[very thin, fill={rgb,1:red,0.587;green,0.823;blue,0.409}] (4.5,2.5) rectangle +(1,1);
\node[black] at (5,3) {58.9\strut};
\draw[very thin, fill={rgb,1:red,0.991;green,0.996;blue,0.737}] (4.5,1.5) rectangle +(1,1);
\node[black] at (5,2) {50.3\strut};
\draw[very thin] (4.5,0.5) rectangle +(1,1);

\node[rotate=55, anchor=west, align=left] at (-0.1,6.6) {Imagen 3};
\node[anchor=east, align=right] at (-0.5,6) {Imagen 3};
\node[rotate=55, anchor=west, align=left] at (0.9,6.6) {SD 3};
\node[anchor=east, align=right] at (-0.5,5) {SD 3};
\node[rotate=55, anchor=west, align=left] at (1.9,6.6) {MJ v6};
\node[anchor=east, align=right] at (-0.5,4) {MJ v6};
\node[rotate=55, anchor=west, align=left] at (2.9,6.6) {DALL·E 3};
\node[anchor=east, align=right] at (-0.5,3) {DALL·E 3};
\node[rotate=55, anchor=west, align=left] at (3.9,6.6) {Imagen 2};
\node[anchor=east, align=right] at (-0.5,2) {Imagen 2};
\node[rotate=55, anchor=west, align=left] at (4.9,6.6) {SDXL 1};
\node[anchor=east, align=right] at (-0.5,1) {SDXL 1};

\node[anchor=east] at (0,-0.5) {};
\end{tikzpicture}}
\end{minipage}}}

\vspace{3mm}
\resizebox{1.01\linewidth}{!}{%
\mbox{%
\hspace{-3mm}%
\begin{minipage}{0.65\linewidth}
\resizebox{\linewidth}{!}{%
\begin{tikzpicture}
\begin{axis} [
        x label style={at={(0.5,-0.1)}},
        y label style={at={(-0.07,0.5)}},
        ylabel={Elo score (and 99\% CI)},
        title={Overall preference on \dalle~Eval},
        width=15cm,
        height=10cm,
    	ybar,
    	major grid style={line width=.2pt,draw=gray!50,dashed},
    	ymajorgrids,
    	tick style={draw=none},
    	xticklabels from table={\modelnamesDalle}{model},
    	xtick={0,...,5},
    	xmin=-0.5,
    	xmax=5.5,
    	xticklabel style={yshift=0mm, align=center,rotate=0},
    	every node near coord/.append style={xshift=0,yshift=-25,color=black},
    	every axis plot/.append style={
          ybar,
          bar width=0.85,
          bar shift=0pt,
          draw=none
        },
        nodes near coords,
]
\addplot+ [mygray, error bars/.cd,
            y dir=both,
            y explicit,
            error bar style={black}]
        table
          [x=index,
            y=elo,
            y error plus expr=\thisrow{high_99_ci}-\thisrow{elo},
            y error minus expr=\thisrow{elo}-\thisrow{low_99_ci}
            ] {\datatableADalle};
\addplot+ [myblue, error bars/.cd,
            y dir=both,
            y explicit,
            error bar style={black}]
        table
          [x=index,
            y=elo,
            y error plus expr=\thisrow{high_99_ci}-\thisrow{elo},
            y error minus expr=\thisrow{elo}-\thisrow{low_99_ci}
            ] {\datatableBDalle};     

\end{axis}
\end{tikzpicture}}
\end{minipage}
\hspace{2mm}
\begin{minipage}{0.35\linewidth}
\resizebox{\linewidth}{!}{%
\begin{tikzpicture}[scale=1.1]

\draw[very thin] (-0.5,5.5) rectangle +(1,1);
\draw[very thin, fill={rgb,1:red,1.000;green,0.988;blue,0.729}] (-0.5,4.5) rectangle +(1,1);
\node[black] at (0,5) {49.5\strut};
\draw[very thin, fill={rgb,1:red,0.999;green,0.955;blue,0.673}] (-0.5,3.5) rectangle +(1,1);
\node[black] at (0,4) {48.0\strut};
\draw[very thin, fill={rgb,1:red,0.997;green,0.917;blue,0.609}] (-0.5,2.5) rectangle +(1,1);
\node[black] at (0,3) {46.3\strut};
\draw[very thin, fill={rgb,1:red,0.863;green,0.230;blue,0.172}] (-0.5,1.5) rectangle +(1,1);
\node[white] at (0,2) {29.0\strut};
\draw[very thin, fill={rgb,1:red,0.647;green,0.000;blue,0.149}] (-0.5,0.5) rectangle +(1,1);
\node[white] at (0,1) {22.3\strut};
\draw[very thin, fill={rgb,1:red,0.985;green,0.994;blue,0.729}] (0.5,5.5) rectangle +(1,1);
\node[black] at (1,6) {50.5\strut};
\draw[very thin] (0.5,4.5) rectangle +(1,1);
\draw[very thin, fill={rgb,1:red,0.999;green,0.964;blue,0.689}] (0.5,3.5) rectangle +(1,1);
\node[black] at (1,4) {48.4\strut};
\draw[very thin, fill={rgb,1:red,0.999;green,0.969;blue,0.697}] (0.5,2.5) rectangle +(1,1);
\node[black] at (1,3) {48.6\strut};
\draw[very thin, fill={rgb,1:red,0.859;green,0.221;blue,0.168}] (0.5,1.5) rectangle +(1,1);
\node[white] at (1,2) {28.6\strut};
\draw[very thin, fill={rgb,1:red,0.943;green,0.399;blue,0.250}] (0.5,0.5) rectangle +(1,1);
\node[white] at (1,1) {32.7\strut};
\draw[very thin, fill={rgb,1:red,0.944;green,0.977;blue,0.673}] (1.5,5.5) rectangle +(1,1);
\node[black] at (2,6) {52.0\strut};
\draw[very thin, fill={rgb,1:red,0.956;green,0.982;blue,0.689}] (1.5,4.5) rectangle +(1,1);
\node[black] at (2,5) {51.6\strut};
\draw[very thin] (1.5,3.5) rectangle +(1,1);
\draw[very thin, fill={rgb,1:red,0.998;green,0.945;blue,0.657}] (1.5,2.5) rectangle +(1,1);
\node[black] at (2,3) {47.5\strut};
\draw[very thin, fill={rgb,1:red,0.854;green,0.212;blue,0.164}] (1.5,1.5) rectangle +(1,1);
\node[white] at (2,2) {28.4\strut};
\draw[very thin, fill={rgb,1:red,0.801;green,0.148;blue,0.152}] (1.5,0.5) rectangle +(1,1);
\node[white] at (2,1) {26.8\strut};
\draw[very thin, fill={rgb,1:red,0.898;green,0.957;blue,0.609}] (2.5,5.5) rectangle +(1,1);
\node[black] at (3,6) {53.7\strut};
\draw[very thin, fill={rgb,1:red,0.962;green,0.984;blue,0.697}] (2.5,4.5) rectangle +(1,1);
\node[black] at (3,5) {51.4\strut};
\draw[very thin, fill={rgb,1:red,0.933;green,0.972;blue,0.657}] (2.5,3.5) rectangle +(1,1);
\node[black] at (3,4) {52.5\strut};
\draw[very thin] (2.5,2.5) rectangle +(1,1);
\draw[very thin, fill={rgb,1:red,0.747;green,0.096;blue,0.151}] (2.5,1.5) rectangle +(1,1);
\node[white] at (3,2) {25.2\strut};
\draw[very thin, fill={rgb,1:red,0.921;green,0.352;blue,0.228}] (2.5,0.5) rectangle +(1,1);
\node[white] at (3,1) {31.8\strut};
\draw[very thin, fill={rgb,1:red,0.155;green,0.622;blue,0.327}] (3.5,5.5) rectangle +(1,1);
\node[white] at (4,6) {71.0\strut};
\draw[very thin, fill={rgb,1:red,0.143;green,0.616;blue,0.324}] (3.5,4.5) rectangle +(1,1);
\node[white] at (4,5) {71.4\strut};
\draw[very thin, fill={rgb,1:red,0.131;green,0.610;blue,0.321}] (3.5,3.5) rectangle +(1,1);
\node[white] at (4,4) {71.6\strut};
\draw[very thin, fill={rgb,1:red,0.052;green,0.504;blue,0.266}] (3.5,2.5) rectangle +(1,1);
\node[white] at (4,3) {74.8\strut};
\draw[very thin] (3.5,1.5) rectangle +(1,1);
\draw[very thin, fill={rgb,1:red,0.998;green,0.931;blue,0.633}] (3.5,0.5) rectangle +(1,1);
\node[black] at (4,1) {46.9\strut};
\draw[very thin, fill={rgb,1:red,0.000;green,0.408;blue,0.216}] (4.5,5.5) rectangle +(1,1);
\node[white] at (5,6) {77.7\strut};
\draw[very thin, fill={rgb,1:red,0.365;green,0.724;blue,0.379}] (4.5,4.5) rectangle +(1,1);
\node[white] at (5,5) {67.3\strut};
\draw[very thin, fill={rgb,1:red,0.080;green,0.555;blue,0.293}] (4.5,3.5) rectangle +(1,1);
\node[white] at (5,4) {73.2\strut};
\draw[very thin, fill={rgb,1:red,0.306;green,0.696;blue,0.365}] (4.5,2.5) rectangle +(1,1);
\node[white] at (5,3) {68.2\strut};
\draw[very thin, fill={rgb,1:red,0.915;green,0.964;blue,0.633}] (4.5,1.5) rectangle +(1,1);
\node[black] at (5,2) {53.1\strut};
\draw[very thin] (4.5,0.5) rectangle +(1,1);

\node[rotate=55, anchor=west, align=left] at (-0.1,6.6) {Imagen 3};
\node[anchor=east, align=right] at (-0.5,6) {Imagen 3};
\node[rotate=55, anchor=west, align=left] at (0.9,6.6) {MJ v6};
\node[anchor=east, align=right] at (-0.5,5) {MJ v6};
\node[rotate=55, anchor=west, align=left] at (1.9,6.6) {SD 3};
\node[anchor=east, align=right] at (-0.5,4) {SD 3};
\node[rotate=55, anchor=west, align=left] at (2.9,6.6) {DALL·E 3};
\node[anchor=east, align=right] at (-0.5,3) {DALL·E 3};
\node[rotate=55, anchor=west, align=left] at (3.9,6.6) {Imagen 2};
\node[anchor=east, align=right] at (-0.5,2) {Imagen 2};
\node[rotate=55, anchor=west, align=left] at (4.9,6.6) {SDXL 1};
\node[anchor=east, align=right] at (-0.5,1) {SDXL 1};

\node[anchor=east] at (0,-0.5) {};
\end{tikzpicture}}
\end{minipage}}}

\caption{{\bf Overall preference:} Elo scores and win-rate percentages on \genaibench{}, \drawbench, and \dalle~Eval. Please refer to Appendix~\ref{app:imagen3-002} for updated human evaluation results as of December 2024.}
\label{fig:elo_overall_preference}
\end{figure}

%% file: assets/human_evals/elo_t2i_alignment.tikz
\pgfplotstableread[col sep=comma]{
index,model,elo,low_99_ci,high_99_ci
0,SDXL 1,873,866,881
1,Imagen 2,950,942,957
2,MJ v6,1019,1012,1026
3,DALL·E 3,1028,1021,1035
4,SD 3,1047,1040,1054
5,Imagen 3,1083,1078,1089
}\modelnamesGenAI

\pgfplotstableread[col sep=comma]{
index,model,elo,low_99_ci,high_99_ci
0,SDXL 1,873,866,881
2,MJ v6,1019,1012,1026
3,DALL·E 3,1028,1021,1035
4,SD 3,1047,1040,1054
}\datatableAGenAI

\pgfplotstableread[col sep=comma]{
index,model,elo,low_99_ci,high_99_ci
1,Imagen 2,950,942,957
5,Imagen 3,1083,1078,1089
}\datatableBGenAI

\pgfplotstableread[col sep=comma]{
index,model,elo,low_99_ci,high_99_ci
0,SDXL 1,931,924,939
1,Imagen 2,934,926,942
2,MJ v6,1011,1004,1018
3,DALL·E 3,1013,1005,1021
4,SD 3,1047,1039,1055
5,Imagen 3,1064,1056,1072
}\modelnamesDrawbench

\pgfplotstableread[col sep=comma]{
index,model,elo,low_99_ci,high_99_ci
0,SDXL 1,931,924,939
2,MJ v6,1011,1004,1018
3,DALL·E 3,1013,1005,1021
4,SD 3,1047,1039,1055
}\datatableADrawbench

\pgfplotstableread[col sep=comma]{
index,model,elo,low_99_ci,high_99_ci
1,Imagen 2,934,926,942
5,Imagen 3,1064,1056,1072
}\datatableBDrawbench

\pgfplotstableread[col sep=comma]{
index,model,elo,low_99_ci,high_99_ci
0,SDXL 1,848,839,858
1,Imagen 2,876,867,885
2,MJ v6,1052,1044,1061
3,SD 3,1069,1061,1077
4,DALL·E 3,1077,1068,1085
5,Imagen 3,1078,1069,1086
}\modelnamesDalle

\pgfplotstableread[col sep=comma]{
index,model,elo,low_99_ci,high_99_ci
0,SDXL 1,848,839,858
2,MJ v6,1052,1044,1061
3,SD 3,1069,1061,1077
4,DALL·E 3,1077,1068,1085
}\datatableADalle

\pgfplotstableread[col sep=comma]{
index,model,elo,low_99_ci,high_99_ci
1,Imagen 2,876,867,885
5,Imagen 3,1078,1069,1086
}\datatableBDalle

\begin{figure}
\centering
\resizebox{1.01\linewidth}{!}{%
\mbox{%
\hspace{-3mm}%
\begin{minipage}{0.65\linewidth}
\resizebox{\linewidth}{!}{%
\begin{tikzpicture}
\begin{axis} [
        x label style={at={(0.5,-0.1)}},
        y label style={at={(-0.07,0.5)}},
        ylabel={Elo score (and 99\% CI)},
        title={Prompt-image alignment on \genaibench{}},
        width=15cm,
        height=10cm,
    	ybar,
    	major grid style={line width=.2pt,draw=gray!50,dashed},
    	ymajorgrids,
    	tick style={draw=none},
    	xticklabels from table={\modelnamesGenAI}{model},
    	xtick={0,...,5},
    	xmin=-0.5,
    	xmax=5.5,
    	xticklabel style={yshift=0mm, align=center,rotate=0},
    	every node near coord/.append style={xshift=0,yshift=-25,color=black},
    	every axis plot/.append style={
          ybar,
          bar width=0.85,
          bar shift=0pt,
          draw=none
        },
        nodes near coords,
]
\addplot+ [mygray, error bars/.cd,
            y dir=both,
            y explicit,
            error bar style={black}]
        table
          [x=index,
            y=elo,
            y error plus expr=\thisrow{high_99_ci}-\thisrow{elo},
            y error minus expr=\thisrow{elo}-\thisrow{low_99_ci}
            ] {\datatableAGenAI};
\addplot+ [myblue, error bars/.cd,
            y dir=both,
            y explicit,
            error bar style={black}]
        table
          [x=index,
            y=elo,
            y error plus expr=\thisrow{high_99_ci}-\thisrow{elo},
            y error minus expr=\thisrow{elo}-\thisrow{low_99_ci}
            ] {\datatableBGenAI};     

\end{axis}
\end{tikzpicture}}
\end{minipage}
\hspace{2mm}
\begin{minipage}{0.35\linewidth}
\resizebox{\linewidth}{!}{%
\begin{tikzpicture}[scale=1.1]

\draw[very thin] (-0.5,5.5) rectangle +(1,1);
\draw[very thin, fill={rgb,1:red,0.996;green,0.878;blue,0.545}] (-0.5,4.5) rectangle +(1,1);
\node[black] at (0,5) {44.9\strut};
\draw[very thin, fill={rgb,1:red,0.993;green,0.717;blue,0.409}] (-0.5,3.5) rectangle +(1,1);
\node[black] at (0,4) {40.6\strut};
\draw[very thin, fill={rgb,1:red,0.993;green,0.732;blue,0.422}] (-0.5,2.5) rectangle +(1,1);
\node[black] at (0,3) {40.9\strut};
\draw[very thin, fill={rgb,1:red,0.958;green,0.437;blue,0.267}] (-0.5,1.5) rectangle +(1,1);
\node[black] at (0,2) {34.6\strut};
\draw[very thin, fill={rgb,1:red,0.647;green,0.000;blue,0.149}] (-0.5,0.5) rectangle +(1,1);
\node[white] at (0,1) {23.9\strut};
\draw[very thin, fill={rgb,1:red,0.851;green,0.937;blue,0.545}] (0.5,5.5) rectangle +(1,1);
\node[black] at (1,6) {55.1\strut};
\draw[very thin] (0.5,4.5) rectangle +(1,1);
\draw[very thin, fill={rgb,1:red,0.999;green,0.979;blue,0.713}] (0.5,3.5) rectangle +(1,1);
\node[black] at (1,4) {49.0\strut};
\draw[very thin, fill={rgb,1:red,0.996;green,0.883;blue,0.553}] (0.5,2.5) rectangle +(1,1);
\node[black] at (1,3) {45.1\strut};
\draw[very thin, fill={rgb,1:red,0.971;green,0.527;blue,0.309}] (0.5,1.5) rectangle +(1,1);
\node[black] at (1,2) {36.4\strut};
\draw[very thin, fill={rgb,1:red,0.778;green,0.125;blue,0.152}] (0.5,0.5) rectangle +(1,1);
\node[white] at (1,1) {27.5\strut};
\draw[very thin, fill={rgb,1:red,0.686;green,0.866;blue,0.439}] (1.5,5.5) rectangle +(1,1);
\node[black] at (2,6) {59.4\strut};
\draw[very thin, fill={rgb,1:red,0.974;green,0.989;blue,0.713}] (1.5,4.5) rectangle +(1,1);
\node[black] at (2,5) {51.0\strut};
\draw[very thin] (1.5,3.5) rectangle +(1,1);
\draw[very thin, fill={rgb,1:red,0.999;green,0.974;blue,0.705}] (1.5,2.5) rectangle +(1,1);
\node[black] at (2,3) {48.8\strut};
\draw[very thin, fill={rgb,1:red,0.993;green,0.702;blue,0.397}] (1.5,1.5) rectangle +(1,1);
\node[black] at (2,2) {40.2\strut};
\draw[very thin, fill={rgb,1:red,0.854;green,0.212;blue,0.164}] (1.5,0.5) rectangle +(1,1);
\node[white] at (2,1) {29.8\strut};
\draw[very thin, fill={rgb,1:red,0.702;green,0.873;blue,0.449}] (2.5,5.5) rectangle +(1,1);
\node[black] at (3,6) {59.1\strut};
\draw[very thin, fill={rgb,1:red,0.857;green,0.940;blue,0.553}] (2.5,4.5) rectangle +(1,1);
\node[black] at (3,5) {54.9\strut};
\draw[very thin, fill={rgb,1:red,0.968;green,0.986;blue,0.705}] (2.5,3.5) rectangle +(1,1);
\node[black] at (3,4) {51.2\strut};
\draw[very thin] (2.5,2.5) rectangle +(1,1);
\draw[very thin, fill={rgb,1:red,0.983;green,0.617;blue,0.350}] (2.5,1.5) rectangle +(1,1);
\node[black] at (3,2) {38.3\strut};
\draw[very thin, fill={rgb,1:red,0.903;green,0.315;blue,0.211}] (2.5,0.5) rectangle +(1,1);
\node[white] at (3,1) {32.0\strut};
\draw[very thin, fill={rgb,1:red,0.410;green,0.745;blue,0.389}] (3.5,5.5) rectangle +(1,1);
\node[black] at (4,6) {65.4\strut};
\draw[very thin, fill={rgb,1:red,0.498;green,0.784;blue,0.399}] (3.5,4.5) rectangle +(1,1);
\node[black] at (4,5) {63.6\strut};
\draw[very thin, fill={rgb,1:red,0.671;green,0.859;blue,0.428}] (3.5,3.5) rectangle +(1,1);
\node[black] at (4,4) {59.8\strut};
\draw[very thin, fill={rgb,1:red,0.587;green,0.823;blue,0.409}] (3.5,2.5) rectangle +(1,1);
\node[black] at (4,3) {61.7\strut};
\draw[very thin] (3.5,1.5) rectangle +(1,1);
\draw[very thin, fill={rgb,1:red,0.993;green,0.732;blue,0.422}] (3.5,0.5) rectangle +(1,1);
\node[black] at (4,1) {41.0\strut};
\draw[very thin, fill={rgb,1:red,0.000;green,0.408;blue,0.216}] (4.5,5.5) rectangle +(1,1);
\node[white] at (5,6) {76.1\strut};
\draw[very thin, fill={rgb,1:red,0.068;green,0.533;blue,0.281}] (4.5,4.5) rectangle +(1,1);
\node[white] at (5,5) {72.5\strut};
\draw[very thin, fill={rgb,1:red,0.131;green,0.610;blue,0.321}] (4.5,3.5) rectangle +(1,1);
\node[white] at (5,4) {70.2\strut};
\draw[very thin, fill={rgb,1:red,0.260;green,0.673;blue,0.353}] (4.5,2.5) rectangle +(1,1);
\node[white] at (5,3) {68.0\strut};
\draw[very thin, fill={rgb,1:red,0.702;green,0.873;blue,0.449}] (4.5,1.5) rectangle +(1,1);
\node[black] at (5,2) {59.0\strut};
\draw[very thin] (4.5,0.5) rectangle +(1,1);

\node[rotate=55, anchor=west, align=left] at (-0.1,6.6) {Imagen 3};
\node[anchor=east, align=right] at (-0.5,6) {Imagen 3};
\node[rotate=55, anchor=west, align=left] at (0.9,6.6) {SD 3};
\node[anchor=east, align=right] at (-0.5,5) {SD 3};
\node[rotate=55, anchor=west, align=left] at (1.9,6.6) {DALL·E 3};
\node[anchor=east, align=right] at (-0.5,4) {DALL·E 3};
\node[rotate=55, anchor=west, align=left] at (2.9,6.6) {MJ v6};
\node[anchor=east, align=right] at (-0.5,3) {MJ v6};
\node[rotate=55, anchor=west, align=left] at (3.9,6.6) {Imagen 2};
\node[anchor=east, align=right] at (-0.5,2) {Imagen 2};
\node[rotate=55, anchor=west, align=left] at (4.9,6.6) {SDXL 1};
\node[anchor=east, align=right] at (-0.5,1) {SDXL 1};

\node[anchor=east] at (0,-0.5) {};
\end{tikzpicture}}
\end{minipage}}}

\vspace{3mm}
\resizebox{1.01\linewidth}{!}{%
\mbox{%
\hspace{-3mm}%
\begin{minipage}{0.65\linewidth}
\resizebox{\linewidth}{!}{%
\begin{tikzpicture}
\begin{axis} [
        x label style={at={(0.5,-0.1)}},
        y label style={at={(-0.07,0.5)}},
        ylabel={Elo score (and 99\% CI)},
        title={Prompt-image alignment on \drawbench{}},
        width=15cm,
        height=10cm,
    	ybar,
    	major grid style={line width=.2pt,draw=gray!50,dashed},
    	ymajorgrids,
    	tick style={draw=none},
    	xticklabels from table={\modelnamesDrawbench}{model},
    	xtick={0,...,5},
    	xmin=-0.5,
    	xmax=5.5,
    	xticklabel style={yshift=0mm, align=center,rotate=0},
    	every node near coord/.append style={xshift=0,yshift=-25,color=black},
    	every axis plot/.append style={
          ybar,
          bar width=0.85,
          bar shift=0pt,
          draw=none
        },
        nodes near coords,
]
\addplot+ [mygray, error bars/.cd,
            y dir=both,
            y explicit,
            error bar style={black}]
        table
          [x=index,
            y=elo,
            y error plus expr=\thisrow{high_99_ci}-\thisrow{elo},
            y error minus expr=\thisrow{elo}-\thisrow{low_99_ci}
            ] {\datatableADrawbench};
\addplot+ [myblue, error bars/.cd,
            y dir=both,
            y explicit,
            error bar style={black}]
        table
          [x=index,
            y=elo,
            y error plus expr=\thisrow{high_99_ci}-\thisrow{elo},
            y error minus expr=\thisrow{elo}-\thisrow{low_99_ci}
            ] {\datatableBDrawbench};

\end{axis}
\end{tikzpicture}}
\end{minipage}
\hspace{2mm}
\begin{minipage}{0.35\linewidth}
\resizebox{\linewidth}{!}{%
\begin{tikzpicture}[scale=1.1]

\draw[very thin] (-0.5,5.5) rectangle +(1,1);
\draw[very thin, fill={rgb,1:red,0.997;green,0.907;blue,0.593}] (-0.5,4.5) rectangle +(1,1);
\node[black] at (0,5) {47.4\strut};
\draw[very thin, fill={rgb,1:red,0.993;green,0.725;blue,0.416}] (-0.5,3.5) rectangle +(1,1);
\node[black] at (0,4) {44.0\strut};
\draw[very thin, fill={rgb,1:red,0.976;green,0.567;blue,0.327}] (-0.5,2.5) rectangle +(1,1);
\node[black] at (0,3) {41.7\strut};
\draw[very thin, fill={rgb,1:red,0.670;green,0.022;blue,0.149}] (-0.5,1.5) rectangle +(1,1);
\node[white] at (0,2) {33.5\strut};
\draw[very thin, fill={rgb,1:red,0.755;green,0.103;blue,0.151}] (-0.5,0.5) rectangle +(1,1);
\node[white] at (0,1) {34.9\strut};
\draw[very thin, fill={rgb,1:red,0.886;green,0.952;blue,0.593}] (0.5,5.5) rectangle +(1,1);
\node[black] at (1,6) {52.6\strut};
\draw[very thin] (0.5,4.5) rectangle +(1,1);
\draw[very thin, fill={rgb,1:red,0.994;green,0.771;blue,0.455}] (0.5,3.5) rectangle +(1,1);
\node[black] at (1,4) {44.8\strut};
\draw[very thin, fill={rgb,1:red,0.997;green,0.898;blue,0.577}] (0.5,2.5) rectangle +(1,1);
\node[black] at (1,3) {47.2\strut};
\draw[very thin, fill={rgb,1:red,0.693;green,0.044;blue,0.150}] (0.5,1.5) rectangle +(1,1);
\node[white] at (1,2) {33.9\strut};
\draw[very thin, fill={rgb,1:red,0.647;green,0.000;blue,0.149}] (0.5,0.5) rectangle +(1,1);
\node[white] at (1,1) {33.0\strut};
\draw[very thin, fill={rgb,1:red,0.694;green,0.870;blue,0.444}] (1.5,5.5) rectangle +(1,1);
\node[black] at (2,6) {56.0\strut};
\draw[very thin, fill={rgb,1:red,0.741;green,0.890;blue,0.474}] (1.5,4.5) rectangle +(1,1);
\node[black] at (2,5) {55.2\strut};
\draw[very thin] (1.5,3.5) rectangle +(1,1);
\draw[very thin, fill={rgb,1:red,0.997;green,0.999;blue,0.745}] (1.5,2.5) rectangle +(1,1);
\node[black] at (2,3) {50.1\strut};
\draw[very thin, fill={rgb,1:red,0.975;green,0.557;blue,0.323}] (1.5,1.5) rectangle +(1,1);
\node[black] at (2,2) {41.5\strut};
\draw[very thin, fill={rgb,1:red,0.903;green,0.315;blue,0.211}] (1.5,0.5) rectangle +(1,1);
\node[white] at (2,1) {38.2\strut};
\draw[very thin, fill={rgb,1:red,0.538;green,0.801;blue,0.403}] (2.5,5.5) rectangle +(1,1);
\node[black] at (3,6) {58.3\strut};
\draw[very thin, fill={rgb,1:red,0.874;green,0.947;blue,0.577}] (2.5,4.5) rectangle +(1,1);
\node[black] at (3,5) {52.8\strut};
\draw[very thin, fill={rgb,1:red,1.000;green,0.998;blue,0.745}] (2.5,3.5) rectangle +(1,1);
\node[black] at (3,4) {49.9\strut};
\draw[very thin] (2.5,2.5) rectangle +(1,1);
\draw[very thin, fill={rgb,1:red,0.961;green,0.457;blue,0.277}] (2.5,1.5) rectangle +(1,1);
\node[black] at (3,2) {40.3\strut};
\draw[very thin, fill={rgb,1:red,0.935;green,0.381;blue,0.241}] (2.5,0.5) rectangle +(1,1);
\node[white] at (3,1) {39.2\strut};
\draw[very thin, fill={rgb,1:red,0.012;green,0.430;blue,0.227}] (3.5,5.5) rectangle +(1,1);
\node[white] at (4,6) {66.5\strut};
\draw[very thin, fill={rgb,1:red,0.024;green,0.452;blue,0.239}] (3.5,4.5) rectangle +(1,1);
\node[white] at (4,5) {66.1\strut};
\draw[very thin, fill={rgb,1:red,0.518;green,0.793;blue,0.401}] (3.5,3.5) rectangle +(1,1);
\node[black] at (4,4) {58.5\strut};
\draw[very thin, fill={rgb,1:red,0.430;green,0.754;blue,0.391}] (3.5,2.5) rectangle +(1,1);
\node[black] at (4,3) {59.7\strut};
\draw[very thin] (3.5,1.5) rectangle +(1,1);
\draw[very thin, fill={rgb,1:red,0.944;green,0.977;blue,0.673}] (3.5,0.5) rectangle +(1,1);
\node[black] at (4,1) {51.2\strut};
\draw[very thin, fill={rgb,1:red,0.056;green,0.511;blue,0.270}] (4.5,5.5) rectangle +(1,1);
\node[white] at (5,6) {65.1\strut};
\draw[very thin, fill={rgb,1:red,0.000;green,0.408;blue,0.216}] (4.5,4.5) rectangle +(1,1);
\node[white] at (5,5) {67.0\strut};
\draw[very thin, fill={rgb,1:red,0.260;green,0.673;blue,0.353}] (4.5,3.5) rectangle +(1,1);
\node[white] at (5,4) {61.8\strut};
\draw[very thin, fill={rgb,1:red,0.342;green,0.713;blue,0.374}] (4.5,2.5) rectangle +(1,1);
\node[white] at (5,3) {60.8\strut};
\draw[very thin, fill={rgb,1:red,0.999;green,0.955;blue,0.673}] (4.5,1.5) rectangle +(1,1);
\node[black] at (5,2) {48.8\strut};
\draw[very thin] (4.5,0.5) rectangle +(1,1);

\node[rotate=55, anchor=west, align=left] at (-0.1,6.6) {Imagen 3};
\node[anchor=east, align=right] at (-0.5,6) {Imagen 3};
\node[rotate=55, anchor=west, align=left] at (0.9,6.6) {SD 3};
\node[anchor=east, align=right] at (-0.5,5) {SD 3};
\node[rotate=55, anchor=west, align=left] at (1.9,6.6) {DALL·E 3};
\node[anchor=east, align=right] at (-0.5,4) {DALL·E 3};
\node[rotate=55, anchor=west, align=left] at (2.9,6.6) {MJ v6};
\node[anchor=east, align=right] at (-0.5,3) {MJ v6};
\node[rotate=55, anchor=west, align=left] at (3.9,6.6) {Imagen 2};
\node[anchor=east, align=right] at (-0.5,2) {Imagen 2};
\node[rotate=55, anchor=west, align=left] at (4.9,6.6) {SDXL 1};
\node[anchor=east, align=right] at (-0.5,1) {SDXL 1};

\node[anchor=east] at (0,-0.5) {};
\end{tikzpicture}}
\end{minipage}}}

\vspace{3mm}
\resizebox{1.01\linewidth}{!}{%
\mbox{%
\hspace{-3mm}%
\begin{minipage}{0.65\linewidth}
\resizebox{\linewidth}{!}{%
\begin{tikzpicture}
\begin{axis} [
        x label style={at={(0.5,-0.1)}},
        y label style={at={(-0.07,0.5)}},
        ylabel={Elo score (and 99\% CI)},
        title={Prompt-image alignment on \dalle~Eval},
        width=15cm,
        height=10cm,
    	ybar,
    	major grid style={line width=.2pt,draw=gray!50,dashed},
    	ymajorgrids,
    	tick style={draw=none},
    	xticklabels from table={\modelnamesDalle}{model},
    	xtick={0,...,5},
    	xmin=-0.5,
    	xmax=5.5,
    	xticklabel style={yshift=0mm, align=center,rotate=0},
    	every node near coord/.append style={xshift=0,yshift=-25,color=black},
    	every axis plot/.append style={
          ybar,
          bar width=0.85,
          bar shift=0pt,
          draw=none
        },
        nodes near coords,
]
\addplot+ [mygray, error bars/.cd,
            y dir=both,
            y explicit,
            error bar style={black}]
        table
          [x=index,
            y=elo,
            y error plus expr=\thisrow{high_99_ci}-\thisrow{elo},
            y error minus expr=\thisrow{elo}-\thisrow{low_99_ci}
            ] {\datatableADalle};
\addplot+ [myblue, error bars/.cd,
            y dir=both,
            y explicit,
            error bar style={black}]
        table
          [x=index,
            y=elo,
            y error plus expr=\thisrow{high_99_ci}-\thisrow{elo},
            y error minus expr=\thisrow{elo}-\thisrow{low_99_ci}
            ] {\datatableBDalle};     

\end{axis}
\end{tikzpicture}}
\end{minipage}
\hspace{2mm}
\begin{minipage}{0.35\linewidth}
\resizebox{\linewidth}{!}{%
\begin{tikzpicture}[scale=1.1]

\draw[very thin] (-0.5,5.5) rectangle +(1,1);
\draw[very thin, fill={rgb,1:red,1.000;green,0.998;blue,0.745}] (-0.5,4.5) rectangle +(1,1);
\node[black] at (0,5) {49.8\strut};
\draw[very thin, fill={rgb,1:red,0.998;green,0.950;blue,0.665}] (-0.5,3.5) rectangle +(1,1);
\node[black] at (0,4) {47.7\strut};
\draw[very thin, fill={rgb,1:red,0.998;green,0.936;blue,0.641}] (-0.5,2.5) rectangle +(1,1);
\node[black] at (0,3) {47.0\strut};
\draw[very thin, fill={rgb,1:red,0.850;green,0.202;blue,0.159}] (-0.5,1.5) rectangle +(1,1);
\node[white] at (0,2) {28.2\strut};
\draw[very thin, fill={rgb,1:red,0.647;green,0.000;blue,0.149}] (-0.5,0.5) rectangle +(1,1);
\node[white] at (0,1) {22.3\strut};
\draw[very thin, fill={rgb,1:red,0.997;green,0.999;blue,0.745}] (0.5,5.5) rectangle +(1,1);
\node[black] at (1,6) {50.2\strut};
\draw[very thin] (0.5,4.5) rectangle +(1,1);
\draw[very thin, fill={rgb,1:red,1.000;green,0.998;blue,0.745}] (0.5,3.5) rectangle +(1,1);
\node[black] at (1,4) {49.9\strut};
\draw[very thin, fill={rgb,1:red,0.999;green,0.969;blue,0.697}] (0.5,2.5) rectangle +(1,1);
\node[black] at (1,3) {48.7\strut};
\draw[very thin, fill={rgb,1:red,0.739;green,0.089;blue,0.151}] (0.5,1.5) rectangle +(1,1);
\node[white] at (1,2) {25.0\strut};
\draw[very thin, fill={rgb,1:red,0.762;green,0.111;blue,0.151}] (0.5,0.5) rectangle +(1,1);
\node[white] at (1,1) {25.6\strut};
\draw[very thin, fill={rgb,1:red,0.939;green,0.974;blue,0.665}] (1.5,5.5) rectangle +(1,1);
\node[black] at (2,6) {52.3\strut};
\draw[very thin, fill={rgb,1:red,0.997;green,0.999;blue,0.745}] (1.5,4.5) rectangle +(1,1);
\node[black] at (2,5) {50.1\strut};
\draw[very thin] (1.5,3.5) rectangle +(1,1);
\draw[very thin, fill={rgb,1:red,0.999;green,0.955;blue,0.673}] (1.5,2.5) rectangle +(1,1);
\node[black] at (2,3) {47.9\strut};
\draw[very thin, fill={rgb,1:red,0.809;green,0.155;blue,0.152}] (1.5,1.5) rectangle +(1,1);
\node[white] at (2,2) {27.0\strut};
\draw[very thin, fill={rgb,1:red,0.670;green,0.022;blue,0.149}] (1.5,0.5) rectangle +(1,1);
\node[white] at (2,1) {23.0\strut};
\draw[very thin, fill={rgb,1:red,0.921;green,0.967;blue,0.641}] (2.5,5.5) rectangle +(1,1);
\node[black] at (3,6) {53.0\strut};
\draw[very thin, fill={rgb,1:red,0.962;green,0.984;blue,0.697}] (2.5,4.5) rectangle +(1,1);
\node[black] at (3,5) {51.3\strut};
\draw[very thin, fill={rgb,1:red,0.944;green,0.977;blue,0.673}] (2.5,3.5) rectangle +(1,1);
\node[black] at (3,4) {52.1\strut};
\draw[very thin] (2.5,2.5) rectangle +(1,1);
\draw[very thin, fill={rgb,1:red,0.917;green,0.343;blue,0.224}] (2.5,1.5) rectangle +(1,1);
\node[white] at (3,2) {31.5\strut};
\draw[very thin, fill={rgb,1:red,0.890;green,0.287;blue,0.198}] (2.5,0.5) rectangle +(1,1);
\node[white] at (3,1) {30.1\strut};
\draw[very thin, fill={rgb,1:red,0.119;green,0.605;blue,0.318}] (3.5,5.5) rectangle +(1,1);
\node[white] at (4,6) {71.8\strut};
\draw[very thin, fill={rgb,1:red,0.048;green,0.496;blue,0.262}] (3.5,4.5) rectangle +(1,1);
\node[white] at (4,5) {75.0\strut};
\draw[very thin, fill={rgb,1:red,0.084;green,0.563;blue,0.296}] (3.5,3.5) rectangle +(1,1);
\node[white] at (4,4) {73.0\strut};
\draw[very thin, fill={rgb,1:red,0.295;green,0.690;blue,0.362}] (3.5,2.5) rectangle +(1,1);
\node[white] at (4,3) {68.5\strut};
\draw[very thin] (3.5,1.5) rectangle +(1,1);
\draw[very thin, fill={rgb,1:red,0.998;green,0.950;blue,0.665}] (3.5,0.5) rectangle +(1,1);
\node[black] at (4,1) {47.7\strut};
\draw[very thin, fill={rgb,1:red,0.000;green,0.408;blue,0.216}] (4.5,5.5) rectangle +(1,1);
\node[white] at (5,6) {77.7\strut};
\draw[very thin, fill={rgb,1:red,0.060;green,0.519;blue,0.273}] (4.5,4.5) rectangle +(1,1);
\node[white] at (5,5) {74.4\strut};
\draw[very thin, fill={rgb,1:red,0.012;green,0.430;blue,0.227}] (4.5,3.5) rectangle +(1,1);
\node[white] at (5,4) {77.0\strut};
\draw[very thin, fill={rgb,1:red,0.225;green,0.656;blue,0.344}] (4.5,2.5) rectangle +(1,1);
\node[white] at (5,3) {69.9\strut};
\draw[very thin, fill={rgb,1:red,0.939;green,0.974;blue,0.665}] (4.5,1.5) rectangle +(1,1);
\node[black] at (5,2) {52.3\strut};
\draw[very thin] (4.5,0.5) rectangle +(1,1);

\node[rotate=55, anchor=west, align=left] at (-0.1,6.6) {Imagen 3};
\node[anchor=east, align=right] at (-0.5,6) {Imagen 3};
\node[rotate=55, anchor=west, align=left] at (0.9,6.6) {DALL·E 3};
\node[anchor=east, align=right] at (-0.5,5) {DALL·E 3};
\node[rotate=55, anchor=west, align=left] at (1.9,6.6) {SD 3};
\node[anchor=east, align=right] at (-0.5,4) {SD 3};
\node[rotate=55, anchor=west, align=left] at (2.9,6.6) {MJ v6};
\node[anchor=east, align=right] at (-0.5,3) {MJ v6};
\node[rotate=55, anchor=west, align=left] at (3.9,6.6) {Imagen 2};
\node[anchor=east, align=right] at (-0.5,2) {Imagen 2};
\node[rotate=55, anchor=west, align=left] at (4.9,6.6) {SDXL 1};
\node[anchor=east, align=right] at (-0.5,1) {SDXL 1};

\node[anchor=east] at (0,-0.5) {};
\end{tikzpicture}}
\end{minipage}}}

\caption{{\bf Prompt-Image Alignment:} Elo scores and win-rate percentages on \genaibench{}, \drawbench{}, and \dalle~Eval. Please refer to Appendix~\ref{app:imagen3-002} for updated human evaluation results as of December 2024.}
\label{fig:elo_t2i_alignment}
\end{figure}

%% file: assets/human_evals/elo_visual_appeal.tikz
\pgfplotstableread[col sep=comma]{
index,model,elo,low_99_ci,high_99_ci
0,SDXL 1,821,812,831
1,Imagen 2,943,935,950
2,DALL·E 3,969,961,977
3,SD 3,1072,1064,1079
4,Imagen 3,1095,1088,1101
5,MJ v6,1101,1093,1109
}\modelnamesGenAI

\pgfplotstableread[col sep=comma]{
index,model,elo,low_99_ci,high_99_ci
0,SDXL 1,821,812,831
2,DALL·E 3,969,961,977
3,SD 3,1072,1064,1079
5,MJ v6,1101,1093,1109
}\datatableAGenAI

\pgfplotstableread[col sep=comma]{
index,model,elo,low_99_ci,high_99_ci
1,Imagen 2,943,935,950
4,Imagen 3,1095,1088,1101
}\datatableBGenAI

\pgfplotstableread[col sep=comma]{
index,model,elo,low_99_ci,high_99_ci
0,DALL·E 3,906,897,914
1,SDXL 1,934,926,942
2,Imagen 2,994,986,1001
3,SD 3,1029,1021,1037
4,Imagen 3,1063,1055,1071
5,MJ v6,1075,1067,1083
}\modelnamesDrawbench

\pgfplotstableread[col sep=comma]{
index,model,elo,low_99_ci,high_99_ci
0,DALL·E 3,906,897,914
1,SDXL 1,934,926,942
3,SD 3,1029,1021,1037
5,MJ v6,1075,1067,1083
}\datatableADrawbench

\pgfplotstableread[col sep=comma]{
index,model,elo,low_99_ci,high_99_ci
2,Imagen 2,994,986,1001
4,Imagen 3,1063,1055,1071
}\datatableBDrawbench

\pgfplotstableread[col sep=comma]{
index,model,elo,low_99_ci,high_99_ci
0,Imagen 2,910,902,919
1,SDXL 1,922,914,930
2,DALL·E 3,1001,994,1009
3,SD 3,1024,1016,1032
4,Imagen 3,1047,1039,1055
5,MJ v6,1095,1087,1104
}\modelnamesDalle

\pgfplotstableread[col sep=comma]{
index,model,elo,low_99_ci,high_99_ci
1,SDXL 1,922,914,930
2,DALL·E 3,1001,994,1009
3,SD 3,1024,1016,1032
5,MJ v6,1095,1087,1104
}\datatableADalle

\pgfplotstableread[col sep=comma]{
index,model,elo,low_99_ci,high_99_ci
0,Imagen 2,910,902,919
4,Imagen 3,1047,1039,1055
}\datatableBDalle

\begin{figure}
\centering
\resizebox{1.01\linewidth}{!}{%
\mbox{%
\hspace{-3mm}%
\begin{minipage}{0.65\linewidth}
\resizebox{\linewidth}{!}{%
\begin{tikzpicture}
\begin{axis} [
        x label style={at={(0.5,-0.1)}},
        y label style={at={(-0.07,0.5)}},
        ylabel={Elo score (and 99\% CI)},
        title={Visual appeal on \genaibench{}},
        width=15cm,
        height=10cm,
    	ybar,
    	major grid style={line width=.2pt,draw=gray!50,dashed},
    	ymajorgrids,
    	tick style={draw=none},
    	xticklabels from table={\modelnamesGenAI}{model},
    	xtick={0,...,5},
    	xmin=-0.5,
    	xmax=5.5,
    	xticklabel style={yshift=0mm, align=center,rotate=0},
    	every node near coord/.append style={xshift=0,yshift=-25,color=black},
    	every axis plot/.append style={
          ybar,
          bar width=0.85,
          bar shift=0pt,
          draw=none
        },
        nodes near coords,
]
\addplot+ [mygray, error bars/.cd,
            y dir=both,
            y explicit,
            error bar style={black}]
        table
          [x=index,
            y=elo,
            y error plus expr=\thisrow{high_99_ci}-\thisrow{elo},
            y error minus expr=\thisrow{elo}-\thisrow{low_99_ci}
            ] {\datatableAGenAI};
\addplot+ [myblue, error bars/.cd,
            y dir=both,
            y explicit,
            error bar style={black}]
        table
          [x=index,
            y=elo,
            y error plus expr=\thisrow{high_99_ci}-\thisrow{elo},
            y error minus expr=\thisrow{elo}-\thisrow{low_99_ci}
            ] {\datatableBGenAI};     

\end{axis}
\end{tikzpicture}}
\end{minipage}
\hspace{2mm}
\begin{minipage}{0.35\linewidth}
\resizebox{\linewidth}{!}{%
\begin{tikzpicture}[scale=1.1]

\draw[very thin] (-0.5,5.5) rectangle +(1,1);
\draw[very thin, fill={rgb,1:red,0.999;green,0.964;blue,0.689}] (-0.5,4.5) rectangle +(1,1);
\node[black] at (0,5) {48.4\strut};
\draw[very thin, fill={rgb,1:red,0.997;green,0.912;blue,0.601}] (-0.5,3.5) rectangle +(1,1);
\node[black] at (0,4) {46.0\strut};
\draw[very thin, fill={rgb,1:red,0.964;green,0.477;blue,0.286}] (-0.5,2.5) rectangle +(1,1);
\node[black] at (0,3) {34.1\strut};
\draw[very thin, fill={rgb,1:red,0.973;green,0.547;blue,0.318}] (-0.5,1.5) rectangle +(1,1);
\node[black] at (0,2) {35.6\strut};
\draw[very thin, fill={rgb,1:red,0.647;green,0.000;blue,0.149}] (-0.5,0.5) rectangle +(1,1);
\node[white] at (0,1) {21.6\strut};
\draw[very thin, fill={rgb,1:red,0.956;green,0.982;blue,0.689}] (0.5,5.5) rectangle +(1,1);
\node[black] at (1,6) {51.6\strut};
\draw[very thin] (0.5,4.5) rectangle +(1,1);
\draw[very thin, fill={rgb,1:red,0.999;green,0.969;blue,0.697}] (0.5,3.5) rectangle +(1,1);
\node[black] at (1,4) {48.5\strut};
\draw[very thin, fill={rgb,1:red,0.989;green,0.657;blue,0.369}] (0.5,2.5) rectangle +(1,1);
\node[black] at (1,3) {38.1\strut};
\draw[very thin, fill={rgb,1:red,0.952;green,0.418;blue,0.258}] (0.5,1.5) rectangle +(1,1);
\node[white] at (1,2) {32.7\strut};
\draw[very thin, fill={rgb,1:red,0.709;green,0.059;blue,0.150}] (0.5,0.5) rectangle +(1,1);
\node[white] at (1,1) {23.5\strut};
\draw[very thin, fill={rgb,1:red,0.892;green,0.954;blue,0.601}] (1.5,5.5) rectangle +(1,1);
\node[black] at (2,6) {54.0\strut};
\draw[very thin, fill={rgb,1:red,0.962;green,0.984;blue,0.697}] (1.5,4.5) rectangle +(1,1);
\node[black] at (2,5) {51.5\strut};
\draw[very thin] (1.5,3.5) rectangle +(1,1);
\draw[very thin, fill={rgb,1:red,0.992;green,0.686;blue,0.384}] (1.5,2.5) rectangle +(1,1);
\node[black] at (2,3) {38.8\strut};
\draw[very thin, fill={rgb,1:red,0.987;green,0.647;blue,0.364}] (1.5,1.5) rectangle +(1,1);
\node[black] at (2,2) {37.8\strut};
\draw[very thin, fill={rgb,1:red,0.859;green,0.221;blue,0.168}] (1.5,0.5) rectangle +(1,1);
\node[white] at (2,1) {28.2\strut};
\draw[very thin, fill={rgb,1:red,0.449;green,0.763;blue,0.394}] (2.5,5.5) rectangle +(1,1);
\node[black] at (3,6) {65.9\strut};
\draw[very thin, fill={rgb,1:red,0.626;green,0.840;blue,0.413}] (2.5,4.5) rectangle +(1,1);
\node[black] at (3,5) {61.9\strut};
\draw[very thin, fill={rgb,1:red,0.655;green,0.853;blue,0.418}] (2.5,3.5) rectangle +(1,1);
\node[black] at (3,4) {61.2\strut};
\draw[very thin] (2.5,2.5) rectangle +(1,1);
\draw[very thin, fill={rgb,1:red,0.997;green,0.912;blue,0.601}] (2.5,1.5) rectangle +(1,1);
\node[black] at (3,2) {45.9\strut};
\draw[very thin, fill={rgb,1:red,0.868;green,0.240;blue,0.177}] (2.5,0.5) rectangle +(1,1);
\node[white] at (3,1) {28.6\strut};
\draw[very thin, fill={rgb,1:red,0.518;green,0.793;blue,0.401}] (3.5,5.5) rectangle +(1,1);
\node[black] at (4,6) {64.4\strut};
\draw[very thin, fill={rgb,1:red,0.388;green,0.735;blue,0.385}] (3.5,4.5) rectangle +(1,1);
\node[white] at (4,5) {67.3\strut};
\draw[very thin, fill={rgb,1:red,0.617;green,0.836;blue,0.412}] (3.5,3.5) rectangle +(1,1);
\node[black] at (4,4) {62.2\strut};
\draw[very thin, fill={rgb,1:red,0.892;green,0.954;blue,0.601}] (3.5,2.5) rectangle +(1,1);
\node[black] at (4,3) {54.1\strut};
\draw[very thin] (3.5,1.5) rectangle +(1,1);
\draw[very thin, fill={rgb,1:red,0.993;green,0.709;blue,0.403}] (3.5,0.5) rectangle +(1,1);
\node[black] at (4,1) {39.5\strut};
\draw[very thin, fill={rgb,1:red,0.000;green,0.408;blue,0.216}] (4.5,5.5) rectangle +(1,1);
\node[white] at (5,6) {78.4\strut};
\draw[very thin, fill={rgb,1:red,0.032;green,0.467;blue,0.246}] (4.5,4.5) rectangle +(1,1);
\node[white] at (5,5) {76.5\strut};
\draw[very thin, fill={rgb,1:red,0.143;green,0.616;blue,0.324}] (4.5,3.5) rectangle +(1,1);
\node[white] at (5,4) {71.8\strut};
\draw[very thin, fill={rgb,1:red,0.166;green,0.627;blue,0.330}] (4.5,2.5) rectangle +(1,1);
\node[white] at (5,3) {71.4\strut};
\draw[very thin, fill={rgb,1:red,0.678;green,0.863;blue,0.433}] (4.5,1.5) rectangle +(1,1);
\node[black] at (5,2) {60.5\strut};
\draw[very thin] (4.5,0.5) rectangle +(1,1);

\node[rotate=55, anchor=west, align=left] at (-0.1,6.6) {MJ v6};
\node[anchor=east, align=right] at (-0.5,6) {MJ v6};
\node[rotate=55, anchor=west, align=left] at (0.9,6.6) {Imagen 3};
\node[anchor=east, align=right] at (-0.5,5) {Imagen 3};
\node[rotate=55, anchor=west, align=left] at (1.9,6.6) {SD 3};
\node[anchor=east, align=right] at (-0.5,4) {SD 3};
\node[rotate=55, anchor=west, align=left] at (2.9,6.6) {DALL·E 3};
\node[anchor=east, align=right] at (-0.5,3) {DALL·E 3};
\node[rotate=55, anchor=west, align=left] at (3.9,6.6) {Imagen 2};
\node[anchor=east, align=right] at (-0.5,2) {Imagen 2};
\node[rotate=55, anchor=west, align=left] at (4.9,6.6) {SDXL 1};
\node[anchor=east, align=right] at (-0.5,1) {SDXL 1};

\node[anchor=east] at (0,-0.5) {};
\end{tikzpicture}}
\end{minipage}}}

\vspace{3mm}
\resizebox{1.01\linewidth}{!}{%
\mbox{%
\hspace{-3mm}%
\begin{minipage}{0.65\linewidth}
\resizebox{\linewidth}{!}{%
\begin{tikzpicture}
\begin{axis} [
        x label style={at={(0.5,-0.1)}},
        y label style={at={(-0.07,0.5)}},
        ylabel={Elo score (and 99\% CI)},
        title={Visual appeal on \drawbench},
        width=15cm,
        height=10cm,
    	ybar,
    	major grid style={line width=.2pt,draw=gray!50,dashed},
    	ymajorgrids,
    	tick style={draw=none},
    	xticklabels from table={\modelnamesDrawbench}{model},
    	xtick={0,...,5},
    	xmin=-0.5,
    	xmax=5.5,
    	xticklabel style={yshift=0mm, align=center,rotate=0},
    	every node near coord/.append style={xshift=0,yshift=-25,color=black},
    	every axis plot/.append style={
          ybar,
          bar width=0.85,
          bar shift=0pt,
          draw=none
        },
        nodes near coords,
]
\addplot+ [mygray, error bars/.cd,
            y dir=both,
            y explicit,
            error bar style={black}]
        table
          [x=index,
            y=elo,
            y error plus expr=\thisrow{high_99_ci}-\thisrow{elo},
            y error minus expr=\thisrow{elo}-\thisrow{low_99_ci}
            ] {\datatableADrawbench};
\addplot+ [myblue, error bars/.cd,
            y dir=both,
            y explicit,
            error bar style={black}]
        table
          [x=index,
            y=elo,
            y error plus expr=\thisrow{high_99_ci}-\thisrow{elo},
            y error minus expr=\thisrow{elo}-\thisrow{low_99_ci}
            ] {\datatableBDrawbench};     

\end{axis}
\end{tikzpicture}}
\end{minipage}
\hspace{2mm}
\begin{minipage}{0.35\linewidth}
\resizebox{\linewidth}{!}{%
\begin{tikzpicture}[scale=1.1]

\draw[very thin] (-0.5,5.5) rectangle +(1,1);
\draw[very thin, fill={rgb,1:red,0.996;green,0.888;blue,0.561}] (-0.5,4.5) rectangle +(1,1);
\node[black] at (0,5) {46.1\strut};
\draw[very thin, fill={rgb,1:red,0.996;green,0.855;blue,0.526}] (-0.5,3.5) rectangle +(1,1);
\node[black] at (0,4) {45.2\strut};
\draw[very thin, fill={rgb,1:red,0.983;green,0.617;blue,0.350}] (-0.5,2.5) rectangle +(1,1);
\node[black] at (0,3) {40.4\strut};
\draw[very thin, fill={rgb,1:red,0.903;green,0.315;blue,0.211}] (-0.5,1.5) rectangle +(1,1);
\node[white] at (0,2) {35.3\strut};
\draw[very thin, fill={rgb,1:red,0.647;green,0.000;blue,0.149}] (-0.5,0.5) rectangle +(1,1);
\node[white] at (0,1) {28.7\strut};
\draw[very thin, fill={rgb,1:red,0.863;green,0.942;blue,0.561}] (0.5,5.5) rectangle +(1,1);
\node[black] at (1,6) {53.9\strut};
\draw[very thin] (0.5,4.5) rectangle +(1,1);
\draw[very thin, fill={rgb,1:red,0.993;green,0.748;blue,0.435}] (0.5,3.5) rectangle +(1,1);
\node[black] at (1,4) {42.9\strut};
\draw[very thin, fill={rgb,1:red,0.973;green,0.547;blue,0.318}] (0.5,2.5) rectangle +(1,1);
\node[black] at (1,3) {39.2\strut};
\draw[very thin, fill={rgb,1:red,0.863;green,0.230;blue,0.172}] (0.5,1.5) rectangle +(1,1);
\node[white] at (1,2) {33.8\strut};
\draw[very thin, fill={rgb,1:red,0.809;green,0.155;blue,0.152}] (0.5,0.5) rectangle +(1,1);
\node[white] at (1,1) {32.2\strut};
\draw[very thin, fill={rgb,1:red,0.820;green,0.924;blue,0.525}] (1.5,5.5) rectangle +(1,1);
\node[black] at (2,6) {54.8\strut};
\draw[very thin, fill={rgb,1:red,0.710;green,0.876;blue,0.454}] (1.5,4.5) rectangle +(1,1);
\node[black] at (2,5) {57.1\strut};
\draw[very thin] (1.5,3.5) rectangle +(1,1);
\draw[very thin, fill={rgb,1:red,0.997;green,0.917;blue,0.609}] (1.5,2.5) rectangle +(1,1);
\node[black] at (2,3) {47.1\strut};
\draw[very thin, fill={rgb,1:red,0.930;green,0.371;blue,0.237}] (1.5,1.5) rectangle +(1,1);
\node[white] at (2,2) {36.2\strut};
\draw[very thin, fill={rgb,1:red,0.877;green,0.259;blue,0.185}] (1.5,0.5) rectangle +(1,1);
\node[white] at (2,1) {34.2\strut};
\draw[very thin, fill={rgb,1:red,0.577;green,0.819;blue,0.408}] (2.5,5.5) rectangle +(1,1);
\node[black] at (3,6) {59.6\strut};
\draw[very thin, fill={rgb,1:red,0.508;green,0.789;blue,0.400}] (2.5,4.5) rectangle +(1,1);
\node[black] at (3,5) {60.8\strut};
\draw[very thin, fill={rgb,1:red,0.898;green,0.957;blue,0.609}] (2.5,3.5) rectangle +(1,1);
\node[black] at (3,4) {52.9\strut};
\draw[very thin] (2.5,2.5) rectangle +(1,1);
\draw[very thin, fill={rgb,1:red,0.992;green,0.694;blue,0.390}] (2.5,1.5) rectangle +(1,1);
\node[black] at (3,2) {41.7\strut};
\draw[very thin, fill={rgb,1:red,0.968;green,0.507;blue,0.300}] (2.5,0.5) rectangle +(1,1);
\node[black] at (3,1) {38.6\strut};
\draw[very thin, fill={rgb,1:red,0.260;green,0.673;blue,0.353}] (3.5,5.5) rectangle +(1,1);
\node[white] at (4,6) {64.7\strut};
\draw[very thin, fill={rgb,1:red,0.155;green,0.622;blue,0.327}] (3.5,4.5) rectangle +(1,1);
\node[white] at (4,5) {66.2\strut};
\draw[very thin, fill={rgb,1:red,0.318;green,0.701;blue,0.368}] (3.5,3.5) rectangle +(1,1);
\node[white] at (4,4) {63.8\strut};
\draw[very thin, fill={rgb,1:red,0.655;green,0.853;blue,0.418}] (3.5,2.5) rectangle +(1,1);
\node[black] at (4,3) {58.3\strut};
\draw[very thin] (3.5,1.5) rectangle +(1,1);
\draw[very thin, fill={rgb,1:red,0.996;green,0.888;blue,0.561}] (3.5,0.5) rectangle +(1,1);
\node[black] at (4,1) {46.0\strut};
\draw[very thin, fill={rgb,1:red,0.000;green,0.408;blue,0.216}] (4.5,5.5) rectangle +(1,1);
\node[white] at (5,6) {71.2\strut};
\draw[very thin, fill={rgb,1:red,0.080;green,0.555;blue,0.293}] (4.5,4.5) rectangle +(1,1);
\node[white] at (5,5) {67.8\strut};
\draw[very thin, fill={rgb,1:red,0.178;green,0.633;blue,0.333}] (4.5,3.5) rectangle +(1,1);
\node[white] at (5,4) {65.8\strut};
\draw[very thin, fill={rgb,1:red,0.479;green,0.776;blue,0.397}] (4.5,2.5) rectangle +(1,1);
\node[black] at (5,3) {61.4\strut};
\draw[very thin, fill={rgb,1:red,0.857;green,0.940;blue,0.553}] (4.5,1.5) rectangle +(1,1);
\node[black] at (5,2) {54.0\strut};
\draw[very thin] (4.5,0.5) rectangle +(1,1);

\node[rotate=55, anchor=west, align=left] at (-0.1,6.6) {MJ v6};
\node[anchor=east, align=right] at (-0.5,6) {MJ v6};
\node[rotate=55, anchor=west, align=left] at (0.9,6.6) {Imagen 3};
\node[anchor=east, align=right] at (-0.5,5) {Imagen 3};
\node[rotate=55, anchor=west, align=left] at (1.9,6.6) {SD 3};
\node[anchor=east, align=right] at (-0.5,4) {SD 3};
\node[rotate=55, anchor=west, align=left] at (2.9,6.6) {Imagen 2};
\node[anchor=east, align=right] at (-0.5,3) {Imagen 2};
\node[rotate=55, anchor=west, align=left] at (3.9,6.6) {SDXL 1};
\node[anchor=east, align=right] at (-0.5,2) {SDXL 1};
\node[rotate=55, anchor=west, align=left] at (4.9,6.6) {DALL·E 3};
\node[anchor=east, align=right] at (-0.5,1) {DALL·E 3};

\node[anchor=east] at (0,-0.5) {};
\end{tikzpicture}}
\end{minipage}}}

\vspace{3mm}
\resizebox{1.01\linewidth}{!}{%
\mbox{%
\hspace{-3mm}%
\begin{minipage}{0.65\linewidth}
\resizebox{\linewidth}{!}{%
\begin{tikzpicture}
\begin{axis} [
        x label style={at={(0.5,-0.1)}},
        y label style={at={(-0.07,0.5)}},
        ylabel={Elo score (and 99\% CI)},
        title={Visual appeal on \dalle~Eval},
        width=15cm,
        height=10cm,
    	ybar,
    	major grid style={line width=.2pt,draw=gray!50,dashed},
    	ymajorgrids,
    	tick style={draw=none},
    	xticklabels from table={\modelnamesDalle}{model},
    	xtick={0,...,5},
    	xmin=-0.5,
    	xmax=5.5,
    	xticklabel style={yshift=0mm, align=center,rotate=0},
    	every node near coord/.append style={xshift=0,yshift=-25,color=black},
    	every axis plot/.append style={
          ybar,
          bar width=0.85,
          bar shift=0pt,
          draw=none
        },
        nodes near coords,
]
\addplot+ [mygray, error bars/.cd,
            y dir=both,
            y explicit,
            error bar style={black}]
        table
          [x=index,
            y=elo,
            y error plus expr=\thisrow{high_99_ci}-\thisrow{elo},
            y error minus expr=\thisrow{elo}-\thisrow{low_99_ci}
            ] {\datatableADalle};
\addplot+ [myblue, error bars/.cd,
            y dir=both,
            y explicit,
            error bar style={black}]
        table
          [x=index,
            y=elo,
            y error plus expr=\thisrow{high_99_ci}-\thisrow{elo},
            y error minus expr=\thisrow{elo}-\thisrow{low_99_ci}
            ] {\datatableBDalle};     

\end{axis}
\end{tikzpicture}}
\end{minipage}
\hspace{2mm}
\begin{minipage}{0.35\linewidth}
\resizebox{\linewidth}{!}{%
\begin{tikzpicture}[scale=1.1]

\draw[very thin] (-0.5,5.5) rectangle +(1,1);
\draw[very thin, fill={rgb,1:red,0.987;green,0.647;blue,0.364}] (-0.5,4.5) rectangle +(1,1);
\node[black] at (0,5) {41.9\strut};
\draw[very thin, fill={rgb,1:red,0.979;green,0.587;blue,0.337}] (-0.5,3.5) rectangle +(1,1);
\node[black] at (0,4) {40.9\strut};
\draw[very thin, fill={rgb,1:red,0.960;green,0.447;blue,0.272}] (-0.5,2.5) rectangle +(1,1);
\node[black] at (0,3) {38.9\strut};
\draw[very thin, fill={rgb,1:red,0.647;green,0.000;blue,0.149}] (-0.5,1.5) rectangle +(1,1);
\node[white] at (0,2) {30.9\strut};
\draw[very thin, fill={rgb,1:red,0.809;green,0.155;blue,0.152}] (-0.5,0.5) rectangle +(1,1);
\node[white] at (0,1) {34.1\strut};
\draw[very thin, fill={rgb,1:red,0.617;green,0.836;blue,0.412}] (0.5,5.5) rectangle +(1,1);
\node[black] at (1,6) {58.1\strut};
\draw[very thin] (0.5,4.5) rectangle +(1,1);
\draw[very thin, fill={rgb,1:red,0.998;green,0.940;blue,0.649}] (0.5,3.5) rectangle +(1,1);
\node[black] at (1,4) {48.2\strut};
\draw[very thin, fill={rgb,1:red,0.991;green,0.677;blue,0.378}] (0.5,2.5) rectangle +(1,1);
\node[black] at (1,3) {42.3\strut};
\draw[very thin, fill={rgb,1:red,0.872;green,0.249;blue,0.181}] (0.5,1.5) rectangle +(1,1);
\node[white] at (1,2) {35.8\strut};
\draw[very thin, fill={rgb,1:red,0.881;green,0.268;blue,0.190}] (0.5,0.5) rectangle +(1,1);
\node[white] at (1,1) {36.1\strut};
\draw[very thin, fill={rgb,1:red,0.557;green,0.810;blue,0.405}] (1.5,5.5) rectangle +(1,1);
\node[black] at (2,6) {59.1\strut};
\draw[very thin, fill={rgb,1:red,0.927;green,0.969;blue,0.649}] (1.5,4.5) rectangle +(1,1);
\node[black] at (2,5) {51.8\strut};
\draw[very thin] (1.5,3.5) rectangle +(1,1);
\draw[very thin, fill={rgb,1:red,0.998;green,0.950;blue,0.665}] (1.5,2.5) rectangle +(1,1);
\node[black] at (2,3) {48.5\strut};
\draw[very thin, fill={rgb,1:red,0.859;green,0.221;blue,0.168}] (1.5,1.5) rectangle +(1,1);
\node[white] at (2,2) {35.3\strut};
\draw[very thin, fill={rgb,1:red,0.881;green,0.268;blue,0.190}] (1.5,0.5) rectangle +(1,1);
\node[white] at (2,1) {36.0\strut};
\draw[very thin, fill={rgb,1:red,0.420;green,0.750;blue,0.390}] (2.5,5.5) rectangle +(1,1);
\node[black] at (3,6) {61.1\strut};
\draw[very thin, fill={rgb,1:red,0.646;green,0.849;blue,0.415}] (2.5,4.5) rectangle +(1,1);
\node[black] at (3,5) {57.7\strut};
\draw[very thin, fill={rgb,1:red,0.939;green,0.974;blue,0.665}] (2.5,3.5) rectangle +(1,1);
\node[black] at (3,4) {51.5\strut};
\draw[very thin] (2.5,2.5) rectangle +(1,1);
\draw[very thin, fill={rgb,1:red,0.993;green,0.732;blue,0.422}] (2.5,1.5) rectangle +(1,1);
\node[black] at (3,2) {43.4\strut};
\draw[very thin, fill={rgb,1:red,0.701;green,0.052;blue,0.150}] (2.5,0.5) rectangle +(1,1);
\node[white] at (3,1) {32.0\strut};
\draw[very thin, fill={rgb,1:red,0.000;green,0.408;blue,0.216}] (3.5,5.5) rectangle +(1,1);
\node[white] at (4,6) {69.1\strut};
\draw[very thin, fill={rgb,1:red,0.178;green,0.633;blue,0.333}] (3.5,4.5) rectangle +(1,1);
\node[white] at (4,5) {64.2\strut};
\draw[very thin, fill={rgb,1:red,0.143;green,0.616;blue,0.324}] (3.5,3.5) rectangle +(1,1);
\node[white] at (4,4) {64.7\strut};
\draw[very thin, fill={rgb,1:red,0.702;green,0.873;blue,0.449}] (3.5,2.5) rectangle +(1,1);
\node[black] at (4,3) {56.6\strut};
\draw[very thin] (3.5,1.5) rectangle +(1,1);
\draw[very thin, fill={rgb,1:red,0.956;green,0.982;blue,0.689}] (3.5,0.5) rectangle +(1,1);
\node[black] at (4,1) {51.1\strut};
\draw[very thin, fill={rgb,1:red,0.084;green,0.563;blue,0.296}] (4.5,5.5) rectangle +(1,1);
\node[white] at (5,6) {65.9\strut};
\draw[very thin, fill={rgb,1:red,0.201;green,0.644;blue,0.339}] (4.5,4.5) rectangle +(1,1);
\node[white] at (5,5) {63.9\strut};
\draw[very thin, fill={rgb,1:red,0.201;green,0.644;blue,0.339}] (4.5,3.5) rectangle +(1,1);
\node[white] at (5,4) {64.0\strut};
\draw[very thin, fill={rgb,1:red,0.028;green,0.460;blue,0.243}] (4.5,2.5) rectangle +(1,1);
\node[white] at (5,3) {68.0\strut};
\draw[very thin, fill={rgb,1:red,0.999;green,0.964;blue,0.689}] (4.5,1.5) rectangle +(1,1);
\node[black] at (5,2) {48.9\strut};
\draw[very thin] (4.5,0.5) rectangle +(1,1);

\node[rotate=55, anchor=west, align=left] at (-0.1,6.6) {MJ v6};
\node[anchor=east, align=right] at (-0.5,6) {MJ v6};
\node[rotate=55, anchor=west, align=left] at (0.9,6.6) {Imagen 3};
\node[anchor=east, align=right] at (-0.5,5) {Imagen 3};
\node[rotate=55, anchor=west, align=left] at (1.9,6.6) {SD 3};
\node[anchor=east, align=right] at (-0.5,4) {SD 3};
\node[rotate=55, anchor=west, align=left] at (2.9,6.6) {DALL·E 3};
\node[anchor=east, align=right] at (-0.5,3) {DALL·E 3};
\node[rotate=55, anchor=west, align=left] at (3.9,6.6) {SDXL 1};
\node[anchor=east, align=right] at (-0.5,2) {SDXL 1};
\node[rotate=55, anchor=west, align=left] at (4.9,6.6) {Imagen 2};
\node[anchor=east, align=right] at (-0.5,1) {Imagen 2};

\node[anchor=east] at (0,-0.5) {};
\end{tikzpicture}}
\end{minipage}}}

\caption{{\bf Visual Appeal:} Elo scores and win-rate percentages on \genaibench{}, \drawbench{}, and \dalle~Eval. Please refer to Appendix~\ref{app:imagen3-002} for updated human evaluation results as of December 2024.}
\label{fig:elo_visual_appeal}
\end{figure}

%% file: assets/human_evals/elo_image_content_recreation.tikz
\pgfplotstableread[col sep=comma]{
index,model,elo,low_99_ci,high_99_ci
0,SDXL 1,793,783,803
1,DALL·E 3,913,904,922
2,Imagen 2,971,963,980
3,SD 3,1051,1043,1060
4,MJ v6,1079,1070,1087
5,Imagen 3,1193,1184,1202
}\modelnames

\pgfplotstableread[col sep=comma]{
index,model,elo,low_99_ci,high_99_ci
0,SDXL 1,793,783,803
1,DALL·E 3,913,904,922
3,SD 3,1051,1043,1060
4,MJ v6,1079,1070,1087
}\datatableA

\pgfplotstableread[col sep=comma]{
index,model,elo,low_99_ci,high_99_ci
2,Imagen 2,971,963,980
5,Imagen 3,1193,1184,1202
}\datatableB

\begin{figure}
\centering
\resizebox{1.01\linewidth}{!}{%
\mbox{%
\hspace{-3mm}%
\begin{minipage}{0.65\linewidth}
\resizebox{\linewidth}{!}{%
\begin{tikzpicture}
\begin{axis} [
        x label style={at={(0.5,-0.1)}},
        y label style={at={(-0.07,0.5)}},
        ylabel={Elo score (and 99\% CI)},
        width=15cm,
        height=10cm,
    	ybar,
    	major grid style={line width=.2pt,draw=gray!50,dashed},
    	ymajorgrids,
    	tick style={draw=none},
    	xticklabels from table={\modelnames}{model},
    	xtick={0,...,5},
    	xmin=-0.5,
    	xmax=5.5,
    	xticklabel style={yshift=0mm, align=center,rotate=0},
    	every node near coord/.append style={xshift=0,yshift=-25,color=black},
    	every axis plot/.append style={
          ybar,
          bar width=0.85,
          bar shift=0pt,
          draw=none
        },
        nodes near coords,
]
\addplot+ [mygray, error bars/.cd,
            y dir=both,
            y explicit,
            error bar style={black}]
        table
          [x=index,
            y=elo,
            y error plus expr=\thisrow{high_99_ci}-\thisrow{elo},
            y error minus expr=\thisrow{elo}-\thisrow{low_99_ci}
            ] {\datatableA};
\addplot+ [myblue, error bars/.cd,
            y dir=both,
            y explicit,
            error bar style={black}]
        table
          [x=index,
            y=elo,
            y error plus expr=\thisrow{high_99_ci}-\thisrow{elo},
            y error minus expr=\thisrow{elo}-\thisrow{low_99_ci}
            ] {\datatableB};     

\end{axis}
\end{tikzpicture}}
\end{minipage}
\hspace{2mm}
\begin{minipage}{0.35\linewidth}
\resizebox{\linewidth}{!}{%
\begin{tikzpicture}[scale=1.1]
\draw[very thin] (-0.5,5.5) rectangle +(1,1);
\draw[very thin, fill={rgb,1:red,0.993;green,0.740;blue,0.429}] (-0.5,4.5) rectangle +(1,1);
\node[black] at (0,5) {37.0\strut};
\draw[very thin, fill={rgb,1:red,0.969;green,0.517;blue,0.304}] (-0.5,3.5) rectangle +(1,1);
\node[black] at (0,4) {29.8\strut};
\draw[very thin, fill={rgb,1:red,0.939;green,0.390;blue,0.246}] (-0.5,2.5) rectangle +(1,1);
\node[white] at (0,3) {25.8\strut};
\draw[very thin, fill={rgb,1:red,0.859;green,0.221;blue,0.168}] (-0.5,1.5) rectangle +(1,1);
\node[white] at (0,2) {20.5\strut};
\draw[very thin, fill={rgb,1:red,0.647;green,0.000;blue,0.149}] (-0.5,0.5) rectangle +(1,1);
\node[white] at (0,1) {11.7\strut};
\draw[very thin, fill={rgb,1:red,0.710;green,0.876;blue,0.454}] (0.5,5.5) rectangle +(1,1);
\node[black] at (1,6) {63.0\strut};
\draw[very thin] (0.5,4.5) rectangle +(1,1);
\draw[very thin, fill={rgb,1:red,0.998;green,0.950;blue,0.665}] (0.5,3.5) rectangle +(1,1);
\node[black] at (1,4) {47.0\strut};
\draw[very thin, fill={rgb,1:red,0.994;green,0.763;blue,0.448}] (0.5,2.5) rectangle +(1,1);
\node[black] at (1,3) {37.9\strut};
\draw[very thin, fill={rgb,1:red,0.964;green,0.477;blue,0.286}] (0.5,1.5) rectangle +(1,1);
\node[black] at (1,2) {28.7\strut};
\draw[very thin, fill={rgb,1:red,0.816;green,0.162;blue,0.152}] (0.5,0.5) rectangle +(1,1);
\node[white] at (1,1) {18.3\strut};
\draw[very thin, fill={rgb,1:red,0.489;green,0.780;blue,0.398}] (1.5,5.5) rectangle +(1,1);
\node[black] at (2,6) {70.2\strut};
\draw[very thin, fill={rgb,1:red,0.939;green,0.974;blue,0.665}] (1.5,4.5) rectangle +(1,1);
\node[black] at (2,5) {53.0\strut};
\draw[very thin] (1.5,3.5) rectangle +(1,1);
\draw[very thin, fill={rgb,1:red,0.994;green,0.786;blue,0.468}] (1.5,2.5) rectangle +(1,1);
\node[black] at (2,3) {38.9\strut};
\draw[very thin, fill={rgb,1:red,0.978;green,0.577;blue,0.332}] (1.5,1.5) rectangle +(1,1);
\node[black] at (2,2) {31.6\strut};
\draw[very thin, fill={rgb,1:red,0.832;green,0.177;blue,0.153}] (1.5,0.5) rectangle +(1,1);
\node[white] at (2,1) {19.1\strut};
\draw[very thin, fill={rgb,1:red,0.353;green,0.718;blue,0.377}] (2.5,5.5) rectangle +(1,1);
\node[white] at (3,6) {74.2\strut};
\draw[very thin, fill={rgb,1:red,0.733;green,0.887;blue,0.469}] (2.5,4.5) rectangle +(1,1);
\node[black] at (3,5) {62.1\strut};
\draw[very thin, fill={rgb,1:red,0.757;green,0.897;blue,0.484}] (2.5,3.5) rectangle +(1,1);
\node[black] at (3,4) {61.1\strut};
\draw[very thin] (2.5,2.5) rectangle +(1,1);
\draw[very thin, fill={rgb,1:red,0.995;green,0.832;blue,0.506}] (2.5,1.5) rectangle +(1,1);
\node[black] at (3,2) {40.6\strut};
\draw[very thin, fill={rgb,1:red,0.980;green,0.597;blue,0.341}] (2.5,0.5) rectangle +(1,1);
\node[black] at (3,1) {32.1\strut};
\draw[very thin, fill={rgb,1:red,0.143;green,0.616;blue,0.324}] (3.5,5.5) rectangle +(1,1);
\node[white] at (4,6) {79.5\strut};
\draw[very thin, fill={rgb,1:red,0.449;green,0.763;blue,0.394}] (3.5,4.5) rectangle +(1,1);
\node[black] at (4,5) {71.3\strut};
\draw[very thin, fill={rgb,1:red,0.548;green,0.806;blue,0.404}] (3.5,3.5) rectangle +(1,1);
\node[black] at (4,4) {68.4\strut};
\draw[very thin, fill={rgb,1:red,0.804;green,0.917;blue,0.515}] (3.5,2.5) rectangle +(1,1);
\node[black] at (4,3) {59.4\strut};
\draw[very thin] (3.5,1.5) rectangle +(1,1);
\draw[very thin, fill={rgb,1:red,0.983;green,0.617;blue,0.350}] (3.5,0.5) rectangle +(1,1);
\node[black] at (4,1) {32.7\strut};
\draw[very thin, fill={rgb,1:red,0.000;green,0.408;blue,0.216}] (4.5,5.5) rectangle +(1,1);
\node[white] at (5,6) {88.3\strut};
\draw[very thin, fill={rgb,1:red,0.088;green,0.570;blue,0.300}] (4.5,4.5) rectangle +(1,1);
\node[white] at (5,5) {81.7\strut};
\draw[very thin, fill={rgb,1:red,0.096;green,0.585;blue,0.308}] (4.5,3.5) rectangle +(1,1);
\node[white] at (5,4) {80.9\strut};
\draw[very thin, fill={rgb,1:red,0.567;green,0.814;blue,0.407}] (4.5,2.5) rectangle +(1,1);
\node[black] at (5,3) {67.9\strut};
\draw[very thin, fill={rgb,1:red,0.587;green,0.823;blue,0.409}] (4.5,1.5) rectangle +(1,1);
\node[black] at (5,2) {67.3\strut};
\draw[very thin] (4.5,0.5) rectangle +(1,1);

\node[rotate=55, anchor=west, align=left] at (-0.1,6.6) {Imagen 3};
\node[anchor=east, align=right] at (-0.5,6) {Imagen 3};
\node[rotate=55, anchor=west, align=left] at (0.9,6.6) {MJ v6};
\node[anchor=east, align=right] at (-0.5,5) {MJ v6};
\node[rotate=55, anchor=west, align=left] at (1.9,6.6) {SD 3};
\node[anchor=east, align=right] at (-0.5,4) {SD 3};
\node[rotate=55, anchor=west, align=left] at (2.9,6.6) {Imagen 2};
\node[anchor=east, align=right] at (-0.5,3) {Imagen 2};
\node[rotate=55, anchor=west, align=left] at (3.9,6.6) {DALL·E 3};
\node[anchor=east, align=right] at (-0.5,2) {DALL·E 3};
\node[rotate=55, anchor=west, align=left] at (4.9,6.6) {SDXL 1};
\node[anchor=east, align=right] at (-0.5,1) {SDXL 1};

\node[anchor=east] at (0,-1) {};
\end{tikzpicture}}
\end{minipage}}}
\caption{{\bf Detailed prompt--image alignment:} Elo scores and win percentages on \doccionek{}.}
\label{fig:docci_elo}
\end{figure}

%% file: assets/human_evals/geckonum_count.tikz
\pgfplotstableread[col sep=comma]{
index,model,accuracy,low_05_ci,high_05_ci
0,SDXL 1,20.530665506742064,18.87954068192288,22.181790331561245
1,Imagen 2,38.25503355704698,36.24012936519642,40.26993774889755
2,MJ v6,42.646420824295014,40.627430754460434,44.66541089412959
3,SD 3,45.50586191923578,43.47205515085093,47.53966868762063
4,DALL·E 3,46.03381014304291,43.999938068893265,48.06768221719256
5,Imagen 3,58.60424794104898,56.59438366566156,60.614112216436396
}\modelnames

\pgfplotstableread[col sep=comma]{
index,model,accuracy,low_05_ci,high_05_ci
0,SDXL 1,20.530665506742064,18.87954068192288,22.181790331561245
2,MJ v6,42.646420824295014,40.627430754460434,44.66541089412959
3,SD 3,45.50586191923578,43.47205515085093,47.53966868762063
4,DALL·E 3,46.03381014304291,43.999938068893265,48.06768221719256
}\datatableA

\pgfplotstableread[col sep=comma]{
index,model,accuracy,low_05_ci,high_05_ci
1,Imagen 2,38.25503355704698,36.24012936519642,40.26993774889755
5,Imagen 3,58.60424794104898,56.59438366566156,60.614112216436396
}\datatableB

\begin{tikzpicture}
\begin{axis} [
        x label style={at={(0.5,-0.1)}},
        y label style={at={(-0.07,0.5)}},
        ylabel={Counting accuracy (Percentage)},
        width=17cm,
        height=10cm,
    	ybar,
    	major grid style={line width=.2pt,draw=gray!50,dashed},
    	ymajorgrids,
    	tick style={draw=none},
    	xticklabels from table={\modelnames}{model},
    	xtick={0,...,5},
    	xmin=-0.5,
    	xmax=5.5,
    	ymin=0,
    	ymax=100,
    	xticklabel style={yshift=0mm, align=center,rotate=0},
    	every node near coord/.append style={xshift=0,yshift=-25,color=black,/pgf/number format/.cd,fixed zerofill,precision=1},
    	every axis plot/.append style={
          ybar,
          bar width=0.85,
          bar shift=0pt,
          draw=none
        },
        nodes near coords,
]
\addplot+ [mygray, error bars/.cd,
            y dir=both,
            y explicit,
            error bar style={black}]
        table
          [x=index,
            y=accuracy,
            y error plus expr=\thisrow{high_05_ci}-\thisrow{accuracy},
            y error minus expr=\thisrow{accuracy}-\thisrow{low_05_ci}
            ] {\datatableA};
\addplot+ [myblue, error bars/.cd,
            y dir=both,
            y explicit,
            error bar style={black}]
        table
          [x=index,
            y=accuracy,
            y error plus expr=\thisrow{high_05_ci}-\thisrow{accuracy},
            y error minus expr=\thisrow{accuracy}-\thisrow{low_05_ci}
            ] {\datatableB};     

\end{axis}
\end{tikzpicture}

%% file: assets/experiments/auto_evals/bar_plots_gecko_rel_filtered.tex
\pgfplotstableread[col sep=comma]{
index,model,accuracy,low_05_ci,high_05_ci
0,SDXL 1,36.050849067944455,33.62850726807694,38.41514602278189
1,Imagen 2,38.16996606960529,35.819115719536654,40.527388580057945
2,Midjourney,51.96258744683868,49.232127964564064,54.5991683620621
3,Dalle·E 3,56.619517442001566,54.17018409299437,58.94632621472189
4,SD3,57.65368021140539,55.24772215706998,60.01363195204359
5,Imagen 3,67.37571304516429,64.01648164131954,70.28590699749853
}\modelnames

\pgfplotstableread[col sep=comma]{
index,model,accuracy,low_05_ci,high_05_ci
0,SDXL 1,36.050849067944455,33.62850726807694,38.41514602278189
2,Midjourney,51.96258744683868,49.232127964564064,54.5991683620621
3,Dalle·E 3,56.619517442001566,54.17018409299437,58.94632621472189
4,SD3,57.65368021140539,55.24772215706998,60.01363195204359
}\datatableA

\pgfplotstableread[col sep=comma]{
index,model,accuracy,low_05_ci,high_05_ci
1,Imagen 2,38.16996606960529,35.819115719536654,40.527388580057945
5,Imagen 3,67.37571304516429,64.01648164131954,70.28590699749853
}\datatableB

\begin{tikzpicture}
\begin{axis} [
        x label style={at={(0.5,-0.1)}},
        y label style={at={(-0.07,0.5)}},
        title={\vqascore~performance on \geckorel},
        width=13.5cm,
        height=9cm,
    	ybar,
    	major grid style={line width=.2pt,draw=gray!50,dashed},
    	ymajorgrids,
    	tick style={draw=none},
    	xticklabels from table={\modelnames}{model},
    	xtick={0,...,5},
    	xmin=-0.5,
    	xmax=5.5,
	    ymin=0,
    	ymax=100,
    	xticklabel style={yshift=0mm, align=center,rotate=0},
    	every node near coord/.append style={xshift=0,yshift=-25,color=black,/pgf/number format/.cd,fixed zerofill,precision=1},
    	every axis plot/.append style={
          ybar,
          bar width=0.85,
          bar shift=0pt,
          draw=none
        },
        nodes near coords,
]
\addplot+ [mygray, error bars/.cd,
            y dir=both,
            y explicit,
            error bar style={black}]
        table
          [x=index,
            y=accuracy,
            y error plus expr=\thisrow{high_05_ci}-\thisrow{accuracy},
            y error minus expr=\thisrow{accuracy}-\thisrow{low_05_ci}
            ] {\datatableA};
\addplot+ [myblue, error bars/.cd,
            y dir=both,
            y explicit,
            error bar style={black}]
        table
          [x=index,
            y=accuracy,
            y error plus expr=\thisrow{high_05_ci}-\thisrow{accuracy},
            y error minus expr=\thisrow{accuracy}-\thisrow{low_05_ci}
            ] {\datatableB};

\node at (3.5, 63) {ns};
\draw (3., 62) -- (3.,63) -- (3.35, 63);
\draw (3.65, 63) -- (4.,63) -- (4, 62);

\node at (0.5, 45) {ns};
\draw (0., 44) -- (0.,45) -- (0.35, 45);
\draw (0.65, 45) -- (1.,45) -- (1, 44);

\end{axis}
\end{tikzpicture}

%% file: assets/experiments/auto_evals/bar_plots_genaibench.tex
\pgfplotstableread[col sep=comma]{
index,model,accuracy,low_05_ci,high_05_ci
0,SDXL 1,32.58993173646906,30.90423697052752,34.234613774410306
1,Imagen 2,40.85565612225782,39.187242224655186,42.553410507107074
2,MJ v6,44.82944893580144,42.95933617697492,46.59585228339058
3,SD 3,50.9340943531429,49.02745850422014,52.81516389496872
4,DALL·E 3,57.892711566740616,56.12522793306369,59.56550755427745
5,Imagen 3,63.40239272792064,61.18966667494099,65.63261814604326
}\modelnames

\pgfplotstableread[col sep=comma]{
index,model,accuracy,low_05_ci,high_05_ci
0,SDXL 1,32.58993173646906,30.90423697052752,34.234613774410306
2,MJ v6,44.82944893580144,42.95933617697492,46.59585228339058
3,SD 3,50.9340943531429,49.02745850422014,52.81516389496872
4,DALL·E 3,57.892711566740616,56.12522793306369,59.56550755427745
}\datatableA

\pgfplotstableread[col sep=comma]{
index,model,accuracy,low_05_ci,high_05_ci
1,Imagen 2,40.85565612225782,39.187242224655186,42.553410507107074
5,Imagen 3,63.40239272792064,61.18966667494099,65.63261814604326
}\datatableB

\begin{tikzpicture}
\begin{axis} [
        x label style={at={(0.5,-0.1)}},
        y label style={at={(-0.07,0.5)}},
        title={\vqascore~performance on \genaibench},
        width=13.5cm,
        height=9cm,
    	ybar,
    	major grid style={line width=.2pt,draw=gray!50,dashed},
    	ymajorgrids,
    	tick style={draw=none},
    	xticklabels from table={\modelnames}{model},
    	xtick={0,...,5},
    	xmin=-0.5,
    	xmax=5.5,
    	ymin=0,
    	ymax=100,
    	xticklabel style={yshift=0mm, align=center,rotate=0},
    	every node near coord/.append style={xshift=0,yshift=-25,color=black,/pgf/number format/.cd,fixed zerofill,precision=1},
    	every axis plot/.append style={
          ybar,
          bar width=0.85,
          bar shift=0pt,
          draw=none
        },
        nodes near coords,
]
\addplot+ [mygray, error bars/.cd,
            y dir=both,
            y explicit,
            error bar style={black}]
        table
          [x=index,
            y=accuracy,
            y error plus expr=\thisrow{high_05_ci}-\thisrow{accuracy},
            y error minus expr=\thisrow{accuracy}-\thisrow{low_05_ci}
            ] {\datatableA};
\addplot+ [myblue, error bars/.cd,
            y dir=both,
            y explicit,
            error bar style={black}]
        table
          [x=index,
            y=accuracy,
            y error plus expr=\thisrow{high_05_ci}-\thisrow{accuracy},
            y error minus expr=\thisrow{accuracy}-\thisrow{low_05_ci}
            ] {\datatableB};

\end{axis}
\end{tikzpicture}

%% file: assets/experiments/auto_evals/bar_plots_gecko_dalle3.tex
\pgfplotstableread[col sep=comma]{
index,model,accuracy,low_05_ci,high_05_ci
0,Imagen 2,35.73472525916375,31.355739015764826,40.19143860135561
1,SDXL 1,36.45857619154467,31.552344510617118,41.511974548495154
2,MJ v6,51.46300594966995,46.11458908799114,56.676315568776346
3,DALL·E 3,52.6272847155852,47.239703383918226,57.86009656939154
4,SD 3,57.58253106510482,52.52332706070435,62.5323058592648
5,Imagen 3,69.51890591167336,61.86616054851349,76.35165201186905
}\modelnames

\pgfplotstableread[col sep=comma]{
index,model,accuracy,low_05_ci,high_05_ci
1,SDXL 1,36.45857619154467,31.552344510617118,41.511974548495154
2,MJ v6,51.46300594966995,46.11458908799114,56.676315568776346
3,DALL·E 3,52.6272847155852,47.239703383918226,57.86009656939154
4,SD 3,57.58253106510482,52.52332706070435,62.5323058592648
}\datatableA

\pgfplotstableread[col sep=comma]{
index,model,accuracy,low_05_ci,high_05_ci
0,Imagen 2,35.73472525916375,31.355739015764826,40.19143860135561
5,Imagen 3,69.51890591167336,61.86616054851349,76.35165201186905
}\datatableB

\begin{tikzpicture}
\begin{axis} [
        x label style={at={(0.5,-0.1)}},
        y label style={at={(-0.07,0.5)}},
        title={\vqascore~performance on \dalleeval},
        width=13.5cm,
        height=9cm,
    	ybar,
    	major grid style={line width=.2pt,draw=gray!50,dashed},
    	ymajorgrids,
    	tick style={draw=none},
    	xticklabels from table={\modelnames}{model},
    	xtick={0,...,5},
    	xmin=-0.5,
    	xmax=5.5,
    	ymin=0,
    	ymax=100,
    	xticklabel style={yshift=0mm, align=center,rotate=0},
    	every node near coord/.append style={xshift=0,yshift=-25,color=black,/pgf/number format/.cd,fixed zerofill,precision=1},
    	every axis plot/.append style={
          ybar,
          bar width=0.85,
          bar shift=0pt,
          draw=none
        },
        nodes near coords,
]
\addplot+ [mygray, error bars/.cd,
            y dir=both,
            y explicit,
            error bar style={black}]
        table
          [x=index,
            y=accuracy,
            y error plus expr=\thisrow{high_05_ci}-\thisrow{accuracy},
            y error minus expr=\thisrow{accuracy}-\thisrow{low_05_ci}
            ] {\datatableA};
\addplot+ [myblue, error bars/.cd,
            y dir=both,
            y explicit,
            error bar style={black}]
        table
          [x=index,
            y=accuracy,
            y error plus expr=\thisrow{high_05_ci}-\thisrow{accuracy},
            y error minus expr=\thisrow{accuracy}-\thisrow{low_05_ci}
            ] {\datatableB};   

\node at (2.5, 61) {ns};
\draw (2., 60) -- (2.,61) -- (2.35, 61);
\draw (2.65, 61) -- (3.,61) -- (3, 60);

\node at (3.5, 65) {ns};
\draw (3., 64) -- (3.,65) -- (3.35, 65);
\draw (3.65, 65) -- (4.,65) -- (4, 64);

\node at (0.5, 48) {ns};
\draw (0., 47) -- (0.,48) -- (0.35, 48);
\draw (0.65, 48) -- (1.,48) -- (1, 47);

\end{axis}
\end{tikzpicture}

%% file: assets/experiments/auto_evals/bar_plots_gecko_docci.tex
\pgfplotstableread[col sep=comma]{
index,model,accuracy,low_05_ci,high_05_ci
0,SDXL 1,21.346852706936488,19.67705279033362,22.966048652866057
1,Imagen 2,36.58854040013978,34.593644558105666,38.595953190655955
2,DALL·E 3,50.04456891506022,48.264658892776225,51.82858180942163
3,SD 3,54.44898662157143,52.43879159838875,56.542830091166664
4,MJ v6,57.86173124664904,56.17270609845229,59.566851363262195
5,Imagen 3,72.9391049407876,70.38785702983643,75.4845405245445
}\modelnames

\pgfplotstableread[col sep=comma]{
index,model,accuracy,low_05_ci,high_05_ci
0,SDXL 1,21.346852706936488,19.67705279033362,22.966048652866057
2,DALL·E 3,50.04456891506022,48.264658892776225,51.82858180942163
3,SD 3,54.44898662157143,52.43879159838875,56.542830091166664
4,MJ v6,57.86173124664904,56.17270609845229,59.566851363262195
}\datatableA

\pgfplotstableread[col sep=comma]{
index,model,accuracy,low_05_ci,high_05_ci
1,Imagen 2,36.58854040013978,34.593644558105666,38.595953190655955
5,Imagen 3,72.9391049407876,70.38785702983643,75.4845405245445
}\datatableB

\begin{tikzpicture}
\begin{axis} [
        x label style={at={(0.5,-0.1)}},
        y label style={at={(-0.07,0.5)}},
        title={\vqascore~performance on \doccionek},
        width=13.5cm,
        height=9cm,
    	ybar,
    	major grid style={line width=.2pt,draw=gray!50,dashed},
    	ymajorgrids,
    	tick style={draw=none},
    	xticklabels from table={\modelnames}{model},
    	xtick={0,...,5},
    	xmin=-0.5,
    	xmax=5.5,
    	ymin=0,
    	ymax=100,
    	xticklabel style={yshift=0mm, align=center,rotate=0},
    	every node near coord/.append style={xshift=0,yshift=-25,color=black,/pgf/number format/.cd,fixed zerofill,precision=1},
    	every axis plot/.append style={
          ybar,
          bar width=0.85,
          bar shift=0pt,
          draw=none
        },
        nodes near coords,
]
\addplot+ [mygray, error bars/.cd,
            y dir=both,
            y explicit,
            error bar style={black}]
        table
          [x=index,
            y=accuracy,
            y error plus expr=\thisrow{high_05_ci}-\thisrow{accuracy},
            y error minus expr=\thisrow{accuracy}-\thisrow{low_05_ci}
            ] {\datatableA};
\addplot+ [myblue, error bars/.cd,
            y dir=both,
            y explicit,
            error bar style={black}]
        table
          [x=index,
            y=accuracy,
            y error plus expr=\thisrow{high_05_ci}-\thisrow{accuracy},
            y error minus expr=\thisrow{accuracy}-\thisrow{low_05_ci}
            ] {\datatableB};     
\end{axis}
\end{tikzpicture}

%% file: assets/experiments/category_barplots.tikz
\pgfplotstableread[col sep=comma]{
index,vqascore,low_05_ci,high_05_ci
0,17.92688907396358,5.001188815606259,4.872654412369162
1,9.220880917124365,5.28411910910628,6.79463081790901
2,19.863176494136983,8.443077839089307,8.897585355852954
3,51.57694199113364,10.887815822861185,11.239328966950652
4,33.651714056112176,7.036276677465725,8.787816431698268
5,31.049132083219718,14.516639447024563,15.011224616634372
6,26.799867192110092,12.789793171045561,14.398295738900622
7,33.1366250446972,9.619637040410272,10.055903863664199
}\sdxl
\pgfplotstableread[col sep=comma]{
index,vqascore,low_05_ci,high_05_ci
0,27.057005906988824,5.303722125910429,6.322489303597674
1,27.601892479224617,10.698461911019074,10.62360350633356
2,23.344663043912913,8.778102058711303,9.807615952763133
3,46.37816553960429,11.325616302613177,12.478087009281502
4,31.800630776697798,8.101555022618808,9.292658620784248
5,28.448077459095543,15.09656883302988,17.012550322645225
6,41.05657023967365,14.917603872187716,16.404356347570122
7,38.24195377931212,9.131518465416136,9.499445997930685
}\imagentwo
\pgfplotstableread[col sep=comma]{
index,vqascore,low_05_ci,high_05_ci
0,43.5591534027789,6.180532983261692,6.154543643473481
1,19.73786192034875,8.386352901454263,10.653833992510396
2,34.60101766706396,10.291106173734079,11.10284596660457
3,67.82301093560473,9.521965125918335,9.064153721696286
4,44.36500696354149,9.112950997499553,8.874713158815439
5,39.312577044217214,16.31658411001781,17.678044650915723
6,58.387449350664475,16.66081238233923,15.878375529759914
7,49.35132593430006,9.636254918464374,9.490879088506546
}\midjourneysix
\pgfplotstableread[col sep=comma]{
index,vqascore,low_05_ci,high_05_ci
0,53.11615863996294,6.178920190454193,5.840305948214763
1,32.35521745893667,10.434892480389223,11.807598411306003
2,36.54538138398197,9.205277420400556,9.477152222411924
3,74.60283317583212,8.961360805294134,7.629160687599679
4,49.71573365729354,7.356856928853269,7.8691794652755505
5,43.3061174737783,15.699107759336247,14.397261956734766
6,54.66473947228003,14.668283144702102,15.556241174832891
7,52.687784054813434,9.02011824555786,8.17122779989159
}\dallethree
\pgfplotstableread[col sep=comma]{
index,vqascore,low_05_ci,high_05_ci
0,44.30477426717508,6.743008117052867,6.335667132331622
1,35.86715691786576,12.549457504145547,13.616992414870282
2,29.29040957057858,9.332201469557756,10.170278306270522
3,69.88023047471275,10.550198633960093,9.109959082132846
4,47.659084736427545,9.11094178714798,8.929131031331778
5,31.844183954491793,15.732409507605652,15.764791785567649
6,63.34873452945446,15.265137004663252,14.249181640233374
7,51.98841083999034,9.14435901912426,9.305901335322567
}\sdthree
\pgfplotstableread[col sep=comma]{
index,vqascore,low_05_ci,high_05_ci
0,66.49993878701507,9.560321464051917,9.318519664946422
1,38.389182692860444,16.198890348745763,15.68121020704028
2,41.60142533849045,12.92535800920664,13.240740703680721
3,84.27653997253445,9.92020063397878,9.201652066879788
4,64.37129228007952,11.214554192508416,10.370032421854674
5,31.824202739906454,18.181833868279487,18.183483427317462
6,64.53467199461898,20.97513477991707,19.751012109896777
7,61.80312855990236,13.53506842045914,13.342442811220565
}\imagenthree

\definecolor{color1bg}{HTML}{ef476f}
\definecolor{color2bg}{HTML}{f78c6b}
\definecolor{color3bg}{HTML}{ffd166}
\definecolor{color4bg}{HTML}{06d6a0}
\definecolor{color5bg}{HTML}{118ab2}
\definecolor{color6bg}{HTML}{00537a}

\begin{tikzpicture}
\begin{axis}[
    ybar,
    reverse legend,
    width=22cm, height=8cm,
    ymin=0,ymax=100,
    bar width=0.12,
    legend style={at={(1.02,0.95)},
      anchor=north west,legend columns=1},
    ylabel={\vqascore},
    xticklabels={Complexity,Compositional,Action,Color,Count,Scale,Shape,Spatial},
    xtick=data,
	every axis plot/.append style={
      draw=none
    },
    ]
\addplot+ [color1bg, error bars/.cd,
            y dir=both,
            y explicit,
            error bar style={black}]
        table
          [x=index,
            y=vqascore,
            y error plus=high_05_ci,
            y error minus=low_05_ci
            ] {\sdxl};
\addplot+ [color2bg, error bars/.cd,
            y dir=both,
            y explicit,
            error bar style={black}]
        table
          [x=index,
            y=vqascore,
            y error plus=high_05_ci,
            y error minus=low_05_ci
            ] {\imagentwo};            
\addplot+ [color3bg, error bars/.cd,
            y dir=both,
            y explicit,
            error bar style={black}]
        table
          [x=index,
            y=vqascore,
            y error plus=high_05_ci,
            y error minus=low_05_ci
            ] {\midjourneysix};
\addplot+ [color4bg, error bars/.cd,
            y dir=both,
            y explicit,
            error bar style={black}]
        table
          [x=index,
            y=vqascore,
            y error plus=high_05_ci,
            y error minus=low_05_ci
            ] {\dallethree};            
\addplot+ [color5bg, error bars/.cd,
            y dir=both,
            y explicit,
            error bar style={black}]
        table
          [x=index,
            y=vqascore,
            y error plus=high_05_ci,
            y error minus=low_05_ci
            ] {\sdthree};            
\addplot+ [color6bg, error bars/.cd,
            y dir=both,
            y explicit,
            error bar style={black}]
        table
          [x=index,
            y=vqascore,
            y error plus=high_05_ci,
            y error minus=low_05_ci
            ] {\imagenthree};
\legend{SDXL 1,Imagen 2,MJ v6,DALL·E 3,SD 3,Imagen 3}
\end{axis}
\end{tikzpicture}

%% file: assets/human_evals/app_geckonum_per_count.tikz
\pgfplotstableread[col sep=comma]{
index,accuracy,low_05_ci,high_05_ci
0,28.743961352657006,24.38450766804852,33.103415037265485
1,34.01360544217687,29.591976127960677,38.43523475639306
2,21.153846153846153,17.453774863824012,24.853917443868294
3,13.590263691683571,10.565301763760168,16.615225619606974
4,7.608695652173914,4.480713938379812,10.736677365968013
}\sdxl
\pgfplotstableread[col sep=comma]{
index,accuracy,low_05_ci,high_05_ci
0,68.5,63.94782805400743,73.05217194599258
1,50.22935779816514,45.53614462004174,54.92257097628854
2,39.38053097345133,34.87624743060421,43.88481451629844
3,24.78813559322034,20.89282402647697,28.68344715996371
4,17.77777777777778,13.217413336253985,22.33814221930157
}\imagentwo
\pgfplotstableread[col sep=comma]{
index,accuracy,low_05_ci,high_05_ci
0,72.77108433734941,68.4883746168685,77.0537940578303
1,55.4054054054054,50.78187429165821,60.028936519152595
2,44.989339019189764,40.48698223113328,49.49169580724625
3,31.781376518218625,27.67534343481674,35.887409601620504
4,13.405797101449277,9.386183746923098,17.425410455975452
}\midjourneysix
\pgfplotstableread[col sep=comma]{
index,accuracy,low_05_ci,high_05_ci
0,88.19277108433735,85.08810763613755,91.29743453253715
1,54.157303370786515,49.52782904898177,58.78677769259125
2,47.97441364605544,43.45299210376292,52.495835188347954
3,32.525252525252526,28.398328592655602,36.65217645784944
4,18.478260869565215,13.899364495555206,23.057157243575226
}\dallethree
\pgfplotstableread[col sep=comma]{
index,accuracy,low_05_ci,high_05_ci
0,65.3012048192771,60.721454122122786,69.88095551643141
1,53.82882882882883,49.191696207321186,58.46596145033648
2,53.5181236673774,49.00420287057828,58.03204446417652
3,40.4040404040404,36.08122890117947,44.72685190690134
4,21.014492753623188,16.208017490714926,25.82096801653145
}\sdthree
\pgfplotstableread[col sep=comma]{
index,accuracy,low_05_ci,high_05_ci
0,86.74698795180723,83.48480329251296,90.00917261110149
1,78.42696629213484,74.60526794459521,82.24866463967446
2,60.3411513859275,55.9138558158801,64.76844695597491
3,43.83838383838384,39.46726556754053,48.209502109227145
4,35.14492753623188,29.512483950511147,40.77737112195262
}\imagenthree

\definecolor{color1bg}{HTML}{ef476f}
\definecolor{color2bg}{HTML}{f78c6b}
\definecolor{color3bg}{HTML}{ffd166}
\definecolor{color4bg}{HTML}{06d6a0}
\definecolor{color5bg}{HTML}{118ab2}
\definecolor{color6bg}{HTML}{00537a}

\begin{tikzpicture}
\begin{axis}[
    ybar,
    reverse legend,
    width=22cm, height=8cm,
    ymin=0,ymax=100,
    bar width=0.12,
    legend style={at={(1.02,0.95)},
      anchor=north west,legend columns=1},
    ylabel={{Counting Accuracy}},
    xlabel={{Ground Truth Number}},
    xticklabels={1,2,3,4,5},
    xtick=data,
	every axis plot/.append style={
      draw=none
    },
    ]
\addplot+ [color1bg, error bars/.cd,
            y dir=both,
            y explicit,
            error bar style={black}]
        table
          [x=index,
            y=accuracy,
            y error plus expr=\thisrow{high_05_ci}-\thisrow{accuracy},
            y error minus expr=\thisrow{accuracy}-\thisrow{low_05_ci}
            ] {\sdxl};
\addplot+ [color2bg, error bars/.cd,
            y dir=both,
            y explicit,
            error bar style={black}]
        table
          [x=index,
            y=accuracy,
            y error plus expr=\thisrow{high_05_ci}-\thisrow{accuracy},
            y error minus expr=\thisrow{accuracy}-\thisrow{low_05_ci}
            ] {\imagentwo};            
\addplot+ [color3bg, error bars/.cd,
            y dir=both,
            y explicit,
            error bar style={black}]
        table
          [x=index,
            y=accuracy,
            y error plus expr=\thisrow{high_05_ci}-\thisrow{accuracy},
            y error minus expr=\thisrow{accuracy}-\thisrow{low_05_ci}
            ] {\midjourneysix};
\addplot+ [color4bg, error bars/.cd,
            y dir=both,
            y explicit,
            error bar style={black}]
        table
          [x=index,
            y=accuracy,
            y error plus expr=\thisrow{high_05_ci}-\thisrow{accuracy},
            y error minus expr=\thisrow{accuracy}-\thisrow{low_05_ci}
            ] {\dallethree};            
\addplot+ [color5bg, error bars/.cd,
            y dir=both,
            y explicit,
            error bar style={black}]
        table
          [x=index,
            y=accuracy,
            y error plus expr=\thisrow{high_05_ci}-\thisrow{accuracy},
            y error minus expr=\thisrow{accuracy}-\thisrow{low_05_ci}
            ] {\sdthree};            
\addplot+ [color6bg, error bars/.cd,
            y dir=both,
            y explicit,
            error bar style={black}]
        table
          [x=index,
            y=accuracy,
            y error plus expr=\thisrow{high_05_ci}-\thisrow{accuracy},
            y error minus expr=\thisrow{accuracy}-\thisrow{low_05_ci}
            ] {\imagenthree};
\legend{SDXL 1,Imagen 2,MJ v6,DALL·E 3,SD 3,Imagen 3}
\end{axis}
\end{tikzpicture}

%% file: assets/human_evals/app_geckonum_per_prompt.tikz
\pgfplotstableread[col sep=comma]{
index,accuracy,low_05_ci,high_05_ci
0,39.55696202531646,34.16571592242752,44.948208128205394
1,17.197452229299362,11.294732392159206,23.100172066439516
2,11.428571428571429,7.444676763559572,15.412466093583282
3,26.582278481012654,16.840660419575098,36.323896542450214
4,39.0625,33.716912586472816,44.408087413527184
5,20.466321243523318,16.441469687670796,24.49117279937584
6,9.584664536741213,6.323404491421243,12.845924582061185
}\sdxl
\pgfplotstableread[col sep=comma]{
index,accuracy,low_05_ci,high_05_ci
0,62.25806451612903,56.86199997357163,67.65412905868644
1,37.16216216216216,29.376808659958307,44.94751566436601
2,19.421487603305785,14.437335622661779,24.40563958394979
3,73.33333333333333,63.32521253853454,83.34145412813211
4,64.3312101910828,59.03288860086635,69.62953178129925
5,46.28099173553719,41.15166812130001,51.41031534977437
6,22.07792207792208,17.44577763436953,26.710066521474623
}\imagentwo
\pgfplotstableread[col sep=comma]{
index,accuracy,low_05_ci,high_05_ci
0,70.12578616352201,65.09516463303609,75.15640769400792
1,36.30573248407643,28.78368552658313,43.82777944156973
2,23.387096774193548,18.11890753342483,28.65528601496226
3,63.74999999999999,53.21590580338784,74.28409419661214
4,81.25,76.97353006917825,85.52646993082175
5,47.92746113989637,42.94377106172876,52.91115121806398
6,26.198083067092654,21.32678728083432,31.069378853350987
}\midjourneysix
\pgfplotstableread[col sep=comma]{
index,accuracy,low_05_ci,high_05_ci
0,67.5,62.36823608301392,72.63176391698609
1,43.31210191082803,35.56127439997811,51.06292942167795
2,35.08064516129033,29.141237509987604,41.02005281259304
3,68.75,58.59302321979798,78.90697678020202
4,76.875,72.25537697841625,81.49462302158376
5,50.259067357512954,45.27115730352255,55.24697741150336
6,40.57507987220447,35.135192385283396,46.01496735912555
}\dallethree
\pgfplotstableread[col sep=comma]{
index,accuracy,low_05_ci,high_05_ci
0,72.10031347962382,67.17854718535912,77.02207977388854
1,37.57961783439491,30.003658129448883,45.15557753934093
2,25.0,19.610817552277087,30.389182447722913
3,61.25000000000001,50.574407264851054,71.92559273514895
4,90.3125,87.07169506254407,93.55330493745593
5,50.0,45.012022991051246,54.987977008948754
6,25.23961661341853,20.427312563287035,30.051920663550025
}\sdthree
\pgfplotstableread[col sep=comma]{
index,accuracy,low_05_ci,high_05_ci
0,78.75,74.26793809821608,83.23206190178392
1,64.3312101910828,56.838251940384524,71.82416844178108
2,50.403225806451616,44.1805362932382,56.62591531966502
3,85.0,77.17547126308475,92.82452873691525
4,76.5625,71.9212303875782,81.2037696124218
5,63.212435233160626,58.401758000214535,68.02311246610671
6,55.59105431309904,50.086608396411414,61.09550022978667
}\imagenthree

\definecolor{color1bg}{HTML}{ef476f}
\definecolor{color2bg}{HTML}{f78c6b}
\definecolor{color3bg}{HTML}{ffd166}
\definecolor{color4bg}{HTML}{06d6a0}
\definecolor{color5bg}{HTML}{118ab2}
\definecolor{color6bg}{HTML}{00537a}

\begin{tikzpicture}
\begin{axis}[
    ybar,
    reverse legend,
    width=22cm, height=8cm,
    ymin=0,ymax=100,
    bar width=0.12,
    legend style={at={(1.02,0.95)},
      anchor=north west,legend columns=1},
    ylabel={{Counting Accuracy}},
    xticklabels={numeric-simple,2-additive,3-additive,numeric-sentence,attribute-color,2-additive-color,attribute-spatial},
    xtick=data,
	every axis plot/.append style={
      draw=none
    },
    ]
\addplot+ [color1bg, error bars/.cd,
            y dir=both,
            y explicit,
            error bar style={black}]
        table
          [x=index,
            y=accuracy,
            y error plus expr=\thisrow{high_05_ci}-\thisrow{accuracy},
            y error minus expr=\thisrow{accuracy}-\thisrow{low_05_ci}
            ] {\sdxl};
\addplot+ [color2bg, error bars/.cd,
            y dir=both,
            y explicit,
            error bar style={black}]
        table
          [x=index,
            y=accuracy,
            y error plus expr=\thisrow{high_05_ci}-\thisrow{accuracy},
            y error minus expr=\thisrow{accuracy}-\thisrow{low_05_ci}
            ] {\imagentwo};            
\addplot+ [color3bg, error bars/.cd,
            y dir=both,
            y explicit,
            error bar style={black}]
        table
          [x=index,
            y=accuracy,
            y error plus expr=\thisrow{high_05_ci}-\thisrow{accuracy},
            y error minus expr=\thisrow{accuracy}-\thisrow{low_05_ci}
            ] {\midjourneysix};
\addplot+ [color4bg, error bars/.cd,
            y dir=both,
            y explicit,
            error bar style={black}]
        table
          [x=index,
            y=accuracy,
            y error plus expr=\thisrow{high_05_ci}-\thisrow{accuracy},
            y error minus expr=\thisrow{accuracy}-\thisrow{low_05_ci}
            ] {\dallethree};            
\addplot+ [color5bg, error bars/.cd,
            y dir=both,
            y explicit,
            error bar style={black}]
        table
          [x=index,
            y=accuracy,
            y error plus expr=\thisrow{high_05_ci}-\thisrow{accuracy},
            y error minus expr=\thisrow{accuracy}-\thisrow{low_05_ci}
            ] {\sdthree};            
\addplot+ [color6bg, error bars/.cd,
            y dir=both,
            y explicit,
            error bar style={black}]
        table
          [x=index,
            y=accuracy,
            y error plus expr=\thisrow{high_05_ci}-\thisrow{accuracy},
            y error minus expr=\thisrow{accuracy}-\thisrow{low_05_ci}
            ] {\imagenthree};
\legend{SDXL 1,Imagen 2,MJ v6,DALL·E 3,SD 3,Imagen 3}
\end{axis}
\end{tikzpicture}

%% file: assets/human_evals/elo_imagen3_002.tex
\pgfplotstableread[col sep=comma]{
index,model,elo,low_99_ci,high_99_ci
0,SDXL 1,817,806,829
1,Imagen 2,889,880,898
2,NovaCanvas,964,957,972
3,MJ v6,982,976,989
4,DALL·E 3,997,991,1004
5,SD 3.5 L,1001,995,1008
6,Flux1.1p,1043,1037,1049
7,Imagen 3-001,1053,1047,1059
8,IdeogramV2,1059,1052,1066
9,RecraftV3,1078,1071,1085
10,Imagen 3-002,1115,1106,1124
}\modelnamesoverall

\pgfplotstableread[col sep=comma]{
index,model,elo,low_99_ci,high_99_ci
0,SDXL 1,817,806,829
2,NovaCanvas,964,957,972
3,MJ v6,982,976,989
4,DALL·E 3,997,991,1004
5,SD 3.5 L,1001,995,1008
6,Flux1.1p,1043,1037,1049
8,IdeogramV2,1059,1052,1066
9,RecraftV3,1078,1071,1085
}\datatableAoverall

\pgfplotstableread[col sep=comma]{
index,model,elo,low_99_ci,high_99_ci
1,Imagen 2,889,880,898
7,Imagen 3-001,1053,1047,1059
10,Imagen 3-002,1115,1106,1124
}\datatableBoverall

\pgfplotstableread[col sep=comma]{
index,model,elo,low_99_ci,high_99_ci
0,SDXL 1,697,685,709
1,Imagen 2,827,818,836
2,NovaCanvas,897,890,904
3,DALL·E 3,961,954,969
4,SD 3.5 L,1033,1028,1039
5,Flux1.1p,1063,1058,1068
6,MJ v6,1066,1060,1072
7,Imagen 3-001,1104,1098,1109
8,IdeogramV2,1104,1098,1110
9,RecraftV3,1112,1106,1119
10,Imagen 3-002,1135,1128,1143
}\modelnamesvisual

\pgfplotstableread[col sep=comma]{
index,model,elo,low_99_ci,high_99_ci
0,SDXL 1,697,685,709
2,NovaCanvas,897,890,904
3,DALL·E 3,961,954,969
4,SD 3.5 L,1033,1028,1039
5,Flux1.1p,1063,1058,1068
6,MJ v6,1066,1060,1072
8,IdeogramV2,1104,1098,1110
9,RecraftV3,1112,1106,1119
}\datatableAvisual

\pgfplotstableread[col sep=comma]{
index,model,elo,low_99_ci,high_99_ci
1,Imagen 2,827,818,836
7,Imagen 3-001,1104,1098,1109
10,Imagen 3-002,1135,1128,1143
}\datatableBvisual

\pgfplotstableread[col sep=comma]{
index,model,elo,low_99_ci,high_99_ci
0,SDXL 1,845,834,855
1,Imagen 2,906,897,915
2,NovaCanvas,971,963,978
3,MJ v6,982,975,989
4,DALL·E 3,1002,995,1009
5,SD 3.5 L,1004,998,1010
6,Flux1.1p,1037,1031,1043
7,Imagen 3-001,1042,1035,1048
8,IdeogramV2,1045,1038,1051
9,RecraftV3,1061,1054,1069
10,Imagen 3-002,1106,1096,1115
}\modelnamesalignment

\pgfplotstableread[col sep=comma]{
index,model,elo,low_99_ci,high_99_ci
0,SDXL 1,845,834,855
2,NovaCanvas,971,963,978
3,MJ v6,982,975,989
4,DALL·E 3,1002,995,1009
5,SD 3.5 L,1004,998,1010
6,Flux1.1p,1037,1031,1043
8,IdeogramV2,1045,1038,1051
9,RecraftV3,1061,1054,1069
}\datatableAalignment

\pgfplotstableread[col sep=comma]{
index,model,elo,low_99_ci,high_99_ci
1,Imagen 2,906,897,915
7,Imagen 3-001,1042,1035,1048
10,Imagen 3-002,1106,1096,1115
}\datatableBalignment

\begin{figure}
\centering
\vspace{-5mm}
\resizebox{\linewidth}{!}{%
\mbox{%
\begin{minipage}{0.54\linewidth}
\resizebox{\linewidth}{!}{%
\begin{tikzpicture}
\begin{axis} [
        x label style={at={(0.5,-0.1)}},
        y label style={at={(-0.07,0.5)}},
        ylabel={Elo rating (with 99\% CI)},
        title={Overall preference},
        width=15cm,
        height=12cm,
    	ybar,
    	major grid style={line width=.2pt,draw=gray!50,dashed},
    	ymajorgrids,
    	tick style={draw=none},
    	xticklabels from table={\modelnamesoverall}{model},
    	xtick={0,...,10},
    	xmin=-0.5,
    	xmax=10.5,
    	xticklabel style={yshift=0mm,xshift=3mm, align=center,rotate=30,anchor=east},
    	every node near coord/.append style={xshift=0,yshift=-22,color=black},
    	every axis plot/.append style={
          ybar,
          bar width=0.85,
          bar shift=0pt,
          draw=none
        },
        nodes near coords,
]
\addplot+ [mygray, error bars/.cd,
            y dir=both,
            y explicit,
            error bar style={black}]
        table
          [x=index,
            y=elo,
            y error plus expr=\thisrow{high_99_ci}-\thisrow{elo},
            y error minus expr=\thisrow{elo}-\thisrow{low_99_ci}
            ] {\datatableAoverall};
\addplot+ [myblue, error bars/.cd,
            y dir=both,
            y explicit,
            error bar style={black}]
        table
          [x=index,
            y=elo,
            y error plus expr=\thisrow{high_99_ci}-\thisrow{elo},
            y error minus expr=\thisrow{elo}-\thisrow{low_99_ci}
            ] {\datatableBoverall};     

\end{axis}
\end{tikzpicture}}
\end{minipage}
\hspace{2mm}
\begin{minipage}{0.46\linewidth}
\resizebox{\linewidth}{!}{%
\begin{tikzpicture}[scale=1.0]
\draw[very thin] (-0.5,10.5) rectangle +(1,1);
\draw[very thin, fill={rgb,1:red,0.995;green,0.809;blue,0.487}] (-0.5,9.5) rectangle +(1,1);
\node[black] at (0,10) {44.8\strut};
\draw[very thin, fill={rgb,1:red,0.983;green,0.617;blue,0.350}] (-0.5,8.5) rectangle +(1,1);
\node[black] at (0,9) {41.2\strut};
\draw[very thin, fill={rgb,1:red,0.983;green,0.617;blue,0.350}] (-0.5,7.5) rectangle +(1,1);
\node[black] at (0,8) {41.2\strut};
\draw[very thin, fill={rgb,1:red,0.976;green,0.567;blue,0.327}] (-0.5,6.5) rectangle +(1,1);
\node[black] at (0,7) {40.5\strut};
\draw[very thin] (-0.5,5.5) rectangle +(1,1);
\draw[very thin] (-0.5,4.5) rectangle +(1,1);
\draw[very thin] (-0.5,3.5) rectangle +(1,1);
\draw[very thin] (-0.5,2.5) rectangle +(1,1);
\draw[very thin] (-0.5,1.5) rectangle +(1,1);
\draw[very thin] (-0.5,0.5) rectangle +(1,1);
\draw[very thin, fill={rgb,1:red,0.780;green,0.907;blue,0.499}] (0.5,10.5) rectangle +(1,1);
\node[black] at (1,11) {55.2\strut};
\draw[very thin] (0.5,9.5) rectangle +(1,1);
\draw[very thin, fill={rgb,1:red,0.999;green,0.974;blue,0.705}] (0.5,8.5) rectangle +(1,1);
\node[black] at (1,9) {49.2\strut};
\draw[very thin, fill={rgb,1:red,0.997;green,0.907;blue,0.593}] (0.5,7.5) rectangle +(1,1);
\node[black] at (1,8) {47.1\strut};
\draw[very thin, fill={rgb,1:red,0.994;green,0.763;blue,0.448}] (0.5,6.5) rectangle +(1,1);
\node[black] at (1,7) {43.9\strut};
\draw[very thin, fill={rgb,1:red,0.952;green,0.418;blue,0.258}] (0.5,5.5) rectangle +(1,1);
\node[white] at (1,6) {38.2\strut};
\draw[very thin] (0.5,4.5) rectangle +(1,1);
\draw[very thin] (0.5,3.5) rectangle +(1,1);
\draw[very thin] (0.5,2.5) rectangle +(1,1);
\draw[very thin] (0.5,1.5) rectangle +(1,1);
\draw[very thin] (0.5,0.5) rectangle +(1,1);
\draw[very thin, fill={rgb,1:red,0.587;green,0.823;blue,0.409}] (1.5,10.5) rectangle +(1,1);
\node[black] at (2,11) {58.8\strut};
\draw[very thin, fill={rgb,1:red,0.968;green,0.986;blue,0.705}] (1.5,9.5) rectangle +(1,1);
\node[black] at (2,10) {50.8\strut};
\draw[very thin] (1.5,8.5) rectangle +(1,1);
\draw[very thin, fill={rgb,1:red,1.000;green,0.993;blue,0.737}] (1.5,7.5) rectangle +(1,1);
\node[black] at (2,8) {49.8\strut};
\draw[very thin, fill={rgb,1:red,0.999;green,0.969;blue,0.697}] (1.5,6.5) rectangle +(1,1);
\node[black] at (2,7) {49.0\strut};
\draw[very thin, fill={rgb,1:red,0.979;green,0.587;blue,0.337}] (1.5,5.5) rectangle +(1,1);
\node[black] at (2,6) {40.8\strut};
\draw[very thin, fill={rgb,1:red,0.978;green,0.577;blue,0.332}] (1.5,4.5) rectangle +(1,1);
\node[black] at (2,5) {40.6\strut};
\draw[very thin] (1.5,3.5) rectangle +(1,1);
\draw[very thin] (1.5,2.5) rectangle +(1,1);
\draw[very thin] (1.5,1.5) rectangle +(1,1);
\draw[very thin] (1.5,0.5) rectangle +(1,1);
\draw[very thin, fill={rgb,1:red,0.587;green,0.823;blue,0.409}] (2.5,10.5) rectangle +(1,1);
\node[black] at (3,11) {58.8\strut};
\draw[very thin, fill={rgb,1:red,0.886;green,0.952;blue,0.593}] (2.5,9.5) rectangle +(1,1);
\node[black] at (3,10) {52.9\strut};
\draw[very thin, fill={rgb,1:red,0.991;green,0.996;blue,0.737}] (2.5,8.5) rectangle +(1,1);
\node[black] at (3,9) {50.2\strut};
\draw[very thin] (2.5,7.5) rectangle +(1,1);
\draw[very thin, fill={rgb,1:red,0.999;green,0.983;blue,0.721}] (2.5,6.5) rectangle +(1,1);
\node[black] at (3,7) {49.4\strut};
\draw[very thin, fill={rgb,1:red,0.994;green,0.763;blue,0.448}] (2.5,5.5) rectangle +(1,1);
\node[black] at (3,6) {43.8\strut};
\draw[very thin, fill={rgb,1:red,0.993;green,0.709;blue,0.403}] (2.5,4.5) rectangle +(1,1);
\node[black] at (3,5) {42.7\strut};
\draw[very thin, fill={rgb,1:red,0.980;green,0.597;blue,0.341}] (2.5,3.5) rectangle +(1,1);
\node[black] at (3,4) {41.0\strut};
\draw[very thin] (2.5,2.5) rectangle +(1,1);
\draw[very thin] (2.5,1.5) rectangle +(1,1);
\draw[very thin] (2.5,0.5) rectangle +(1,1);
\draw[very thin, fill={rgb,1:red,0.538;green,0.801;blue,0.403}] (3.5,10.5) rectangle +(1,1);
\node[black] at (4,11) {59.5\strut};
\draw[very thin, fill={rgb,1:red,0.733;green,0.887;blue,0.469}] (3.5,9.5) rectangle +(1,1);
\node[black] at (4,10) {56.1\strut};
\draw[very thin, fill={rgb,1:red,0.962;green,0.984;blue,0.697}] (3.5,8.5) rectangle +(1,1);
\node[black] at (4,9) {51.0\strut};
\draw[very thin, fill={rgb,1:red,0.980;green,0.991;blue,0.721}] (3.5,7.5) rectangle +(1,1);
\node[black] at (4,8) {50.6\strut};
\draw[very thin] (3.5,6.5) rectangle +(1,1);
\draw[very thin, fill={rgb,1:red,0.996;green,0.855;blue,0.526}] (3.5,5.5) rectangle +(1,1);
\node[black] at (4,6) {45.6\strut};
\draw[very thin, fill={rgb,1:red,0.994;green,0.778;blue,0.461}] (3.5,4.5) rectangle +(1,1);
\node[black] at (4,5) {44.2\strut};
\draw[very thin, fill={rgb,1:red,0.973;green,0.547;blue,0.318}] (3.5,3.5) rectangle +(1,1);
\node[black] at (4,4) {40.2\strut};
\draw[very thin, fill={rgb,1:red,0.969;green,0.517;blue,0.304}] (3.5,2.5) rectangle +(1,1);
\node[black] at (4,3) {39.7\strut};
\draw[very thin] (3.5,1.5) rectangle +(1,1);
\draw[very thin] (3.5,0.5) rectangle +(1,1);
\draw[very thin] (4.5,10.5) rectangle +(1,1);
\draw[very thin, fill={rgb,1:red,0.388;green,0.735;blue,0.385}] (4.5,9.5) rectangle +(1,1);
\node[white] at (5,10) {61.8\strut};
\draw[very thin, fill={rgb,1:red,0.557;green,0.810;blue,0.405}] (4.5,8.5) rectangle +(1,1);
\node[black] at (5,9) {59.2\strut};
\draw[very thin, fill={rgb,1:red,0.733;green,0.887;blue,0.469}] (4.5,7.5) rectangle +(1,1);
\node[black] at (5,8) {56.2\strut};
\draw[very thin, fill={rgb,1:red,0.827;green,0.927;blue,0.530}] (4.5,6.5) rectangle +(1,1);
\node[black] at (5,7) {54.4\strut};
\draw[very thin] (4.5,5.5) rectangle +(1,1);
\draw[very thin, fill={rgb,1:red,1.000;green,0.988;blue,0.729}] (4.5,4.5) rectangle +(1,1);
\node[black] at (5,5) {49.6\strut};
\draw[very thin, fill={rgb,1:red,0.996;green,0.871;blue,0.539}] (4.5,3.5) rectangle +(1,1);
\node[black] at (5,4) {45.9\strut};
\draw[very thin, fill={rgb,1:red,0.996;green,0.871;blue,0.539}] (4.5,2.5) rectangle +(1,1);
\node[black] at (5,3) {45.9\strut};
\draw[very thin, fill={rgb,1:red,0.845;green,0.193;blue,0.155}] (4.5,1.5) rectangle +(1,1);
\node[white] at (5,2) {34.6\strut};
\draw[very thin] (4.5,0.5) rectangle +(1,1);
\draw[very thin] (5.5,10.5) rectangle +(1,1);
\draw[very thin] (5.5,9.5) rectangle +(1,1);
\draw[very thin, fill={rgb,1:red,0.548;green,0.806;blue,0.404}] (5.5,8.5) rectangle +(1,1);
\node[black] at (6,9) {59.4\strut};
\draw[very thin, fill={rgb,1:red,0.678;green,0.863;blue,0.433}] (5.5,7.5) rectangle +(1,1);
\node[black] at (6,8) {57.3\strut};
\draw[very thin, fill={rgb,1:red,0.749;green,0.893;blue,0.479}] (5.5,6.5) rectangle +(1,1);
\node[black] at (6,7) {55.8\strut};
\draw[very thin, fill={rgb,1:red,0.985;green,0.994;blue,0.729}] (5.5,5.5) rectangle +(1,1);
\node[black] at (6,6) {50.4\strut};
\draw[very thin] (5.5,4.5) rectangle +(1,1);
\draw[very thin, fill={rgb,1:red,1.000;green,0.998;blue,0.745}] (5.5,3.5) rectangle +(1,1);
\node[black] at (6,4) {49.9\strut};
\draw[very thin, fill={rgb,1:red,0.985;green,0.627;blue,0.355}] (5.5,2.5) rectangle +(1,1);
\node[black] at (6,3) {41.4\strut};
\draw[very thin, fill={rgb,1:red,0.908;green,0.324;blue,0.215}] (5.5,1.5) rectangle +(1,1);
\node[white] at (6,2) {36.7\strut};
\draw[very thin, fill={rgb,1:red,0.716;green,0.066;blue,0.150}] (5.5,0.5) rectangle +(1,1);
\node[white] at (6,1) {31.9\strut};
\draw[very thin] (6.5,10.5) rectangle +(1,1);
\draw[very thin] (6.5,9.5) rectangle +(1,1);
\draw[very thin] (6.5,8.5) rectangle +(1,1);
\draw[very thin, fill={rgb,1:red,0.567;green,0.814;blue,0.407}] (6.5,7.5) rectangle +(1,1);
\node[black] at (7,8) {59.0\strut};
\draw[very thin, fill={rgb,1:red,0.518;green,0.793;blue,0.401}] (6.5,6.5) rectangle +(1,1);
\node[black] at (7,7) {59.8\strut};
\draw[very thin, fill={rgb,1:red,0.843;green,0.934;blue,0.540}] (6.5,5.5) rectangle +(1,1);
\node[black] at (7,6) {54.1\strut};
\draw[very thin, fill={rgb,1:red,0.997;green,0.999;blue,0.745}] (6.5,4.5) rectangle +(1,1);
\node[black] at (7,5) {50.1\strut};
\draw[very thin] (6.5,3.5) rectangle +(1,1);
\draw[very thin, fill={rgb,1:red,0.998;green,0.931;blue,0.633}] (6.5,2.5) rectangle +(1,1);
\node[black] at (7,3) {47.8\strut};
\draw[very thin, fill={rgb,1:red,0.980;green,0.597;blue,0.341}] (6.5,1.5) rectangle +(1,1);
\node[black] at (7,2) {40.9\strut};
\draw[very thin, fill={rgb,1:red,0.647;green,0.000;blue,0.149}] (6.5,0.5) rectangle +(1,1);
\node[white] at (7,1) {30.5\strut};
\draw[very thin] (7.5,10.5) rectangle +(1,1);
\draw[very thin] (7.5,9.5) rectangle +(1,1);
\draw[very thin] (7.5,8.5) rectangle +(1,1);
\draw[very thin] (7.5,7.5) rectangle +(1,1);
\draw[very thin, fill={rgb,1:red,0.489;green,0.780;blue,0.398}] (7.5,6.5) rectangle +(1,1);
\node[black] at (8,7) {60.3\strut};
\draw[very thin, fill={rgb,1:red,0.843;green,0.934;blue,0.540}] (7.5,5.5) rectangle +(1,1);
\node[black] at (8,6) {54.1\strut};
\draw[very thin, fill={rgb,1:red,0.597;green,0.827;blue,0.410}] (7.5,4.5) rectangle +(1,1);
\node[black] at (8,5) {58.6\strut};
\draw[very thin, fill={rgb,1:red,0.915;green,0.964;blue,0.633}] (7.5,3.5) rectangle +(1,1);
\node[black] at (8,4) {52.2\strut};
\draw[very thin] (7.5,2.5) rectangle +(1,1);
\draw[very thin, fill={rgb,1:red,0.939;green,0.390;blue,0.246}] (7.5,1.5) rectangle +(1,1);
\node[white] at (8,2) {37.8\strut};
\draw[very thin, fill={rgb,1:red,0.670;green,0.022;blue,0.149}] (7.5,0.5) rectangle +(1,1);
\node[white] at (8,1) {31.1\strut};
\draw[very thin] (8.5,10.5) rectangle +(1,1);
\draw[very thin] (8.5,9.5) rectangle +(1,1);
\draw[very thin] (8.5,8.5) rectangle +(1,1);
\draw[very thin] (8.5,7.5) rectangle +(1,1);
\draw[very thin] (8.5,6.5) rectangle +(1,1);
\draw[very thin, fill={rgb,1:red,0.108;green,0.599;blue,0.315}] (8.5,5.5) rectangle +(1,1);
\node[white] at (9,6) {65.4\strut};
\draw[very thin, fill={rgb,1:red,0.271;green,0.679;blue,0.356}] (8.5,4.5) rectangle +(1,1);
\node[white] at (9,5) {63.3\strut};
\draw[very thin, fill={rgb,1:red,0.567;green,0.814;blue,0.407}] (8.5,3.5) rectangle +(1,1);
\node[black] at (9,4) {59.1\strut};
\draw[very thin, fill={rgb,1:red,0.353;green,0.718;blue,0.377}] (8.5,2.5) rectangle +(1,1);
\node[white] at (9,3) {62.2\strut};
\draw[very thin] (8.5,1.5) rectangle +(1,1);
\draw[very thin, fill={rgb,1:red,0.989;green,0.657;blue,0.369}] (8.5,0.5) rectangle +(1,1);
\node[black] at (9,1) {41.8\strut};
\draw[very thin] (9.5,10.5) rectangle +(1,1);
\draw[very thin] (9.5,9.5) rectangle +(1,1);
\draw[very thin] (9.5,8.5) rectangle +(1,1);
\draw[very thin] (9.5,7.5) rectangle +(1,1);
\draw[very thin] (9.5,6.5) rectangle +(1,1);
\draw[very thin] (9.5,5.5) rectangle +(1,1);
\draw[very thin, fill={rgb,1:red,0.036;green,0.474;blue,0.250}] (9.5,4.5) rectangle +(1,1);
\node[white] at (10,5) {68.1\strut};
\draw[very thin, fill={rgb,1:red,0.000;green,0.408;blue,0.216}] (9.5,3.5) rectangle +(1,1);
\node[white] at (10,4) {69.5\strut};
\draw[very thin, fill={rgb,1:red,0.012;green,0.430;blue,0.227}] (9.5,2.5) rectangle +(1,1);
\node[white] at (10,3) {68.9\strut};
\draw[very thin, fill={rgb,1:red,0.626;green,0.840;blue,0.413}] (9.5,1.5) rectangle +(1,1);
\node[black] at (10,2) {58.2\strut};
\draw[very thin] (9.5,0.5) rectangle +(1,1);

\node[rotate=35, anchor=west, align=left] at (-0.1,11.6) {Imagen 3-002};
\node[anchor=east, align=right] at (-0.5,11) {Imagen 3-002};
\node[rotate=35, anchor=west, align=left] at (0.9,11.6) {RecraftV3};
\node[anchor=east, align=right] at (-0.5,10) {RecraftV3};
\node[rotate=35, anchor=west, align=left] at (1.9,11.6) {IdeogramV2};
\node[anchor=east, align=right] at (-0.5,9) {IdeogramV2};
\node[rotate=35, anchor=west, align=left] at (2.9,11.6) {Imagen 3-001};
\node[anchor=east, align=right] at (-0.5,8) {Imagen 3-001};
\node[rotate=35, anchor=west, align=left] at (3.9,11.6) {Flux1.1p};
\node[anchor=east, align=right] at (-0.5,7) {Flux1.1p};
\node[rotate=35, anchor=west, align=left] at (4.9,11.6) {SD 3.5 L};
\node[anchor=east, align=right] at (-0.5,6) {SD 3.5 L};
\node[rotate=35, anchor=west, align=left] at (5.9,11.6) {DALL·E 3};
\node[anchor=east, align=right] at (-0.5,5) {DALL·E 3};
\node[rotate=35, anchor=west, align=left] at (6.9,11.6) {MJ v6};
\node[anchor=east, align=right] at (-0.5,4) {MJ v6};
\node[rotate=35, anchor=west, align=left] at (7.9,11.6) {NovaCanvas};
\node[anchor=east, align=right] at (-0.5,3) {NovaCanvas};
\node[rotate=35, anchor=west, align=left] at (8.9,11.6) {Imagen 2};
\node[anchor=east, align=right] at (-0.5,2) {Imagen 2};
\node[rotate=35, anchor=west, align=left] at (9.9,11.6) {SDXL 1};
\node[anchor=east, align=right] at (-0.5,1) {SDXL 1};

\node[anchor=east] at (0,-1) {};
\end{tikzpicture}}
\end{minipage}}}
\\[1mm]
\resizebox{\linewidth}{!}{%
\mbox{%
\begin{minipage}{0.54\linewidth}
\resizebox{\linewidth}{!}{%
\begin{tikzpicture}
\begin{axis} [
        x label style={at={(0.5,-0.1)}},
        y label style={at={(-0.07,0.5)}},
        ylabel={Elo rating (with 99\% CI)},
        title={Visual quality},
        width=15cm,
        height=12cm,
    	ybar,
    	major grid style={line width=.2pt,draw=gray!50,dashed},
    	ymajorgrids,
    	tick style={draw=none},
    	xticklabels from table={\modelnamesvisual}{model},
    	xtick={0,...,10},
    	xmin=-0.5,
    	xmax=10.5,
    	xticklabel style={yshift=0mm,xshift=3mm, align=center,rotate=30,anchor=east},
    	every node near coord/.append style={xshift=0,yshift=-22,color=black,font=\small},
    	every axis plot/.append style={
          ybar,
          bar width=0.85,
          bar shift=0pt,
          draw=none
        },
        nodes near coords,
]
\addplot+ [mygray, error bars/.cd,
            y dir=both,
            y explicit,
            error bar style={black}]
        table
          [x=index,
            y=elo,
            y error plus expr=\thisrow{high_99_ci}-\thisrow{elo},
            y error minus expr=\thisrow{elo}-\thisrow{low_99_ci}
            ] {\datatableAvisual};
\addplot+ [myblue, error bars/.cd,
            y dir=both,
            y explicit,
            error bar style={black}]
        table
          [x=index,
            y=elo,
            y error plus expr=\thisrow{high_99_ci}-\thisrow{elo},
            y error minus expr=\thisrow{elo}-\thisrow{low_99_ci}
            ] {\datatableBvisual};     

\end{axis}
\end{tikzpicture}}
\end{minipage}
\hspace{2mm}
\begin{minipage}{0.46\linewidth}
\resizebox{\linewidth}{!}{%
\begin{tikzpicture}[scale=1]
\draw[very thin] (-0.5,10.5) rectangle +(1,1);
\draw[very thin, fill={rgb,1:red,0.998;green,0.945;blue,0.657}] (-0.5,9.5) rectangle +(1,1);
\node[black] at (0,10) {47.3\strut};
\draw[very thin, fill={rgb,1:red,0.997;green,0.921;blue,0.617}] (-0.5,8.5) rectangle +(1,1);
\node[black] at (0,9) {46.2\strut};
\draw[very thin, fill={rgb,1:red,0.998;green,0.936;blue,0.641}] (-0.5,7.5) rectangle +(1,1);
\node[black] at (0,8) {46.8\strut};
\draw[very thin, fill={rgb,1:red,0.993;green,0.732;blue,0.422}] (-0.5,6.5) rectangle +(1,1);
\node[black] at (0,7) {39.5\strut};
\draw[very thin] (-0.5,5.5) rectangle +(1,1);
\draw[very thin] (-0.5,4.5) rectangle +(1,1);
\draw[very thin] (-0.5,3.5) rectangle +(1,1);
\draw[very thin] (-0.5,2.5) rectangle +(1,1);
\draw[very thin] (-0.5,1.5) rectangle +(1,1);
\draw[very thin] (-0.5,0.5) rectangle +(1,1);
\draw[very thin, fill={rgb,1:red,0.933;green,0.972;blue,0.657}] (0.5,10.5) rectangle +(1,1);
\node[black] at (1,11) {52.7\strut};
\draw[very thin] (0.5,9.5) rectangle +(1,1);
\draw[very thin, fill={rgb,1:red,0.980;green,0.991;blue,0.721}] (0.5,8.5) rectangle +(1,1);
\node[black] at (1,9) {50.9\strut};
\draw[very thin, fill={rgb,1:red,0.980;green,0.991;blue,0.721}] (0.5,7.5) rectangle +(1,1);
\node[black] at (1,8) {50.9\strut};
\draw[very thin, fill={rgb,1:red,0.995;green,0.825;blue,0.500}] (0.5,6.5) rectangle +(1,1);
\node[black] at (1,7) {42.4\strut};
\draw[very thin, fill={rgb,1:red,0.994;green,0.778;blue,0.461}] (0.5,5.5) rectangle +(1,1);
\node[black] at (1,6) {41.0\strut};
\draw[very thin] (0.5,4.5) rectangle +(1,1);
\draw[very thin] (0.5,3.5) rectangle +(1,1);
\draw[very thin] (0.5,2.5) rectangle +(1,1);
\draw[very thin] (0.5,1.5) rectangle +(1,1);
\draw[very thin] (0.5,0.5) rectangle +(1,1);
\draw[very thin, fill={rgb,1:red,0.904;green,0.959;blue,0.617}] (1.5,10.5) rectangle +(1,1);
\node[black] at (2,11) {53.8\strut};
\draw[very thin, fill={rgb,1:red,0.999;green,0.983;blue,0.721}] (1.5,9.5) rectangle +(1,1);
\node[black] at (2,10) {49.1\strut};
\draw[very thin] (1.5,8.5) rectangle +(1,1);
\draw[very thin, fill={rgb,1:red,0.999;green,0.969;blue,0.697}] (1.5,7.5) rectangle +(1,1);
\node[black] at (2,8) {48.5\strut};
\draw[very thin, fill={rgb,1:red,0.991;green,0.996;blue,0.737}] (1.5,6.5) rectangle +(1,1);
\node[black] at (2,7) {50.4\strut};
\draw[very thin, fill={rgb,1:red,0.999;green,0.955;blue,0.673}] (1.5,5.5) rectangle +(1,1);
\node[black] at (2,6) {47.8\strut};
\draw[very thin, fill={rgb,1:red,0.993;green,0.725;blue,0.416}] (1.5,4.5) rectangle +(1,1);
\node[black] at (2,5) {39.3\strut};
\draw[very thin] (1.5,3.5) rectangle +(1,1);
\draw[very thin] (1.5,2.5) rectangle +(1,1);
\draw[very thin] (1.5,1.5) rectangle +(1,1);
\draw[very thin] (1.5,0.5) rectangle +(1,1);
\draw[very thin, fill={rgb,1:red,0.921;green,0.967;blue,0.641}] (2.5,10.5) rectangle +(1,1);
\node[black] at (3,11) {53.2\strut};
\draw[very thin, fill={rgb,1:red,0.999;green,0.983;blue,0.721}] (2.5,9.5) rectangle +(1,1);
\node[black] at (3,10) {49.1\strut};
\draw[very thin, fill={rgb,1:red,0.962;green,0.984;blue,0.697}] (2.5,8.5) rectangle +(1,1);
\node[black] at (3,9) {51.5\strut};
\draw[very thin] (2.5,7.5) rectangle +(1,1);
\draw[very thin, fill={rgb,1:red,0.968;green,0.986;blue,0.705}] (2.5,6.5) rectangle +(1,1);
\node[black] at (3,7) {51.3\strut};
\draw[very thin, fill={rgb,1:red,0.996;green,0.863;blue,0.532}] (2.5,5.5) rectangle +(1,1);
\node[black] at (3,6) {43.6\strut};
\draw[very thin, fill={rgb,1:red,0.994;green,0.771;blue,0.455}] (2.5,4.5) rectangle +(1,1);
\node[black] at (3,5) {40.6\strut};
\draw[very thin, fill={rgb,1:red,0.965;green,0.487;blue,0.290}] (2.5,3.5) rectangle +(1,1);
\node[black] at (3,4) {33.5\strut};
\draw[very thin] (2.5,2.5) rectangle +(1,1);
\draw[very thin] (2.5,1.5) rectangle +(1,1);
\draw[very thin] (2.5,0.5) rectangle +(1,1);
\draw[very thin, fill={rgb,1:red,0.702;green,0.873;blue,0.449}] (3.5,10.5) rectangle +(1,1);
\node[black] at (4,11) {60.5\strut};
\draw[very thin, fill={rgb,1:red,0.796;green,0.914;blue,0.510}] (3.5,9.5) rectangle +(1,1);
\node[black] at (4,10) {57.6\strut};
\draw[very thin, fill={rgb,1:red,1.000;green,0.993;blue,0.737}] (3.5,8.5) rectangle +(1,1);
\node[black] at (4,9) {49.6\strut};
\draw[very thin, fill={rgb,1:red,0.999;green,0.974;blue,0.705}] (3.5,7.5) rectangle +(1,1);
\node[black] at (4,8) {48.7\strut};
\draw[very thin] (3.5,6.5) rectangle +(1,1);
\draw[very thin, fill={rgb,1:red,0.939;green,0.974;blue,0.665}] (3.5,5.5) rectangle +(1,1);
\node[black] at (4,6) {52.4\strut};
\draw[very thin, fill={rgb,1:red,1.000;green,0.988;blue,0.729}] (3.5,4.5) rectangle +(1,1);
\node[black] at (4,5) {49.3\strut};
\draw[very thin, fill={rgb,1:red,0.969;green,0.517;blue,0.304}] (3.5,3.5) rectangle +(1,1);
\node[black] at (4,4) {34.1\strut};
\draw[very thin, fill={rgb,1:red,0.872;green,0.249;blue,0.181}] (3.5,2.5) rectangle +(1,1);
\node[white] at (4,3) {27.5\strut};
\draw[very thin] (3.5,1.5) rectangle +(1,1);
\draw[very thin] (3.5,0.5) rectangle +(1,1);
\draw[very thin] (4.5,10.5) rectangle +(1,1);
\draw[very thin, fill={rgb,1:red,0.749;green,0.893;blue,0.479}] (4.5,9.5) rectangle +(1,1);
\node[black] at (5,10) {59.0\strut};
\draw[very thin, fill={rgb,1:red,0.944;green,0.977;blue,0.673}] (4.5,8.5) rectangle +(1,1);
\node[black] at (5,9) {52.2\strut};
\draw[very thin, fill={rgb,1:red,0.835;green,0.930;blue,0.535}] (4.5,7.5) rectangle +(1,1);
\node[black] at (5,8) {56.4\strut};
\draw[very thin, fill={rgb,1:red,0.998;green,0.950;blue,0.665}] (4.5,6.5) rectangle +(1,1);
\node[black] at (5,7) {47.6\strut};
\draw[very thin] (4.5,5.5) rectangle +(1,1);
\draw[very thin, fill={rgb,1:red,0.999;green,0.983;blue,0.721}] (4.5,4.5) rectangle +(1,1);
\node[black] at (5,5) {49.2\strut};
\draw[very thin, fill={rgb,1:red,0.980;green,0.597;blue,0.341}] (4.5,3.5) rectangle +(1,1);
\node[black] at (5,4) {36.0\strut};
\draw[very thin, fill={rgb,1:red,0.983;green,0.617;blue,0.350}] (4.5,2.5) rectangle +(1,1);
\node[black] at (5,3) {36.4\strut};
\draw[very thin, fill={rgb,1:red,0.958;green,0.437;blue,0.267}] (4.5,1.5) rectangle +(1,1);
\node[black] at (5,2) {32.2\strut};
\draw[very thin] (4.5,0.5) rectangle +(1,1);
\draw[very thin] (5.5,10.5) rectangle +(1,1);
\draw[very thin] (5.5,9.5) rectangle +(1,1);
\draw[very thin, fill={rgb,1:red,0.694;green,0.870;blue,0.444}] (5.5,8.5) rectangle +(1,1);
\node[black] at (6,9) {60.7\strut};
\draw[very thin, fill={rgb,1:red,0.741;green,0.890;blue,0.474}] (5.5,7.5) rectangle +(1,1);
\node[black] at (6,8) {59.4\strut};
\draw[very thin, fill={rgb,1:red,0.985;green,0.994;blue,0.729}] (5.5,6.5) rectangle +(1,1);
\node[black] at (6,7) {50.7\strut};
\draw[very thin, fill={rgb,1:red,0.980;green,0.991;blue,0.721}] (5.5,5.5) rectangle +(1,1);
\node[black] at (6,6) {50.8\strut};
\draw[very thin] (5.5,4.5) rectangle +(1,1);
\draw[very thin, fill={rgb,1:red,0.997;green,0.898;blue,0.577}] (5.5,3.5) rectangle +(1,1);
\node[black] at (6,4) {45.0\strut};
\draw[very thin, fill={rgb,1:red,0.983;green,0.617;blue,0.350}] (5.5,2.5) rectangle +(1,1);
\node[black] at (6,3) {36.4\strut};
\draw[very thin, fill={rgb,1:red,0.912;green,0.334;blue,0.220}] (5.5,1.5) rectangle +(1,1);
\node[white] at (6,2) {29.6\strut};
\draw[very thin, fill={rgb,1:red,0.647;green,0.000;blue,0.149}] (5.5,0.5) rectangle +(1,1);
\node[white] at (6,1) {19.9\strut};
\draw[very thin] (6.5,10.5) rectangle +(1,1);
\draw[very thin] (6.5,9.5) rectangle +(1,1);
\draw[very thin] (6.5,8.5) rectangle +(1,1);
\draw[very thin, fill={rgb,1:red,0.459;green,0.767;blue,0.395}] (6.5,7.5) rectangle +(1,1);
\node[black] at (7,8) {66.5\strut};
\draw[very thin, fill={rgb,1:red,0.489;green,0.780;blue,0.398}] (6.5,6.5) rectangle +(1,1);
\node[black] at (7,7) {65.9\strut};
\draw[very thin, fill={rgb,1:red,0.567;green,0.814;blue,0.407}] (6.5,5.5) rectangle +(1,1);
\node[black] at (7,6) {64.0\strut};
\draw[very thin, fill={rgb,1:red,0.874;green,0.947;blue,0.577}] (6.5,4.5) rectangle +(1,1);
\node[black] at (7,5) {55.0\strut};
\draw[very thin] (6.5,3.5) rectangle +(1,1);
\draw[very thin, fill={rgb,1:red,0.976;green,0.567;blue,0.327}] (6.5,2.5) rectangle +(1,1);
\node[black] at (7,3) {35.3\strut};
\draw[very thin, fill={rgb,1:red,0.997;green,0.917;blue,0.609}] (6.5,1.5) rectangle +(1,1);
\node[black] at (7,2) {45.9\strut};
\draw[very thin, fill={rgb,1:red,0.890;green,0.287;blue,0.198}] (6.5,0.5) rectangle +(1,1);
\node[white] at (7,1) {28.6\strut};
\draw[very thin] (7.5,10.5) rectangle +(1,1);
\draw[very thin] (7.5,9.5) rectangle +(1,1);
\draw[very thin] (7.5,8.5) rectangle +(1,1);
\draw[very thin] (7.5,7.5) rectangle +(1,1);
\draw[very thin, fill={rgb,1:red,0.178;green,0.633;blue,0.333}] (7.5,6.5) rectangle +(1,1);
\node[white] at (8,7) {72.5\strut};
\draw[very thin, fill={rgb,1:red,0.587;green,0.823;blue,0.409}] (7.5,5.5) rectangle +(1,1);
\node[black] at (8,6) {63.6\strut};
\draw[very thin, fill={rgb,1:red,0.587;green,0.823;blue,0.409}] (7.5,4.5) rectangle +(1,1);
\node[black] at (8,5) {63.6\strut};
\draw[very thin, fill={rgb,1:red,0.538;green,0.801;blue,0.403}] (7.5,3.5) rectangle +(1,1);
\node[black] at (8,4) {64.7\strut};
\draw[very thin] (7.5,2.5) rectangle +(1,1);
\draw[very thin, fill={rgb,1:red,0.987;green,0.647;blue,0.364}] (7.5,1.5) rectangle +(1,1);
\node[black] at (8,2) {37.3\strut};
\draw[very thin, fill={rgb,1:red,0.839;green,0.185;blue,0.153}] (7.5,0.5) rectangle +(1,1);
\node[white] at (8,1) {25.9\strut};
\draw[very thin] (8.5,10.5) rectangle +(1,1);
\draw[very thin] (8.5,9.5) rectangle +(1,1);
\draw[very thin] (8.5,8.5) rectangle +(1,1);
\draw[very thin] (8.5,7.5) rectangle +(1,1);
\draw[very thin] (8.5,6.5) rectangle +(1,1);
\draw[very thin, fill={rgb,1:red,0.410;green,0.745;blue,0.389}] (8.5,5.5) rectangle +(1,1);
\node[black] at (9,6) {67.8\strut};
\draw[very thin, fill={rgb,1:red,0.283;green,0.684;blue,0.359}] (8.5,4.5) rectangle +(1,1);
\node[white] at (9,5) {70.4\strut};
\draw[very thin, fill={rgb,1:red,0.898;green,0.957;blue,0.609}] (8.5,3.5) rectangle +(1,1);
\node[black] at (9,4) {54.1\strut};
\draw[very thin, fill={rgb,1:red,0.617;green,0.836;blue,0.412}] (8.5,2.5) rectangle +(1,1);
\node[black] at (9,3) {62.7\strut};
\draw[very thin] (8.5,1.5) rectangle +(1,1);
\draw[very thin, fill={rgb,1:red,0.993;green,0.732;blue,0.422}] (8.5,0.5) rectangle +(1,1);
\node[black] at (9,1) {39.5\strut};
\draw[very thin] (9.5,10.5) rectangle +(1,1);
\draw[very thin] (9.5,9.5) rectangle +(1,1);
\draw[very thin] (9.5,8.5) rectangle +(1,1);
\draw[very thin] (9.5,7.5) rectangle +(1,1);
\draw[very thin] (9.5,6.5) rectangle +(1,1);
\draw[very thin] (9.5,5.5) rectangle +(1,1);
\draw[very thin, fill={rgb,1:red,0.000;green,0.408;blue,0.216}] (9.5,4.5) rectangle +(1,1);
\node[white] at (10,5) {80.1\strut};
\draw[very thin, fill={rgb,1:red,0.225;green,0.656;blue,0.344}] (9.5,3.5) rectangle +(1,1);
\node[white] at (10,4) {71.4\strut};
\draw[very thin, fill={rgb,1:red,0.100;green,0.592;blue,0.312}] (9.5,2.5) rectangle +(1,1);
\node[white] at (10,3) {74.1\strut};
\draw[very thin, fill={rgb,1:red,0.702;green,0.873;blue,0.449}] (9.5,1.5) rectangle +(1,1);
\node[black] at (10,2) {60.5\strut};
\draw[very thin] (9.5,0.5) rectangle +(1,1);

\node[rotate=35, anchor=west, align=left] at (-0.1,11.6) {Imagen 3-002};
\node[anchor=east, align=right] at (-0.5,11) {Imagen 3-002};
\node[rotate=35, anchor=west, align=left] at (0.9,11.6) {RecraftV3};
\node[anchor=east, align=right] at (-0.5,10) {RecraftV3};
\node[rotate=35, anchor=west, align=left] at (1.9,11.6) {IdeogramV2};
\node[anchor=east, align=right] at (-0.5,9) {IdeogramV2};
\node[rotate=35, anchor=west, align=left] at (2.9,11.6) {Imagen 3-001};
\node[anchor=east, align=right] at (-0.5,8) {Imagen 3-001};
\node[rotate=35, anchor=west, align=left] at (3.9,11.6) {MJ v6};
\node[anchor=east, align=right] at (-0.5,7) {MJ v6};
\node[rotate=35, anchor=west, align=left] at (4.9,11.6) {Flux1.1p};
\node[anchor=east, align=right] at (-0.5,6) {Flux1.1p};
\node[rotate=35, anchor=west, align=left] at (5.9,11.6) {SD 3.5 L};
\node[anchor=east, align=right] at (-0.5,5) {SD 3.5 L};
\node[rotate=35, anchor=west, align=left] at (6.9,11.6) {DALL·E 3};
\node[anchor=east, align=right] at (-0.5,4) {DALL·E 3};
\node[rotate=35, anchor=west, align=left] at (7.9,11.6) {NovaCanvas};
\node[anchor=east, align=right] at (-0.5,3) {NovaCanvas};
\node[rotate=35, anchor=west, align=left] at (8.9,11.6) {Imagen 2};
\node[anchor=east, align=right] at (-0.5,2) {Imagen 2};
\node[rotate=35, anchor=west, align=left] at (9.9,11.6) {SDXL 1};
\node[anchor=east, align=right] at (-0.5,1) {SDXL 1};

\node[anchor=east] at (0,-1) {};
\end{tikzpicture}}
\end{minipage}}}
\\[1mm]
\resizebox{\linewidth}{!}{%
\mbox{%
\begin{minipage}{0.54\linewidth}
\resizebox{\linewidth}{!}{%
\begin{tikzpicture}
\begin{axis} [
        x label style={at={(0.5,-0.1)}},
        y label style={at={(-0.07,0.5)}},
        ylabel={Elo rating (with 99\% CI)},
        title={Prompt-image alignment},
        width=15cm,
        height=12cm,
    	ybar,
    	major grid style={line width=.2pt,draw=gray!50,dashed},
    	ymajorgrids,
    	tick style={draw=none},
    	xticklabels from table={\modelnamesalignment}{model},
    	xtick={0,...,10},
    	xmin=-0.5,
    	xmax=10.5,
    	xticklabel style={yshift=0mm,xshift=3mm, align=center,rotate=30,anchor=east},
    	every node near coord/.append style={xshift=0,yshift=-22,color=black,font=\small},
    	every axis plot/.append style={
          ybar,
          bar width=0.85,
          bar shift=0pt,
          draw=none
        },
        nodes near coords,
]
\addplot+ [mygray, error bars/.cd,
            y dir=both,
            y explicit,
            error bar style={black}]
        table
          [x=index,
            y=elo,
            y error plus expr=\thisrow{high_99_ci}-\thisrow{elo},
            y error minus expr=\thisrow{elo}-\thisrow{low_99_ci}
            ] {\datatableAalignment};
\addplot+ [myblue, error bars/.cd,
            y dir=both,
            y explicit,
            error bar style={black}]
        table
          [x=index,
            y=elo,
            y error plus expr=\thisrow{high_99_ci}-\thisrow{elo},
            y error minus expr=\thisrow{elo}-\thisrow{low_99_ci}
            ] {\datatableBalignment};     

\end{axis}
\end{tikzpicture}}
\end{minipage}
\hspace{2mm}
\begin{minipage}{0.46\linewidth}
\resizebox{\linewidth}{!}{%
\begin{tikzpicture}[scale=1]
\draw[very thin] (-0.5,10.5) rectangle +(1,1);
\draw[very thin, fill={rgb,1:red,0.993;green,0.740;blue,0.429}] (-0.5,9.5) rectangle +(1,1);
\node[black] at (0,10) {43.2\strut};
\draw[very thin, fill={rgb,1:red,0.989;green,0.657;blue,0.369}] (-0.5,8.5) rectangle +(1,1);
\node[black] at (0,9) {41.6\strut};
\draw[very thin, fill={rgb,1:red,0.992;green,0.686;blue,0.384}] (-0.5,7.5) rectangle +(1,1);
\node[black] at (0,8) {42.1\strut};
\draw[very thin, fill={rgb,1:red,0.971;green,0.527;blue,0.309}] (-0.5,6.5) rectangle +(1,1);
\node[black] at (0,7) {39.5\strut};
\draw[very thin] (-0.5,5.5) rectangle +(1,1);
\draw[very thin] (-0.5,4.5) rectangle +(1,1);
\draw[very thin] (-0.5,3.5) rectangle +(1,1);
\draw[very thin] (-0.5,2.5) rectangle +(1,1);
\draw[very thin] (-0.5,1.5) rectangle +(1,1);
\draw[very thin] (-0.5,0.5) rectangle +(1,1);
\draw[very thin, fill={rgb,1:red,0.710;green,0.876;blue,0.454}] (0.5,10.5) rectangle +(1,1);
\node[black] at (1,11) {56.8\strut};
\draw[very thin] (0.5,9.5) rectangle +(1,1);
\draw[very thin, fill={rgb,1:red,0.999;green,0.959;blue,0.681}] (0.5,8.5) rectangle +(1,1);
\node[black] at (1,9) {48.6\strut};
\draw[very thin, fill={rgb,1:red,0.998;green,0.936;blue,0.641}] (0.5,7.5) rectangle +(1,1);
\node[black] at (1,8) {47.8\strut};
\draw[very thin, fill={rgb,1:red,0.995;green,0.848;blue,0.519}] (0.5,6.5) rectangle +(1,1);
\node[black] at (1,7) {45.3\strut};
\draw[very thin, fill={rgb,1:red,0.986;green,0.637;blue,0.360}] (0.5,5.5) rectangle +(1,1);
\node[black] at (1,6) {41.2\strut};
\draw[very thin] (0.5,4.5) rectangle +(1,1);
\draw[very thin] (0.5,3.5) rectangle +(1,1);
\draw[very thin] (0.5,2.5) rectangle +(1,1);
\draw[very thin] (0.5,1.5) rectangle +(1,1);
\draw[very thin] (0.5,0.5) rectangle +(1,1);
\draw[very thin, fill={rgb,1:red,0.626;green,0.840;blue,0.413}] (1.5,10.5) rectangle +(1,1);
\node[black] at (2,11) {58.4\strut};
\draw[very thin, fill={rgb,1:red,0.950;green,0.979;blue,0.681}] (1.5,9.5) rectangle +(1,1);
\node[black] at (2,10) {51.4\strut};
\draw[very thin] (1.5,8.5) rectangle +(1,1);
\draw[very thin, fill={rgb,1:red,0.999;green,0.979;blue,0.713}] (1.5,7.5) rectangle +(1,1);
\node[black] at (2,8) {49.3\strut};
\draw[very thin, fill={rgb,1:red,0.968;green,0.986;blue,0.705}] (1.5,6.5) rectangle +(1,1);
\node[black] at (2,7) {50.9\strut};
\draw[very thin, fill={rgb,1:red,0.994;green,0.755;blue,0.442}] (1.5,5.5) rectangle +(1,1);
\node[black] at (2,6) {43.5\strut};
\draw[very thin, fill={rgb,1:red,0.993;green,0.748;blue,0.435}] (1.5,4.5) rectangle +(1,1);
\node[black] at (2,5) {43.3\strut};
\draw[very thin] (1.5,3.5) rectangle +(1,1);
\draw[very thin] (1.5,2.5) rectangle +(1,1);
\draw[very thin] (1.5,1.5) rectangle +(1,1);
\draw[very thin] (1.5,0.5) rectangle +(1,1);
\draw[very thin, fill={rgb,1:red,0.655;green,0.853;blue,0.418}] (2.5,10.5) rectangle +(1,1);
\node[black] at (3,11) {57.9\strut};
\draw[very thin, fill={rgb,1:red,0.921;green,0.967;blue,0.641}] (2.5,9.5) rectangle +(1,1);
\node[black] at (3,10) {52.2\strut};
\draw[very thin, fill={rgb,1:red,0.974;green,0.989;blue,0.713}] (2.5,8.5) rectangle +(1,1);
\node[black] at (3,9) {50.7\strut};
\draw[very thin] (2.5,7.5) rectangle +(1,1);
\draw[very thin, fill={rgb,1:red,0.999;green,0.969;blue,0.697}] (2.5,6.5) rectangle +(1,1);
\node[black] at (3,7) {48.9\strut};
\draw[very thin, fill={rgb,1:red,0.995;green,0.809;blue,0.487}] (2.5,5.5) rectangle +(1,1);
\node[black] at (3,6) {44.5\strut};
\draw[very thin, fill={rgb,1:red,0.994;green,0.778;blue,0.461}] (2.5,4.5) rectangle +(1,1);
\node[black] at (3,5) {43.9\strut};
\draw[very thin, fill={rgb,1:red,0.995;green,0.809;blue,0.487}] (2.5,3.5) rectangle +(1,1);
\node[black] at (3,4) {44.5\strut};
\draw[very thin] (2.5,2.5) rectangle +(1,1);
\draw[very thin] (2.5,1.5) rectangle +(1,1);
\draw[very thin] (2.5,0.5) rectangle +(1,1);
\draw[very thin, fill={rgb,1:red,0.498;green,0.784;blue,0.399}] (3.5,10.5) rectangle +(1,1);
\node[black] at (4,11) {60.5\strut};
\draw[very thin, fill={rgb,1:red,0.820;green,0.924;blue,0.525}] (3.5,9.5) rectangle +(1,1);
\node[black] at (4,10) {54.7\strut};
\draw[very thin, fill={rgb,1:red,0.999;green,0.974;blue,0.705}] (3.5,8.5) rectangle +(1,1);
\node[black] at (4,9) {49.1\strut};
\draw[very thin, fill={rgb,1:red,0.962;green,0.984;blue,0.697}] (3.5,7.5) rectangle +(1,1);
\node[black] at (4,8) {51.1\strut};
\draw[very thin] (3.5,6.5) rectangle +(1,1);
\draw[very thin, fill={rgb,1:red,0.995;green,0.840;blue,0.513}] (3.5,5.5) rectangle +(1,1);
\node[black] at (4,6) {45.2\strut};
\draw[very thin, fill={rgb,1:red,0.994;green,0.786;blue,0.468}] (3.5,4.5) rectangle +(1,1);
\node[black] at (4,5) {44.1\strut};
\draw[very thin, fill={rgb,1:red,0.992;green,0.686;blue,0.384}] (3.5,3.5) rectangle +(1,1);
\node[black] at (4,4) {42.1\strut};
\draw[very thin, fill={rgb,1:red,0.989;green,0.657;blue,0.369}] (3.5,2.5) rectangle +(1,1);
\node[black] at (4,3) {41.5\strut};
\draw[very thin] (3.5,1.5) rectangle +(1,1);
\draw[very thin] (3.5,0.5) rectangle +(1,1);
\draw[very thin] (4.5,10.5) rectangle +(1,1);
\draw[very thin, fill={rgb,1:red,0.607;green,0.832;blue,0.411}] (4.5,9.5) rectangle +(1,1);
\node[black] at (5,10) {58.8\strut};
\draw[very thin, fill={rgb,1:red,0.725;green,0.883;blue,0.464}] (4.5,8.5) rectangle +(1,1);
\node[black] at (5,9) {56.5\strut};
\draw[very thin, fill={rgb,1:red,0.780;green,0.907;blue,0.499}] (4.5,7.5) rectangle +(1,1);
\node[black] at (5,8) {55.5\strut};
\draw[very thin, fill={rgb,1:red,0.812;green,0.920;blue,0.520}] (4.5,6.5) rectangle +(1,1);
\node[black] at (5,7) {54.8\strut};
\draw[very thin] (4.5,5.5) rectangle +(1,1);
\draw[very thin, fill={rgb,1:red,0.927;green,0.969;blue,0.649}] (4.5,4.5) rectangle +(1,1);
\node[black] at (5,5) {51.9\strut};
\draw[very thin, fill={rgb,1:red,0.995;green,0.817;blue,0.493}] (4.5,3.5) rectangle +(1,1);
\node[black] at (5,4) {44.7\strut};
\draw[very thin, fill={rgb,1:red,0.995;green,0.809;blue,0.487}] (4.5,2.5) rectangle +(1,1);
\node[black] at (5,3) {44.6\strut};
\draw[very thin, fill={rgb,1:red,0.899;green,0.305;blue,0.207}] (4.5,1.5) rectangle +(1,1);
\node[white] at (5,2) {35.8\strut};
\draw[very thin] (4.5,0.5) rectangle +(1,1);
\draw[very thin] (5.5,10.5) rectangle +(1,1);
\draw[very thin] (5.5,9.5) rectangle +(1,1);
\draw[very thin, fill={rgb,1:red,0.718;green,0.880;blue,0.459}] (5.5,8.5) rectangle +(1,1);
\node[black] at (6,9) {56.7\strut};
\draw[very thin, fill={rgb,1:red,0.749;green,0.893;blue,0.479}] (5.5,7.5) rectangle +(1,1);
\node[black] at (6,8) {56.1\strut};
\draw[very thin, fill={rgb,1:red,0.757;green,0.897;blue,0.484}] (5.5,6.5) rectangle +(1,1);
\node[black] at (6,7) {55.9\strut};
\draw[very thin, fill={rgb,1:red,0.998;green,0.940;blue,0.649}] (5.5,5.5) rectangle +(1,1);
\node[black] at (6,6) {48.1\strut};
\draw[very thin] (5.5,4.5) rectangle +(1,1);
\draw[very thin, fill={rgb,1:red,0.999;green,0.964;blue,0.689}] (5.5,3.5) rectangle +(1,1);
\node[black] at (6,4) {48.8\strut};
\draw[very thin, fill={rgb,1:red,0.993;green,0.740;blue,0.429}] (5.5,2.5) rectangle +(1,1);
\node[black] at (6,3) {43.1\strut};
\draw[very thin, fill={rgb,1:red,0.976;green,0.567;blue,0.327}] (5.5,1.5) rectangle +(1,1);
\node[black] at (6,2) {40.2\strut};
\draw[very thin, fill={rgb,1:red,0.647;green,0.000;blue,0.149}] (5.5,0.5) rectangle +(1,1);
\node[white] at (6,1) {29.8\strut};
\draw[very thin] (6.5,10.5) rectangle +(1,1);
\draw[very thin] (6.5,9.5) rectangle +(1,1);
\draw[very thin] (6.5,8.5) rectangle +(1,1);
\draw[very thin, fill={rgb,1:red,0.780;green,0.907;blue,0.499}] (6.5,7.5) rectangle +(1,1);
\node[black] at (7,8) {55.5\strut};
\draw[very thin, fill={rgb,1:red,0.655;green,0.853;blue,0.418}] (6.5,6.5) rectangle +(1,1);
\node[black] at (7,7) {57.9\strut};
\draw[very thin, fill={rgb,1:red,0.788;green,0.910;blue,0.504}] (6.5,5.5) rectangle +(1,1);
\node[black] at (7,6) {55.3\strut};
\draw[very thin, fill={rgb,1:red,0.956;green,0.982;blue,0.689}] (6.5,4.5) rectangle +(1,1);
\node[black] at (7,5) {51.2\strut};
\draw[very thin] (6.5,3.5) rectangle +(1,1);
\draw[very thin, fill={rgb,1:red,0.991;green,0.996;blue,0.737}] (6.5,2.5) rectangle +(1,1);
\node[black] at (7,3) {50.3\strut};
\draw[very thin, fill={rgb,1:red,0.960;green,0.447;blue,0.272}] (6.5,1.5) rectangle +(1,1);
\node[black] at (7,2) {38.3\strut};
\draw[very thin, fill={rgb,1:red,0.747;green,0.096;blue,0.151}] (6.5,0.5) rectangle +(1,1);
\node[white] at (7,1) {32.0\strut};
\draw[very thin] (7.5,10.5) rectangle +(1,1);
\draw[very thin] (7.5,9.5) rectangle +(1,1);
\draw[very thin] (7.5,8.5) rectangle +(1,1);
\draw[very thin] (7.5,7.5) rectangle +(1,1);
\draw[very thin, fill={rgb,1:red,0.626;green,0.840;blue,0.413}] (7.5,6.5) rectangle +(1,1);
\node[black] at (8,7) {58.5\strut};
\draw[very thin, fill={rgb,1:red,0.780;green,0.907;blue,0.499}] (7.5,5.5) rectangle +(1,1);
\node[black] at (8,6) {55.4\strut};
\draw[very thin, fill={rgb,1:red,0.710;green,0.876;blue,0.454}] (7.5,4.5) rectangle +(1,1);
\node[black] at (8,5) {56.9\strut};
\draw[very thin, fill={rgb,1:red,1.000;green,0.993;blue,0.737}] (7.5,3.5) rectangle +(1,1);
\node[black] at (8,4) {49.7\strut};
\draw[very thin] (7.5,2.5) rectangle +(1,1);
\draw[very thin, fill={rgb,1:red,0.976;green,0.567;blue,0.327}] (7.5,1.5) rectangle +(1,1);
\node[black] at (8,2) {40.1\strut};
\draw[very thin, fill={rgb,1:red,0.824;green,0.170;blue,0.153}] (7.5,0.5) rectangle +(1,1);
\node[white] at (8,1) {33.5\strut};
\draw[very thin] (8.5,10.5) rectangle +(1,1);
\draw[very thin] (8.5,9.5) rectangle +(1,1);
\draw[very thin] (8.5,8.5) rectangle +(1,1);
\draw[very thin] (8.5,7.5) rectangle +(1,1);
\draw[very thin] (8.5,6.5) rectangle +(1,1);
\draw[very thin, fill={rgb,1:red,0.248;green,0.667;blue,0.350}] (8.5,5.5) rectangle +(1,1);
\node[white] at (9,6) {64.2\strut};
\draw[very thin, fill={rgb,1:red,0.538;green,0.801;blue,0.403}] (8.5,4.5) rectangle +(1,1);
\node[black] at (9,5) {59.8\strut};
\draw[very thin, fill={rgb,1:red,0.420;green,0.750;blue,0.390}] (8.5,3.5) rectangle +(1,1);
\node[black] at (9,4) {61.7\strut};
\draw[very thin, fill={rgb,1:red,0.538;green,0.801;blue,0.403}] (8.5,2.5) rectangle +(1,1);
\node[black] at (9,3) {59.9\strut};
\draw[very thin] (8.5,1.5) rectangle +(1,1);
\draw[very thin, fill={rgb,1:red,0.983;green,0.617;blue,0.350}] (8.5,0.5) rectangle +(1,1);
\node[black] at (9,1) {41.0\strut};
\draw[very thin] (9.5,10.5) rectangle +(1,1);
\draw[very thin] (9.5,9.5) rectangle +(1,1);
\draw[very thin] (9.5,8.5) rectangle +(1,1);
\draw[very thin] (9.5,7.5) rectangle +(1,1);
\draw[very thin] (9.5,6.5) rectangle +(1,1);
\draw[very thin] (9.5,5.5) rectangle +(1,1);
\draw[very thin, fill={rgb,1:red,0.000;green,0.408;blue,0.216}] (9.5,4.5) rectangle +(1,1);
\node[white] at (10,5) {70.2\strut};
\draw[very thin, fill={rgb,1:red,0.052;green,0.504;blue,0.266}] (9.5,3.5) rectangle +(1,1);
\node[white] at (10,4) {68.0\strut};
\draw[very thin, fill={rgb,1:red,0.092;green,0.578;blue,0.304}] (9.5,2.5) rectangle +(1,1);
\node[white] at (10,3) {66.5\strut};
\draw[very thin, fill={rgb,1:red,0.587;green,0.823;blue,0.409}] (9.5,1.5) rectangle +(1,1);
\node[black] at (10,2) {59.0\strut};
\draw[very thin] (9.5,0.5) rectangle +(1,1);

\node[rotate=35, anchor=west, align=left] at (-0.1,11.6) {Imagen 3-002};
\node[anchor=east, align=right] at (-0.5,11) {Imagen 3-002};
\node[rotate=35, anchor=west, align=left] at (0.9,11.6) {RecraftV3};
\node[anchor=east, align=right] at (-0.5,10) {RecraftV3};
\node[rotate=35, anchor=west, align=left] at (1.9,11.6) {IdeogramV2};
\node[anchor=east, align=right] at (-0.5,9) {IdeogramV2};
\node[rotate=35, anchor=west, align=left] at (2.9,11.6) {Imagen 3-001};
\node[anchor=east, align=right] at (-0.5,8) {Imagen 3-001};
\node[rotate=35, anchor=west, align=left] at (3.9,11.6) {Flux1.1p};
\node[anchor=east, align=right] at (-0.5,7) {Flux1.1p};
\node[rotate=35, anchor=west, align=left] at (4.9,11.6) {SD 3.5 L};
\node[anchor=east, align=right] at (-0.5,6) {SD 3.5 L};
\node[rotate=35, anchor=west, align=left] at (5.9,11.6) {DALL·E 3};
\node[anchor=east, align=right] at (-0.5,5) {DALL·E 3};
\node[rotate=35, anchor=west, align=left] at (6.9,11.6) {MJ v6};
\node[anchor=east, align=right] at (-0.5,4) {MJ v6};
\node[rotate=35, anchor=west, align=left] at (7.9,11.6) {NovaCanvas};
\node[anchor=east, align=right] at (-0.5,3) {NovaCanvas};
\node[rotate=35, anchor=west, align=left] at (8.9,11.6) {Imagen 2};
\node[anchor=east, align=right] at (-0.5,2) {Imagen 2};
\node[rotate=35, anchor=west, align=left] at (9.9,11.6) {SDXL 1};
\node[anchor=east, align=right] at (-0.5,1) {SDXL 1};

\node[anchor=east] at (0,-1) {};
\end{tikzpicture}}
\end{minipage}}}
\caption{{\bf Updated human evaluation on \genaibench{}:} Elo scores and win-rate percentages for (i) overall preference, (ii) visual quality, and (iii) prompt-image alignment.}
\label{fig:elo_updated_genai_bench}
\end{figure}

%% file: contributions.tex
\section*{Contributions}
\begin{minipage}[t]{1.0\linewidth}
\noindent
\begin{multicols}{2}
\setlength{\parindent}{0pt}
\textbf{Core Contributors} \\
Jason Baldridge \\
Jakob Bauer \\
Mukul Bhutani \\
Nicole Brichtova \\
Andrew Bunner \\
Lluis Castrejon \\
Kelvin Chan \\
Yichang Chen \\
Sander Dieleman \\
Yuqing Du \\
Zach Eaton-Rosen \\
Hongliang Fei \\
Nando de Freitas \\
Yilin Gao \\
Evgeny Gladchenko \\
Sergio Gómez Colmenarejo \\
Mandy Guo \\
Alex Haig \\
Will Hawkins \\
Hexiang (Frank) Hu \\
Huilian Huang \\
Tobenna Peter Igwe \\
Christos Kaplanis \\
Siavash Khodadadeh \\
Yelin Kim \\
Ksenia Konyushkova \\
Karol Langner \\
Eric Lau \\
Rory Lawton \\ 
Shixin Luo \\
\columnbreak
\\
\\
Soňa Mokrá \\
Henna Nandwani \\
Yasumasa Onoe \\
Aäron van den Oord \\
Zarana Parekh \\
Jordi Pont-Tuset \\
Hang Qi \\
Rui Qian \\
Deepak Ramachandran \\
Poorva Rane \\
Abdullah Rashwan \\
Ali Razavi \\
Robert Riachi \\
Hansa Srinivasan \\
Srivatsan Srinivasan \\
Robin Strudel \\
Benigno Uria \\
Oliver Wang \\
Su Wang \\
Austin Waters \\
Chris Wolff \\
Auriel Wright \\
Zhisheng Xiao \\
Hao Xiong \\
Keyang Xu \\
Marc van Zee \\
Junlin Zhang \\
Katie Zhang \\
Wenlei Zhou \\
Konrad Zolna \\
\end{multicols}
\end{minipage}

\begin{minipage}[t]{1.0\linewidth}
\noindent
\begin{multicols}{2}
\setlength{\parindent}{0pt}
\textbf{Contributors} \\
Ola Aboubakar \\
Canfer Akbulut \\
Oscar Akerlund \\
Isabela Albuquerque \\
Nina Anderson \\
Marco Andreetto \\
Lora Aroyo \\
Ben Bariach \\
David Barker \\
Praseem Banzal \\
Sherry Ben \\
Dana Berman \\
Courtney Biles \\
Irina Blok \\
Pankil Botadra \\
Jenny Brennan \\
Karla Brown \\
John Buckley \\
Rudy Bunel \\
Elie Bursztein \\
Christina Butterfield \\
Ben Caine \\
Viral Carpenter \\
Norman Casagrande \\
Ming-Wei Chang \\
Solomon Chang \\
Shamik Chaudhuri \\
Tony Chen \\
John Choi \\
Dmitry Churbanau \\
Nathan Clement \\
Matan Cohen \\
Forrester Cole \\
Romina Datta \\
Mikhail Dektiarev \\
Vincent Du \\
Praneet Dutta \\
Tom Eccles \\
Ndidi Elue \\
Ashley Feden \\
Shlomi Fruchter \\
Frankie Garcia \\
Roopal Garg \\
Weina Ge \\
Ahmed Ghazy \\
Bryant Gipson \\
Andrew Goodman \\
Dawid Górny \\
\columnbreak
\\
\\
Sven Gowal \\
Khyatti Gupta \\
Yoni Halpern \\
Yena Han \\
Susan Hao \\
Jamie Hayes \\
Jonathan Heek \\
Amir Hertz \\
Ed Hirst \\
Emiel Hoogeboom \\
Tingbo Hou \\
Heidi Howard \\
Mohamed Ibrahim \\
Dirichi Ike-Njoku \\
Joana Iljazi \\
Vlad Ionescu \\
William Isaac \\
Komal Jalan \\
Reena Jana \\
Gemma Jennings \\
Donovon Jenson \\
Xuhui Jia \\
Kerry Jones \\
Xiaoen Ju \\
Ivana Kajic \\
Christos Kaplanis \\
Burcu Karagol Ayan \\
Jacob Kelly \\
Suraj Kothawade \\
Christina Kouridi \\
Ira Ktena \\
Jolanda Kumakaw \\
Dana Kurniawan \\
Dmitry Lagun \\
Lily Lavitas \\
Jason Lee \\
Tao Li \\
Marco Liang \\
Ricky Liang \\
Maggie Li-Calis \\
Rui Lin \\
Jasmine Liu \\
Yuchi Liu \\
Javier Lopez Alberca \\
Matthieu Kim Lorrain \\
Peggy Lu \\
Kristian Lum \\
Yukun Ma \\
\end{multicols}
\end{minipage}

\begin{minipage}[t]{1.0\linewidth}
\noindent
\begin{multicols}{2}
\setlength{\parindent}{0pt}
Chase Malik \\
John Mellor \\
Thomas Mensink \\
Inbar Mosseri \\
Tom Murray \\
Aida Nematzadeh \\
Paul Nicholas \\
Signe Nørly \\
João Gabriel Oliveira \\
Guillermo Ortiz-Jimenez \\
Michela Paganini \\
Tom Le Paine \\
Roni Paiss \\
Alicia Parrish \\
Anne Peckham \\
Vikas Peswani \\
Igor Petrovski \\
Tobias Pfaff \\
Alex Pirozhenko \\
Ryan Poplin \\
Utsav Prabhu \\
Yuan Qi \\
Matthew Rahtz \\
Cyrus Rashtchian \\
Charvi Rastogi \\
Amit Raul \\
Ali Razavi \\
Sylvestre-Alvise Rebuffi \\
Susanna Ricco \\
Felix Riedel \\
Dirk Robinson \\
Pankaj Rohatgi \\
Bill Rosgen \\
Sarah Rumbley \\
Moonkyung Ryu \\
Anthony Salgado \\
Tim Salimans \\
Eleni Shaw \\
Gregory Shaw \\
Sahil Singla \\
Florian Schroff \\
Candice Schumann \\
Tanmay Shah \\
Brendan Shillingford \\
Kaushik Shivakumar \\
Dennis Shtatnov \\
Zach Singer \\
Evgeny Sluzhaev \\
Valerii Sokolov \\
Thibault Sottiaux \\
Florian Stimberg \\
Brad Stone \\
David Stutz \\
Yu-Chuan Su \\
Eric Tabellion \\
Amit Talreja \\
Shuai Tang \\
David Tao \\
Kurt Thomas \\
Gregory Thornton \\
Andeep Toor \\
Cristian Udrescu \\
Aayush Upadhyay \\
Cristina Vasconcelos \\
Shanthal Vasanth \\
Alex Vasiloff \\
Andrey Voynov \\
Amanda Walker \\
Luyu Wang \\
Miaosen Wang \\
Simon Wang \\
Stanley Wang \\
Qifei Wang \\
Yuxiao Wang \\
Ágoston Weisz \\
Olivia Wiles \\
Chenxia Wu \\
Xingyu Federico Xu \\
Andrew Xue \\
Jianbo Yang \\
Luo Yu \\
Mete Yurtoglu \\
Ali Zand \\
Han Zhang \\
Jiageng Zhang \\
Catherine Zhao \\
Adilet Zhaxybay \\
Miao Zhou \\
Shengqi Zhu \\
Zhenkai Zhu \\
\end{multicols}
\end{minipage}

\begin{minipage}[t]{1.0\linewidth}
\noindent
\begin{multicols}{2}
\setlength{\parindent}{0pt}
\textbf{Advisors} \\
Dawn Bloxwich \\
Mahyar Bordbar \\
Luis C. Cobo \\
Eli Collins \\
Shengyang Dai \\
Tulsee Doshi \\
Anca Dragan \\
Douglas Eck \\
Demis Hassabis \\
Sissie Hsiao \\
Tom Hume \\
\columnbreak
\\
\\
Koray Kavukcuoglu \\
Helen King \\
Jack Krawczyk \\
Yeqing Li \\
Kathy Meier-Hellstern \\
Andras Orban \\
Yury Pinsky \\
Amar Subramanya \\
Oriol Vinyals \\
Ting Yu \\
Yori Zwols \\
\end{multicols}
\end{minipage}

\setlength{\footskip}{0.4cm}
\hspace{10pt}

\noindent The roles are defined as below:
\begin{itemize}
    \item \textit{Core Contributor}: Individual that had significant impact throughout the project.
    \item \textit{Contributor}: Individual that had contributions to the project and was partially involved with the effort. 
    \item \textit{Advisor}: Individual who provided guidance and expertise to the project.
\end{itemize}

Within each role, contributions are equal, and are listed in alphabetical order. Ordering within each role does not indicate ordering of the contributions.